\documentclass[10pt,twocolumn,letterpaper]{article}

\usepackage[pagenumbers]{iccv} 
\usepackage{amsmath}
\usepackage{multirow}
\usepackage{makecell}
\usepackage{colortbl}  

\definecolor{myred}{RGB}{249, 221, 223} 
\definecolor{myblue}{RGB}{221, 222, 237} 
\newcommand{\best}[1]{\cellcolor{myred}\textbf{#1}}
\newcommand{\second}[1]{\cellcolor{myblue}\underline{#1}}

\definecolor{iccvblue}{rgb}{0.21,0.49,0.74}

\usepackage[pagebackref,breaklinks,colorlinks,allcolors=iccvblue]{hyperref}

\def\paperID{6812} 
\def\confName{ICCV}
\def\confYear{2025}

\title{Adaptive Blind All-in-One Image Restoration}

\author{David Serrano-Lozano$^{1,2}$ \quad Luis Herranz$^{3}$ \quad Shaolin Su$^{1}$ \quad Javier Vazquez-Corral$^{1,2}$\\
{\normalsize $^1$Computer Vision Center} \quad
{\normalsize $^2$Universitat Autònoma de Barcelona} \quad
{\normalsize $^3$Universidad Autónoma de Madrid} \\
{\normalsize \texttt{\{dserrano, shaolin, javier.vazquez\}@cvc.uab.cat \quad luis.herranz@uam.es}
}}

\begin{document}
\maketitle

\begin{abstract}
Blind all-in-one image restoration models aim to recover a high-quality image from an input degraded with unknown distortions. However, these models require all the possible degradation types to be defined during the training stage while showing limited generalization to unseen degradations, which limits their practical application in complex cases. In this paper, we introduce ABAIR, a simple yet effective adaptive blind all-in-one restoration model that not only handles multiple degradations and generalizes well to unseen distortions but also efficiently integrates new degradations by training only a small subset of parameters. We first train our baseline model on a large dataset of natural images with multiple synthetic degradations. To enhance its ability to recognize distortions, we incorporate a segmentation head that estimates per-pixel degradation types. Second, we adapt our initial model to varying image restoration tasks using independent low-rank adapters. Third, we learn to adaptively combine adapters to versatile images via a flexible and lightweight degradation estimator. This specialize-then-merge approach is both powerful in addressing specific distortions and flexible in adapting to complex tasks. Moreover, our model not only surpasses state-of-the-art performance on five- and three-task IR setups but also demonstrates superior generalization to unseen degradations and composite distortions. \\\url{https://aba-ir.github.io/}
\end{abstract}


\section{Introduction}\label{sec:intro}
\begin{figure}[t] \centering
    \includegraphics[width=\linewidth]{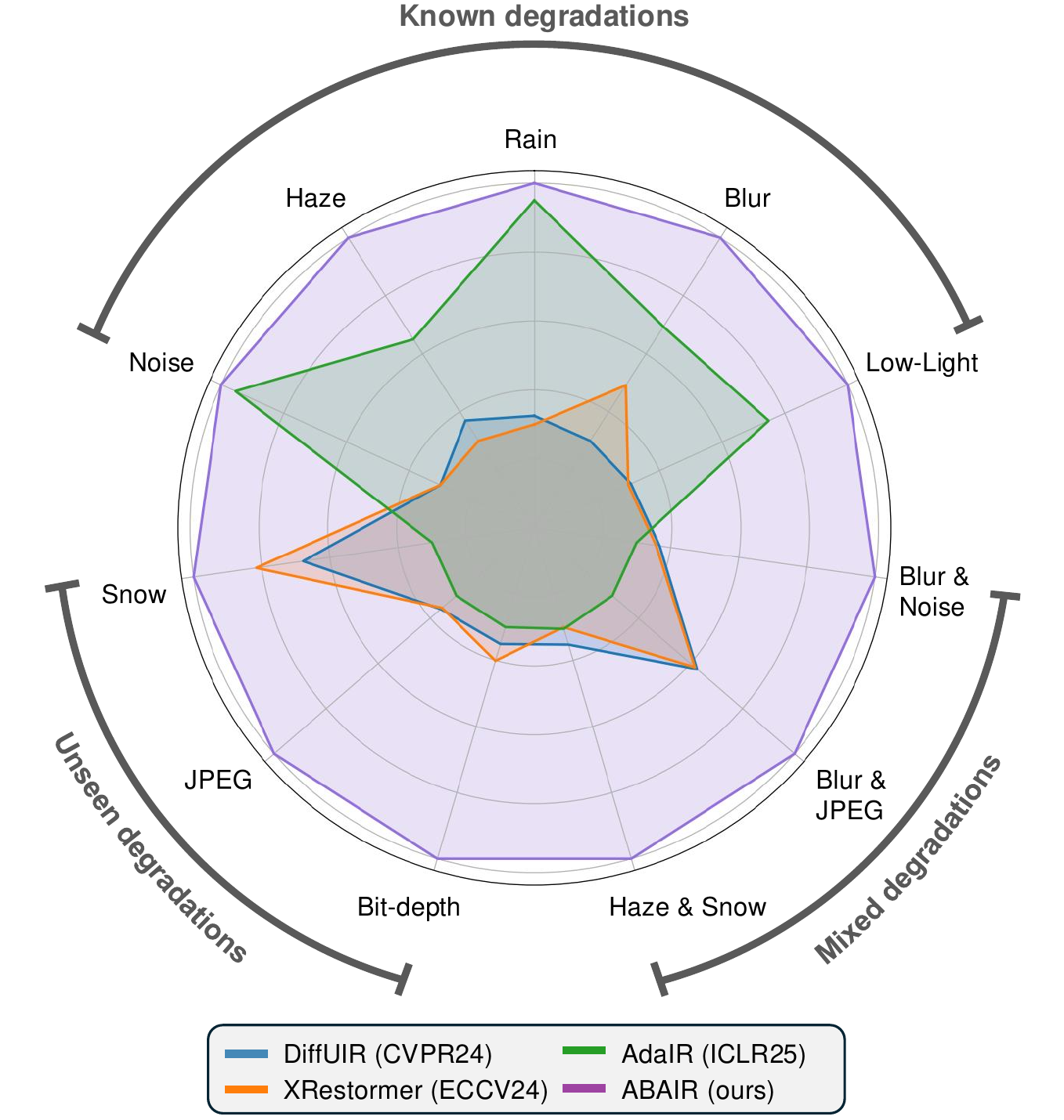}
    \vspace{-5mm}
    \caption{Our model significantly outperforms state-of-the-art all-in-one image restoration (IR) methods, DiffUIR~\cite{zheng2024selective}, X-Restormer~\cite{chen2023x-restormer}, and AdaIR~\cite{cui2025adair}, across five known IR tasks, three unseen tasks, and three mixed degradation scenarios. The plot is normalized along each axis, with the lowest value positioned on the second circle and the highest value on the outermost circle.} 
    \vspace{-5mm}
    \label{fig:teaser}
\end{figure}

Image restoration (IR) is a fundamental task in computer vision, essential for enhancing visual quality and optimizing the performance of downstream tasks~\cite{sun2022rethinking, niu2020effective}. IR aims to reconstruct high-fidelity images by systematically removing various degradations present in low-quality inputs. These degradations often emerge through a complex interplay of external environmental conditions and camera limitations during the image acquisition process such as adverse weather conditions~\cite{Li2020allinone, valanarasu2022transweather, wang2024selfpromer}, noise~\cite{dong2018denoising}, blur~\cite{ren2023multiscale} and low-light environments~\cite{cai2023retinexformer, zhou2023pyramid}. 

The inherently ill-posed nature of IR presents a significant challenge for conventional approaches limiting their effectiveness~\cite{oliveira2009adaptive, kindermann2005deblurring, michaeli2013nonparametric, kim2010single, timofte2013anchored, dong2011image, he2010darkchennelprior, farsiu2004fast}. Recent advances in deep learning techniques have led to remarkable progress in IR~\cite{li2022airnet, zamir2022restormer, potlapalli2024promptir, ren2023multiscale, delbracio2023inversion}, achieving substantial improvements in reconstruction accuracy. However, these frameworks demonstrate superior performance only in modeling dedicated degradations, as they are typically trained on IR datasets with specific degradations.

To address the limitations of using separate models for each degradation type, recent IR models have adopted an all-in-one approach, \emph{i.e.}, they are designed to handle multiple degradation types within a single model, alleviating the dependency on dedicated models for various IR tasks. While these models have demonstrated success in integrating diverse IR tasks into a unified framework, several challenges persist when applied to real-world IR problems. First, these models assume that the type of degradations are specified in advance, allowing them to target and remove them specifically. In practice, however, IR must operate in a blind setting, with no prior knowledge of the degradation present in a given image, making blind IR significantly more challenging. Second, real images frequently contain composite degradations --- \emph{e.g.}, a moving vehicle in a low-light scene, or a compressed image captured in hazy weather. Although existing models are designed to handle diverse degradation types, they typically process only a single type of distortion at a time, limiting their effectiveness in practical cases. Third, current all-in-one methods require access to all degradation types during training, which restricts their generalizability to unseen degradations. Adapting these models to new, unobserved degradation types while retaining the all-in-one functionality typically necessitates retraining the entire model with the expanded set of degradations --- a process that is both computationally expensive and time-consuming.

\textbf{Contribution: }This paper presents an Adaptive Blind All-in-One Image Restoration (ABAIR) method, designed to bridge the gap between IR techniques and their application in practical complex scenarios. ABAIR effectively addresses multiple and composite degradations with a flexible structure that can be easily updated to include new degradations. Our approach is a simple yet effective scheme that combines three main components. First, we propose a large-scale pretraining with synthetic degradations to obtain a robust weight initialization. To increase variability, we introduce Degradation CutMix~\cite{yun2019cutmix}, a technique for generating images with controlled mixtures of degradations. Additionally, we enhance this pretraining by incorporating a segmentation head to recognize per-pixel degradation types. Second, we train independent restoration adapters --- specifically LoRA~\cite{hu2021lora} --- to bridge the gap between synthetic and real data. Third, we train a lightweight image-degradation estimator to select the most suitable combination of adapters, enabling a blind all-in-one IR method. By leveraging the synthetic pretraining and the specialize-then-merge scheme, our model achieves handling composite distortions within a single image. Moreover, our pretraining and adapter-based design provides a flexible and extensible model structure that can be easily updated to address new distortions. By training a new adapter for the added degradation and retraining the lightweight estimator, our approach preserves prior knowledge of other IR tasks without requiring full model retraining. This enables a blind all-in-one model for versatile IR, achieving state-of-the-art performance on all-in-one IR benchmarks, and generalizing to unseen and mixed degradations, as shown in Figure~\ref{fig:teaser}.

\section{Related Work}\label{sec:relatedwork}
\vspace{-1mm}
\paragraph{Single Degradation Image Restoration}
Most prior work on IR has typically considered removing a single type of degradation in an image to recover its clean counterpart. Typical single degradation IR tasks include denoising~\cite{dong2018denoising}, deblurring~\cite{ren2023multiscale}, deraining~\cite{chen2023learning}, dehazing~\cite{yang2022self}, low-light enhancement~\cite{cai2023retinexformer, zhou2023pyramid}, etc. Though achieving promising progress on individual tasks, these methods are only capable of dealing with a specific type of distortion, thus limiting their generalizability to a wider range of IR scenarios.

\vspace{-4mm}
\paragraph{All-in-One Image Restoration}
Recently, multi-degradation and all-in-one IR approaches have gained significant attention. Multi-degradation methods~\cite{zamir2022restormer, chen2022nafnet, zamir2021multi, yue2024efficient, park2024colora, tang2024RCOT, guo2024mambair, luo2023controlling, guo2024test, conde2024instructir} proposed model architectures that are effective on distinct IR tasks. However, these methods are trained so that one set of parameters can handle only one specific type of degradation. Therefore, one still needs to allocate different weights for different degradations. Furthermore, to restore a versatile image, the type of degradation has to be known so that the corresponding parameters can be loaded, this non-blind scheme further hinders the efficiency and effectiveness for real-world applications.


Meanwhile, blind all-in-one IR approaches utilize specialized modules to distinguish degradation types blindly. For instance, AirNet~\cite{li2022airnet} and OneRestore~\cite{guo2024onerestore} use a contrastive loss to extract a latent degradation representation from the input images. X-Restormer~\cite{chen2023x-restormer} adds a spatial self-attention module to the transformer block to enhance spatial mapping capabilities. AdaIR~\cite{cui2025adair} operates in the frequency domain to distinguish the distinct degradation types while IDR~\cite{zhang2023idr} learns degradation-specific priors and incorporates them into restoration. Despite the ability to blindly process images, these methods are still incapable of dealing with composite distortions or unseen distortions, due to their focus on dedicated degradations and standard IR benchmarks. 


\begin{figure*}[ht!]
    \centering
    \includegraphics[width=\linewidth]{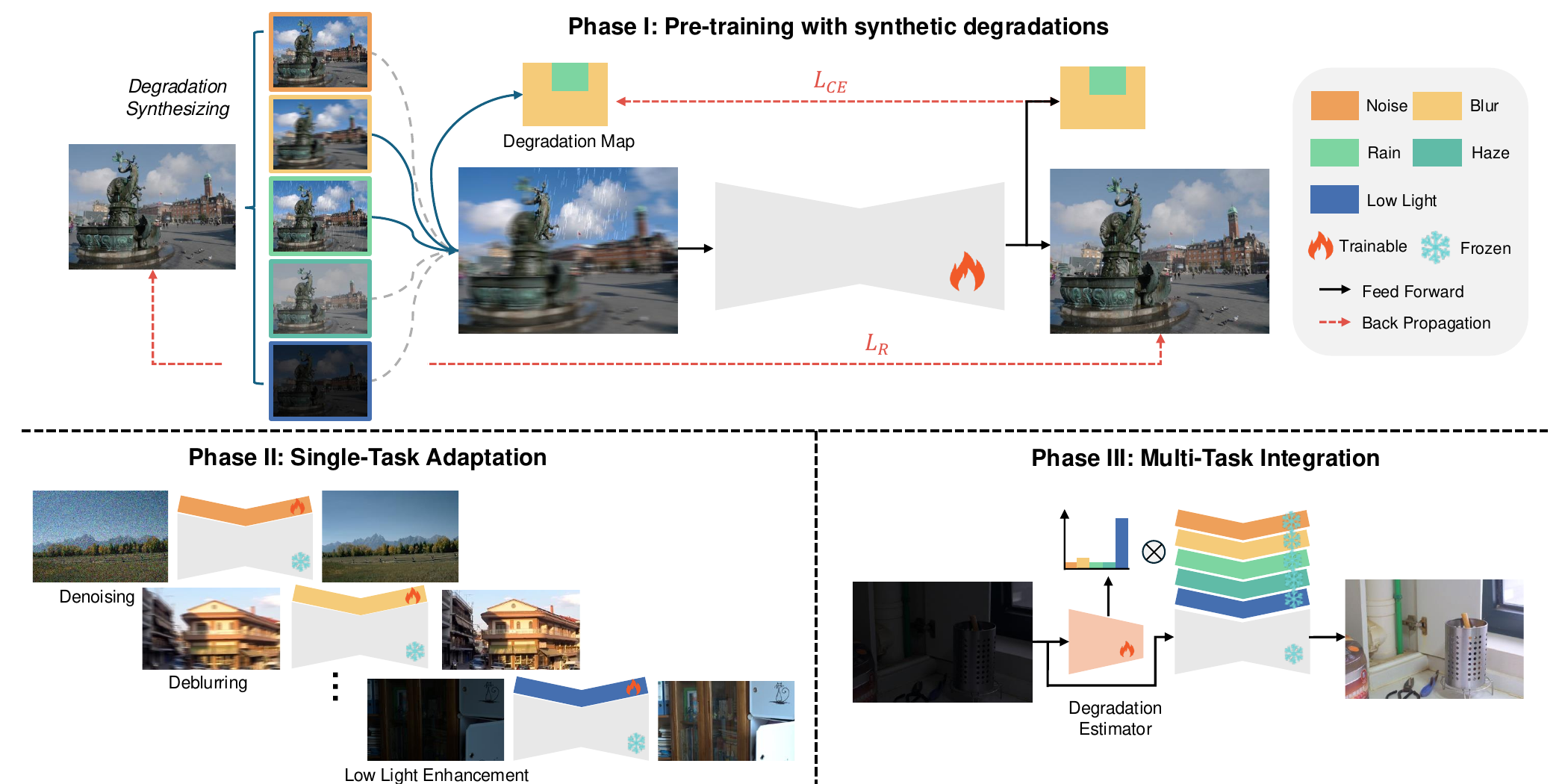}
    \vspace{-6mm}
    \caption{General schema of our proposed method. Our method is divided into three phases. In Phase I we pretrain our baseline with synthetic degradations on high-fidelity images. Each image contains different degradations in different regions --- Degradation CutMix~\cite{yun2019cutmix}, and a segmentation head learns to predict them, while a restoration loss aims at restoring the image. In this way, the model can distinguish and generalize well to multiple degradations in a controlled manner. In Phase II, we learn degradation-specific adaptors using standard image restoration datasets. In Phase III, we learn a lightweight degradation estimator to adaptively blend the adapters based on the degradation profile of the input image. This three-phase methodology makes our method flexible to deal with images containing multiple degradations and easy to update for new ones as it only requires training an adapter for the new distortion and retraining the lightweight estimator.}
    \label{fig:method}
    \vspace{-4mm}
\end{figure*}

Prompt learning techniques, which capture task-specific context, have shown promise in guiding adaptation for vision tasks~\cite{jia2022visual}. In all-in-one IR, PromptIR~\cite{potlapalli2024promptir} integrates a dedicated prompt block to capture degradation-specific features from input images, while DA-CLIP~\cite{luo2023daclip}, MPerceiver~\cite{ai2024multimodal}, ProRes~\cite{ma2023prores} and Painter~\cite{wang2023painter} leverage large pretrained models as prompt generators. However, the latter approaches, as well as diffusion-based methods~\cite{zheng2024selective, jiang2024autodir, lin2024diffbir}, are often constrained by the high memory demands of large models and iterative processes, respectively. Additionally, existing all-in-one approaches require all degradation types to be predefined during training, limiting the addition of new distortion to already trained models. All-in-one IR also places great emphasis on adverse weather removal~\cite{valanarasu2022transweather, ozdenizci2023restoring, zhu2024mwformer, zhu2023learning}, compression~\cite{zeng2025all, ren2024moediffir, li2024promptcir}, video~\cite{zhao2025avernet} or techniques to improve existing architectures~\cite{qin2024restore, wu2024harmony}.

In contrast, aiming at simple and versatile image restoration, we propose a three-phase approach to blindly process images with various degradations, either in single or composite forms. First, we introduce a synthetic pretraining stage to establish a strong baseline model. Next, we employ a specialize-then-merge strategy, where the model is first adapted to individual real-world degradations before learning to address the problem in a blind all-in-one manner.


\begin{figure*}[ht!]
    \centering
    \begin{minipage}{0.16\textwidth}
        \centering
        \includegraphics[width=\linewidth]{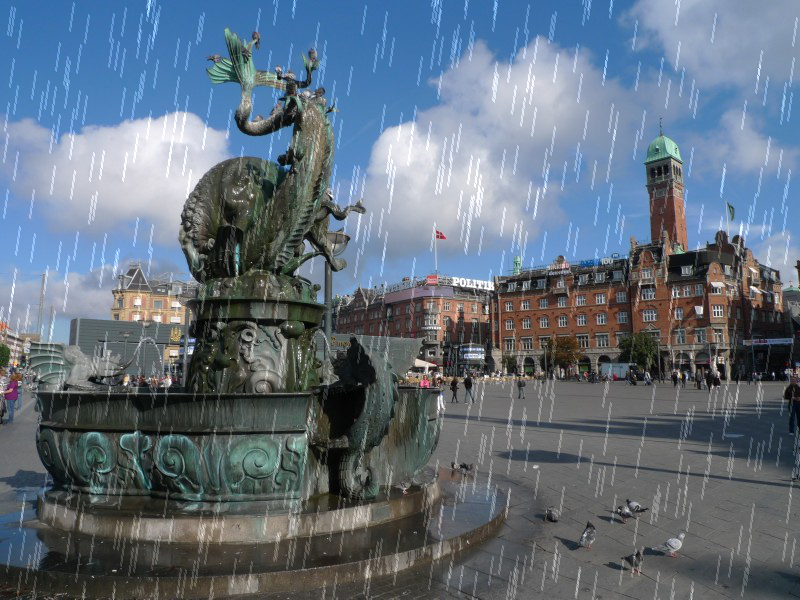}
        \subcaption{Rain}\label{fig:rain}
    \end{minipage}
    \begin{minipage}{0.16\textwidth}
        \centering
        \includegraphics[width=\linewidth]{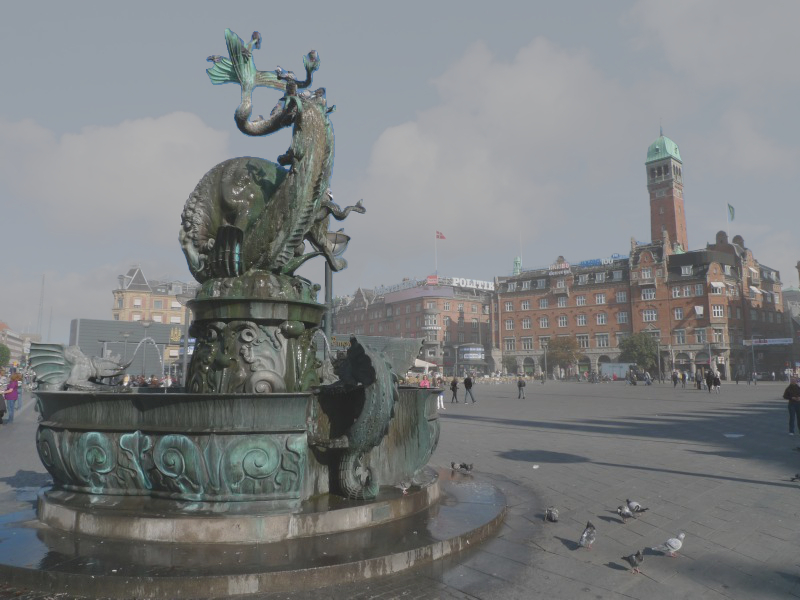}
        \subcaption{Haze}\label{fig:haze}
    \end{minipage}
    \begin{minipage}{0.16\textwidth}
        \centering
        \includegraphics[width=\linewidth]{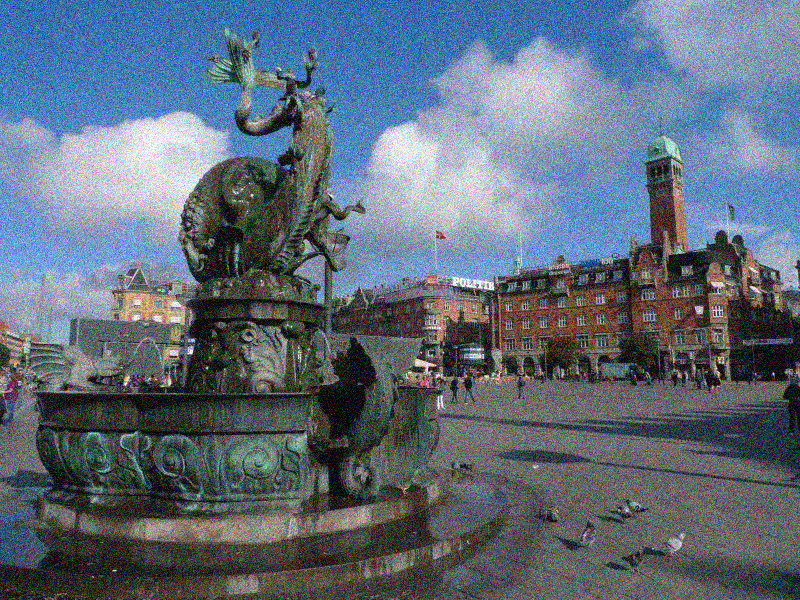}
        \subcaption{Noise}\label{fig:noise}
    \end{minipage}
    \begin{minipage}{0.16\textwidth}
        \centering
        \includegraphics[width=\linewidth]{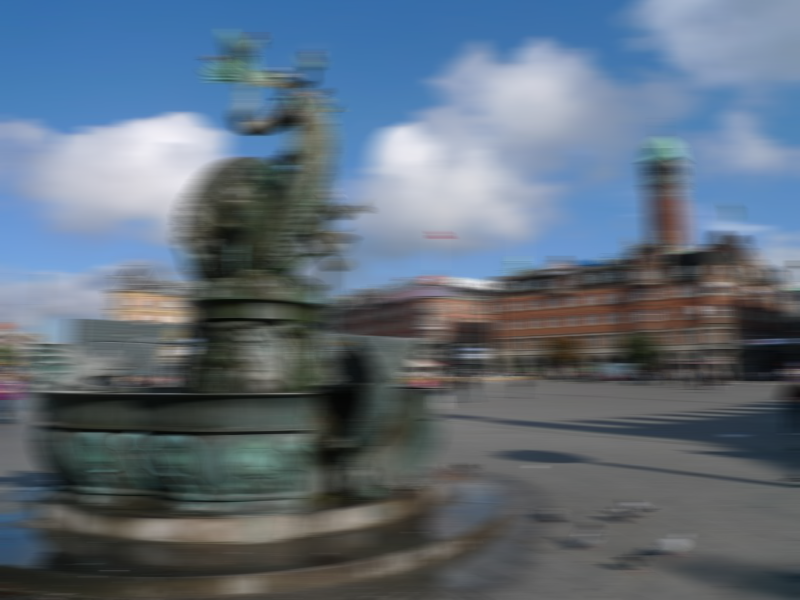}
        \subcaption{Blur}\label{fig:blur}
    \end{minipage}
    \begin{minipage}{0.16\textwidth}
        \centering
        \includegraphics[width=\linewidth]{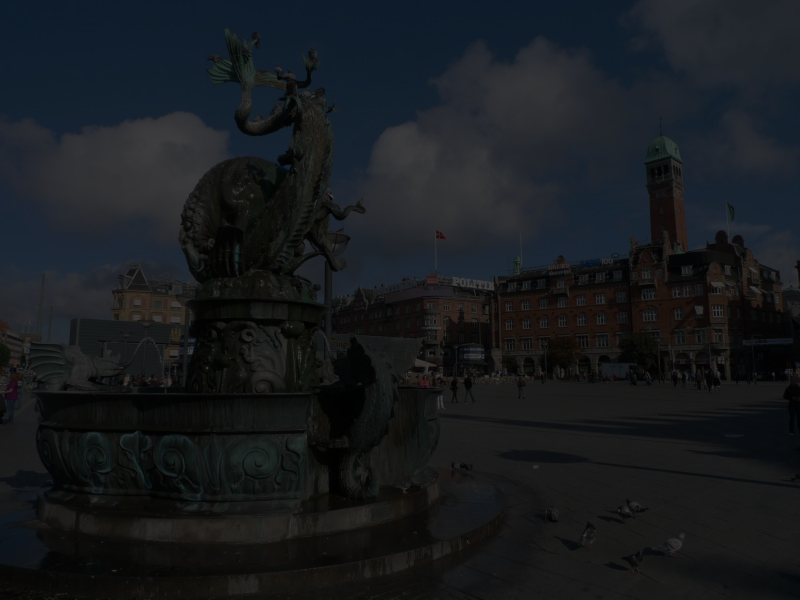}
        \subcaption{Low-light}\label{fig:low-light}
    \end{minipage}
    \begin{minipage}{0.16\textwidth}
       \centering
       \includegraphics[width=\linewidth]{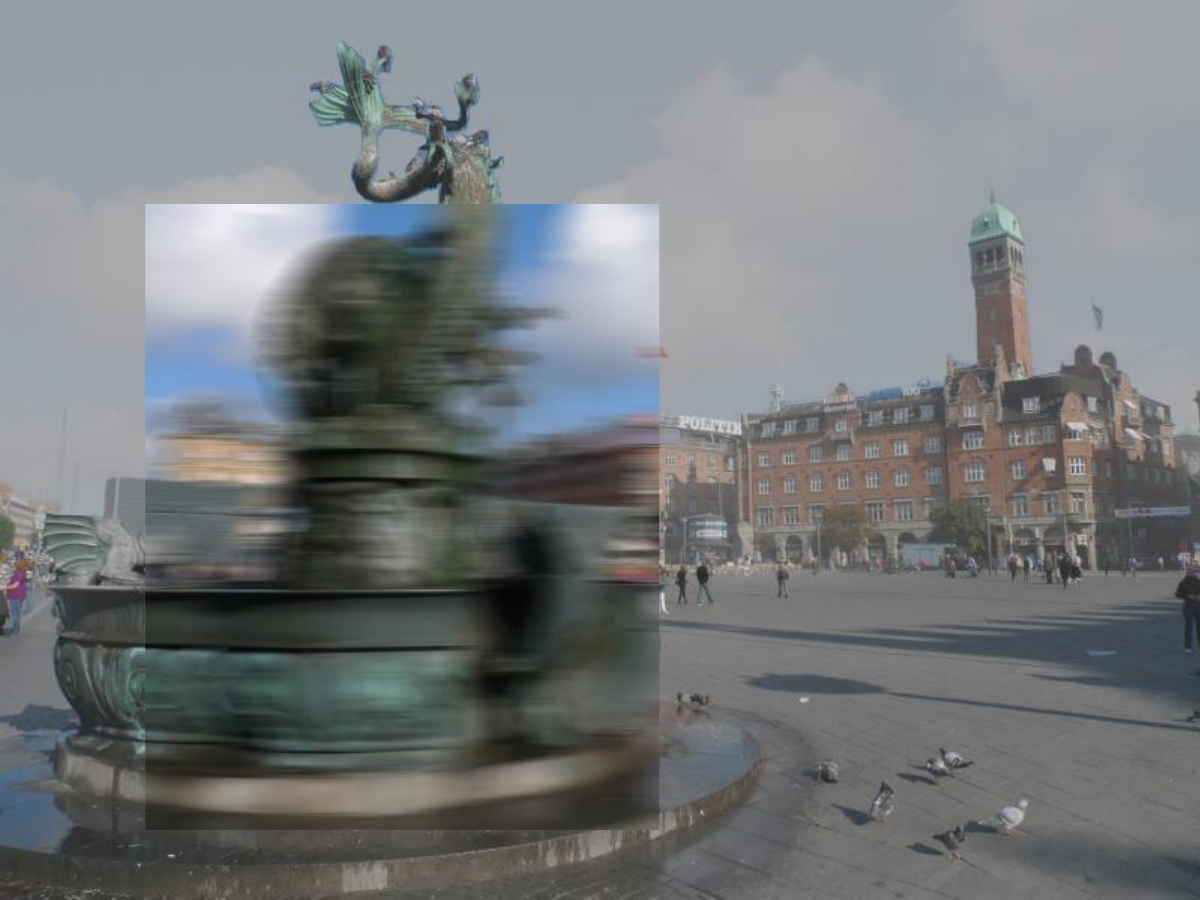}
       \subcaption{Deg. Cutmix}\label{fig:cutmix}
    \end{minipage}
    \vspace{-3mm}
    \caption{Examples of our synthetic degradation generation for five traditional degradations (a)-(e) and our Degradation Cutmix (f).}
    \vspace{-4mm}

    \label{fig:synthetic_examples}
\end{figure*}

\vspace{-4mm}
\paragraph{Parameter Efficient Fine-Tuning}
Fine-tuning all model parameters is computationally expensive, especially for large-scale models, but Parameter Efficient Fine-Tuning (PEFT) addresses this by reducing the number of trainable parameters and memory usage while maintaining performance~\cite{xu2023parameter}. Low-Rank Adaptation (LoRA)~\cite{hu2021lora} introduced a reparametrization strategy for fine-tuning, where the weights of a specific layer $W' = W + \Delta W$ are represented as a linear combination of the pretrained weights $W$ and the updated term $\Delta W$ obtained through a low-rank decomposition, $\Delta W = BA$. Here, $B \in \mathbb{R}^{d \times k}$, $A \in \mathbb{R}^{r \times k}$ with rank $r \ll \text{min}(d,k)$. By optimizing only the low-rank matrices $A$ and $B$, LoRA reduces the number of parameters required for adapting that layer, minimizing memory and computational demands. Building upon LoRA, alternative decomposition methods have been proposed, such as VeRA~\cite{kopiczko2024vera} and Conv-LoRA\cite{zhong2024convlora}.

PEFT techniques have proven valuable not only for domain adaptation in large models but also for applications like task arithmetic and continual learning, mitigating the problem of catastrophic forgetting --- where models lose previously acquired knowledge upon learning new tasks~\cite{DeLange2022catastrophic}. In IR, Park et al.~\cite{park2024colora} applied low-rank decomposition within single-task IR frameworks to enhance model performance. 
Conversely, we propose a simple approach to merge different adaptors to obtain a flexible blind all-in-one model capable of learning new tasks without forgetting previously learned IR tasks. This is achieved through a disentangled scheme that requires minimal retraining for each new task.

\section{Method}
Figure~\ref{fig:method} presents an overview of our proposed approach for enhancing low-quality images by systematically addressing the key limitations in current IR methods. Our approach consists of three phases, roughly targeting the following challenges: (i) robust generalization to varying types of degradations (Phase I), (ii) effective adaptation to specific degradations (Phase II), (iii) flexible blind all-in-one mechanism able to handle unknown and mixed degradations (Phase III).
Note that we freeze the parameters of the baseline after Phase I and adapters after Phase II, preserving maximal knowledge learned from previous phases and avoiding catastrophic forgetting. This design also allows us to adapt to new distortions with minimal training as only a new adapter and the lightweight estimator will be trained.

\vspace{-4mm}
\paragraph{Phase I: pretraining with synthetic degradations.}~Recent advances demonstrate that large-scale pretrained models can significantly improve performance across various tasks~\cite{abnar2022exploring}. Moreover, pretraining with synthetic data has also shown effectiveness when the domain gap is properly addressed~\cite{yang2024depth}. We hypothesize that applying large-scale pretraining to IR can yield a notable performance boost compared to training solely on traditional IR datasets. Thus, as depicted in Figure~\ref{fig:method}, Phase I involves training a baseline IR model with large-scale data and synthetic degradations to provide a robust weight initialization. 

To obtain large-scale data for pretraining, we define a degradation pipeline that generates a low-quality counterpart for each high-quality input image by introducing different synthetic degradations.
We pre-train on five distortions: noise, blur, rain, haze, and low-light conditions, as: (i) they are the most studied and commonly observed in real-world scenarios, (ii) they provide a diverse and balanced mix of low- and high-frequency degradations, and (iii) they enable a strong foundation for other distortions. Each synthetic distortion closely approximates its real-world counterpart. For instance, low-light conditions are simulated by compressing the image histogram, reducing the dynamic range of pixel intensities. In contrast, haze is introduced by adding an achromatic layer based on the depth map estimated from DepthAnythingv2~\cite{yang2024depth}. Figure \ref{fig:synthetic_examples} (a)-(e) illustrate examples of each synthetic degradation. We use the Google Landmarks dataset (GLD)~\cite{weyand2020google}, a large-scale collection of five million images of diverse landmarks worldwide. To ensure that the input images are high quality, we filter out the images with a resolution smaller than 400 pixels and exclude images with a NIMA score~\cite{talebi2018nima} below the 4.90 threshold.

To construct our baseline model, we build on recent advancements in all-in-one IR methods. Particularly, we use the Restormer architecture~\cite{zamir2022restormer} with spatial attention mechanisms~\cite{chen2023x-restormer} and incorporate a modified version of PromptIR~\cite{potlapalli2024promptir} prompt blocks. This combination enables the baseline model to effectively capture the inherent information of various degradations, establishing a strong foundation for generalized IR tasks.

While training on single distortions allows the model to handle each type individually, real-world IR often involves mixed degradations. However, applying multiple synthetic degradations simultaneously can severely deteriorate image quality, making it harder for the model to reconstruct the original image (see Table \ref{tab:synthetic_ablation}). To address this, we implement two strategies: (i) a Degradation CutMix technique, and (ii) a segmentation head with a cross-entropy loss. In Figure \ref{fig:synthetic_examples}~(f), we show our first strategy --- inspired by Yang et al.~\cite{yang2024depth} for depth estimation, where two different degradations are applied in separated random regions. Although this setup is not fully realistic, it helps the model distinguish and manage multiple degradation types within the same image. Our second strategy incorporates a segmentation head into the baseline model, producing a per-pixel distortion map. By comparing this map with the ground truth map, we guide the model to recognize and differentiate multiple types of degradation. The segmentation head is removed in subsequent phases. Additional details on our baseline and synthetic degradation generation are provided in the Supplementary Material.

\begin{table*}[ht!]
    \centering
    \caption{5-degradations setup. Quantitative results on five IR datasets comparing the state-of-the-art all-in-one methods and our approach.}    
    \small
    \vspace{-2pt}
    \setlength{\tabcolsep}{5.3pt} 
    \begin{tabular}{lccccccccccccc}
        
        \toprule
        \multirow{2}{*}{PSNR/SSIM} & \multicolumn{2}{c}{Deraining} & \multicolumn{2}{c}{Dehazing} & \multicolumn{2}{c}{Denoising} & \multicolumn{2}{c}{Deblurring} & \multicolumn{2}{c}{Low-Light} & \multicolumn{2}{c}{\multirow{2}{*}{Average}} & \multirow{2}{*}{Param.}\\
        
        \cmidrule{2-11}

        & \multicolumn{2}{c}{Rain100L} & \multicolumn{2}{c}{SOTS (Out)} & \multicolumn{2}{c}{BSD68 $_{\sigma=25}$} & \multicolumn{2}{c}{GoPro} & \multicolumn{2}{c}{LoLv1} & & & \\
        \midrule
    
        AirNet~\cite{li2022airnet} & 32.98 & .951 & 21.04 & .884 & 30.91 & .882 & 24.35 & .781 & 18.18 & .735 & 25.49 & .847 & 9M\\
        Uformer~\cite{wang2022uformer} & 35.48 & .967 & 27.20 & .958 & 30.59 & .869 & 26.41 & .809 & 21.40 & .808 & 28.21 & .882 & 52M \\
        IDR~\cite{zhang2023idr} & 35.63 & .965 & 25.24 & .943 & \best{31.60} & .887 & 27.87 & .846 & 21.34 & .826 & 28.34 & .893 & 15M \\
        X-Restormer~\cite{chen2023x-restormer} & 35.42 & .968 & 27.58 & .959 & 30.92 & .880 & 27.54 & .835 & 20.88 & .817 & 28.47 & .891 & 26M \\
        DA-CLIP~\cite{luo2023daclip} & 35.49 & .970 & 28.10 & .962 & 30.42 & .859 & 26.50 & .807 & 21.94 & .817 & 28.49 & .880 & 174M \\ 
        DiffUIR~\cite{zheng2024selective} & 35.52 & .969 & 28.17 & .964 & 30.92 & .879 & 26.99 & .821 & 20.92 & .789 & 28.50 & .880 & 36M \\
        Restormer~\cite{zamir2022restormer} & 35.56 & .970 & 27.94 & .962 & 30.74 & .875 & 26.84 & .818 & 21.74 & .815 & 28.56 & .888 & 26M \\
        PromptIR~\cite{potlapalli2024promptir} & 35.40 & .967 & 28.26 & .965 & 30.89 & .872 & 26.55 & .808 & 21.80 & .815 & 28.58 & .885 & 36M \\
        AdaIR~\cite{cui2025adair} & 38.02 & .981 & 30.53 & .978 & 31.35 & .889 & 28.12 & .858 & 23.00 & .845 & 30.20 & .910 & 28M \\ 
        \midrule
        Ours OH & \second{38.18} & \second{.983} & \second{33.46} & \second{.983} & \second{31.38} & \best{.898} & \best{29.00} & \best{.878} & \best{24.20} & \best{.865} & \second{31.24} & \second{.921} & 41M\\
        Ours SW & \best{38.22} & \best{.984} & \best{33.48} & \best{.984} & \second{31.38} & \best{.898} & \best{29.00} & \best{.878} & \second{24.19} & \best{.865} & \best{31.25} & \best{.921} & 41M\\
        \bottomrule
        
    \end{tabular}
    \vspace{-3mm}
    \label{tab:5tasks}
\end{table*}

\paragraph{Phase II: Single-task adaptation.}~While synthetic degradations provide a robust weight initialization for our baseline model, a domain shift remains between the pretraining data (due to the different dataset and synthetic degradations), standard IR datasets, and real-world conditions. For instance, accurately simulating haze is particularly challenging, as it depends on light scattering by particles, which varies with depth and atmospheric conditions. To mitigate this gap, we introduce Phase II of our approach, where the baseline model is adapted using task-specific adapters specializing on a single degradation by using a standard IR dataset, as outlined in Figure~\ref{fig:method}.

In this phase, we augment each linear and convolutional layer with a set of LoRA adapters, one per degradation type and parametrized by $\left\{A_n\right\}$ and $\left\{B_n\right\}$ where $n$ denotes the degradation index. These low-rank matrices refine the initial frozen weights through linear combinations and are trained individually for each degradation type. For simplicity, we omit the layer-specific indices, but each layer has its own set of adapters. By leveraging pretraining and subsequent specialization via adapters, our approach achieves superior performance across diverse tasks and exhibits improved generalization.

\paragraph{Phase III: Multi-task integration.}~LoRA adapters serve as a plug-and-play solution to refine our pretrained model. However, we still need to select the most suitable adapter based on the input image, when the degradation type is unknown. To address this limitation and derive a blind all-in-one IR method, we propose to use a lightweight degradation estimator $p\left(n\vert x;\theta\right)$ to estimate the probability of each degradation $n$ given the input image $x$, parametrized by $\theta$. This estimator is trained to identify the type of degradation in the image using the combination of all the IR datasets in Phase II. As shown in Figure~\ref{fig:method}, the probabilities of the estimator are used to linearly combine the task-specific adapters with the baseline weights. In this way, the baseline is adapted with the task-specific LoRAs according to the lightweight estimator and, therefore, each input image individually. Specifically, given an input image $x$, the weight update for a specific layer with baseline weights $W$ and adapter weights $\left\{A_n\right\}$ and $\left\{B_n\right\}$ is computed as:
\begin{equation} \label{eq:phase3}
    W'\left(x\right) = W + \sum_{n=1}^{N} p\left(n\vert x;\theta\right) B_n A_n,
\end{equation}
where $N$ is the number of seen degradations. 

We propose two variants of the estimator: (i) one-hot (OH) and (ii) soft-weights (SW). The one-hot variant simply selects the adapter corresponding to the degradation with the maximum probability, while the soft-weight variant computes the weighted average as in Eq.~\ref{eq:phase3}.

With a strong baseline model capable of generalizing effectively, our approach not only achieves superior performance on known degradations but also performs well on unseen and mixed degradations. Moreover, thanks to its modular architecture, we empirically demonstrate that performance on unseen degradation types can be further improved with minimal training effort. Specifically, one only needs to train an additional adapter for the new degradation type and retrain the lightweight estimator --- together requiring fewer than four million parameters.


\begin{table*}[t]
    \centering
    \caption{\small 3-degradations setup. Quantitative results on three IR datasets comparing the state-of-the-art all-in-one methods and our approach.}
    \vspace{-1mm}
    \setlength{\tabcolsep}{5.5pt}
    \small
    \begin{tabular}{lcccccccccccc}
        \toprule
        \multirow{2}{*}{PSNR/SSIM} & \multicolumn{2}{c}{Deraining} & \multicolumn{2}{c}{Dehazing} & \multicolumn{6}{c}{Denoising} & \multicolumn{2}{c}{\multirow{2}{*}{Average}}\\
        
        \cmidrule{2-11}

        & \multicolumn{2}{c}{Rain100L} & \multicolumn{2}{c}{SOTS (Out)} & \multicolumn{2}{c}{BSD68 $_{\sigma=15}$} & \multicolumn{2}{c}{BSD68 $_{\sigma=25}$} & \multicolumn{2}{c}{BSD68 $_{\sigma=50}$} & & \\
        
        \midrule

        DL~\cite{fan2019general}  & 32.62 & .931 & 26.92 & .931 & 33.05 & .914 & 30.41 & .861 & 26.90 & .740 & 29.98 & .875 \\
        MPRNet~\cite{zamir2021multi} & 33.57 & .954 & 25.28 & .954 & 33.54 & .927 & 30.89 & .880 & 27.56 & .779 & 30.17 & .899 \\
        AirNet~\cite{li2022airnet} & 34.90 & .967 & 27.94 & .962 & 33.92 & .933 & 31.26 & .888 & 28.00 & .797 & 31.20 & .909 \\
        Restormer~\cite{zamir2022restormer}  & 35.56 & .969 & 29.92 & .970 & 33.86 & .933 & 31.20 & .888 & 27.90 & .794 & 31.69 & .911 \\
        PromptIR~\cite{potlapalli2024promptir} & 36.37 & .972 & 30.58 & .974 & 33.98 & .933 & 31.31 & .888 & 28.06 & .799 & 32.06 & .913 \\
        AdaIR~\cite{cui2025adair} & \second{38.64} & \best{.983} & 31.06 & .980 & 34.12 & \best{.935} & \best{31.45} & \best{.892} & 28.19 & .802 & 32.69 & .918 \\
        
        \midrule
        Ours OH & \best{38.69} & \second{.982} & \best{33.53} & \best{.984} & \best{34.18} & \best{.935} & \second{31.38} & \second{.890} & \best{28.25} & \best{.804} & \best{33.21} & \best{.919} \\
        Ours SW & \second{38.64} & \second{.982} & \second{33.50} & \best{.984} & \best{34.18} & \best{.935} & \second{31.38} & \second{.890} & \second{28.22} & \second{.803} & \second{33.18} & \best{.919} \\
        \bottomrule
        \end{tabular}
    \label{tab:3tasks}
\end{table*}

\begin{table}[t!]
    \centering
    \caption{\small Quantitative results on additional test datasets with the learned degradations.}
    \vspace{-1mm}
    \setlength{\tabcolsep}{3.2pt} 
    \small
    \begin{tabular}{lcccccc}
    \toprule
    \multirow{2}{*}{PSNR/SSIM} & \multicolumn{2}{c}{Deraining} & \multicolumn{2}{c}{Deblurring} & \multicolumn{2}{c}{Low-Light} \\
    \cmidrule{2-7}
    & \multicolumn{2}{c}{Rain100H} & \multicolumn{2}{c}{HIDE} & \multicolumn{2}{c}{Lolv2-Real} \\
    \midrule 
    IDR~\cite{zhang2023idr} & 11.32 & .397 & 16.83 & .621 & 17.61 & .697 \\
    X-Restormer~\cite{chen2023x-restormer} & 14.08 & .437 & 25.40 & .801 & 25.42 & .876  \\
    DiffUIR~\cite{zheng2024selective} & 14.78 & .487 & 23.98 & .739 & 26.12 & .861  \\
    Restormer~\cite{zamir2022restormer} & 14.50 & .464 & 24.42 & .781 & 26.12 & .877 \\
    AdaIR~\cite{cui2025adair} & 14.13 & .438 & 24.43 & .771 & 26.54 & .867 \\
    PromptIR~\cite{potlapalli2024promptir} & 14.28 & .444 & 24.49 & .762 & 26.70 & .870 \\

    \midrule
    
    Ours OH & \best{18.76} & \best{.612} & \second{27.04} & \best{.850} & \best{28.09} & \best{.907} \\
    Ours SW & \second{17.13} & \second{.564} & \best{27.05} & \best{.850} & \best{28.09} & \second{.906} \\
    \bottomrule
    \end{tabular}
    \label{tab:unseen_datasets}
\end{table}

\section{Experiments}\label{sec:experiments}
We evaluate our method on two all-in-one IR setups: five- and three-degradation. Additionally, we assess its performance on datasets excluded from training, unseen IR degradation types, and mixed degradation scenarios. The accuracy of the methods is measured using two well-established metrics: PSNR and SSIM. In all cases we report the mean value across all the test images, and we highlight \colorbox{myred}{\textbf{best}} and \colorbox{myblue}{\underline{second-best}} values for each metric. Our approach is compared against recent all-in-one IR methods. However, some existing methods have been evaluated under different setups or lack publicly available training code and models~\cite{jiang2024autodir, ai2024multimodal, wang2023painter}. To ensure a fair comparison, we train five state-of-the-art methods, Restormer~\cite{zamir2022restormer}, DA-CLIP~\cite{luo2023daclip}, DiffUIR~\cite{luo2023controlling}, PromptIR~\cite{potlapalli2024promptir}, X-Restormer~\cite{chen2023x-restormer}, on the five-degradation IR setup. We used the official code provided by the authors. For our approach, we evaluate the two variations, one-hot (OH) and soft weights (SW).

\vspace{-2mm}
\paragraph{Implementation details:}~Our training follows a three-phase scheme, using the Adam~\cite{kingma2014adam} optimizer with weight decay~\cite{loshchilov2017decoupled}, an initial learning rate of $2 \times 10^{-4}$, a cosine learning rate scheduler, and a warmup start of one epoch. Following Potlapalli et al.~\cite{potlapalli2024promptir}, we train our model on four NVIDIA A40 GPUs with a batch size of 4 and image patches of 384.
In the Phase I, we jointly optimize a weighted combination of two objectives: a cross-entropy loss $L_{CE}$ for per-pixel degradation classification and a reconstruction loss $L_R$, which consists of $L_1$ loss and SSIM loss. The cross-entropy and SSIM components are both weighted at 0.5 to balance classification and reconstruction. During Phase II, we optimize only the reconstruction loss $L_R$, maintaining the same SSIM weighting as in the previous phase. In Phase III, we train the lightweight estimator separately using a cross-entropy loss to classify the degradation type of the input image, ensuring that the model accurately learns degradation prediction.

\begin{table}[t!]
    \centering
    \caption{\small Quantitative results for unseen IR tasks. Note that the models have not been trained for these degradations. Ours* shows results for the lightweight re-training scenario. New adapters are trained for the new tasks and the estimator is retrained with 8 tasks.}
    \vspace{-1mm}
    \setlength{\tabcolsep}{3.5pt} 
    \small
    \begin{tabular}{lcccccc}
    \toprule
    \multirow{2}{*}{PSNR/SSIM} & \multicolumn{2}{c}{4-to-8 bits} & \multicolumn{2}{c}{JPEG Q20} & \multicolumn{2}{c}{Desnowing} \\
    \cmidrule{2-7}
    & \multicolumn{2}{c}{Live1} & \multicolumn{2}{c}{Live1} & \multicolumn{2}{c}{City-Snow} \\
    \midrule
    IDR~\cite{zhang2023idr} & 24.02 & .738 & 26.51 & .913 & 18.00 & .649\\
    X-Restormer~\cite{chen2023x-restormer} & 24.73         & .745                        & 26.86                    & .922                          & 18.51         & .681               \\
    DiffUIR~\cite{zheng2024selective} & 24.68         & .743                        & 26.88                    & .921                          & 18.39         & .671               \\
    Restormer~\cite{zamir2022restormer} & 24.64         & .743                        & 26.90                     & .929                          & 18.14         & .655               \\
    AdaIR~\cite{cui2025adair} & 24.63 & .739 & 26.65 & .924 & 18.06 & .649 \\
    PromptIR~\cite{potlapalli2024promptir} & 24.70         & .740                         & 26.60                     & .920                           & 18.49         & .673               \\
\midrule

Ours OH & \second{25.25}         & \second{.742}                        & \second{29.20}                     & \second{.931}                          & \best{18.71}             & \best{.684}                \\
Ours SW & \best{25.32}         & \best{.743}                        & \best{29.35}                    & \best{.926}                          & \second{18.67}         & \second{.683}               \\
\midrule
Ours OH$^*$ & 29.14 & .826 & 30.82 & .943 & 24.19 & .797 \\
Ours SW$^*$ & 29.03 & .810 & 30.71 & .939 & 24.02 & .779 \\

    \bottomrule
    \end{tabular}
    \label{tab:unseen_tasks}
\end{table}

\paragraph{5-Degradation blind IR.}~For the blind five-task setup, we follow the protocol of Zhang et al.~\cite{zhang2023idr}. Specifically, for training we use Rain200L\cite{yang2017deep} for deraining, RESIDE~\cite{li2018benchmarking} for dehazing, BSD400~\cite{martin2001database} and WED~\cite{ma2016waterloo} for denoising with $\sigma{=}25$, GoPro~\cite{Nah2017gopro} for deblurring, and LOL~\cite{wei2018loldataset} for low-light enhancement. Evaluation is conducted on Rain100L~\cite{yang2017deep}, SOTS-Outdoor~\cite{li2018benchmarking}, BSD68~\cite{martin2001database}, GoPro~\cite{Nah2017gopro}, and LOL~\cite{wei2018loldataset}, with results reported in Table~\ref{tab:5tasks}. We outperform all the methods across all tasks, except PSNR in denoising. Note that denoising gains are typically small due to the identical noise distribution in training and test sets in all-in-one setups. Our method achieves a substantial improvement over the state of the art average, with a PSNR gain of 1.05 dB. Remarkably, our method outperforms AdaIR~\cite{cui2025adair} by 2.95 dB PSNR and 1.19 dB PSNR on dehazing and low-light image enhancement, respectively.

\begin{figure*}[t] \centering
    \raisebox{0.1\height}{\makebox[0.02\textwidth]{\rotatebox{90}{\makecell{\footnotesize GoPro~\cite{Nah2017gopro}}}}}
    \includegraphics[width=0.155\textwidth]{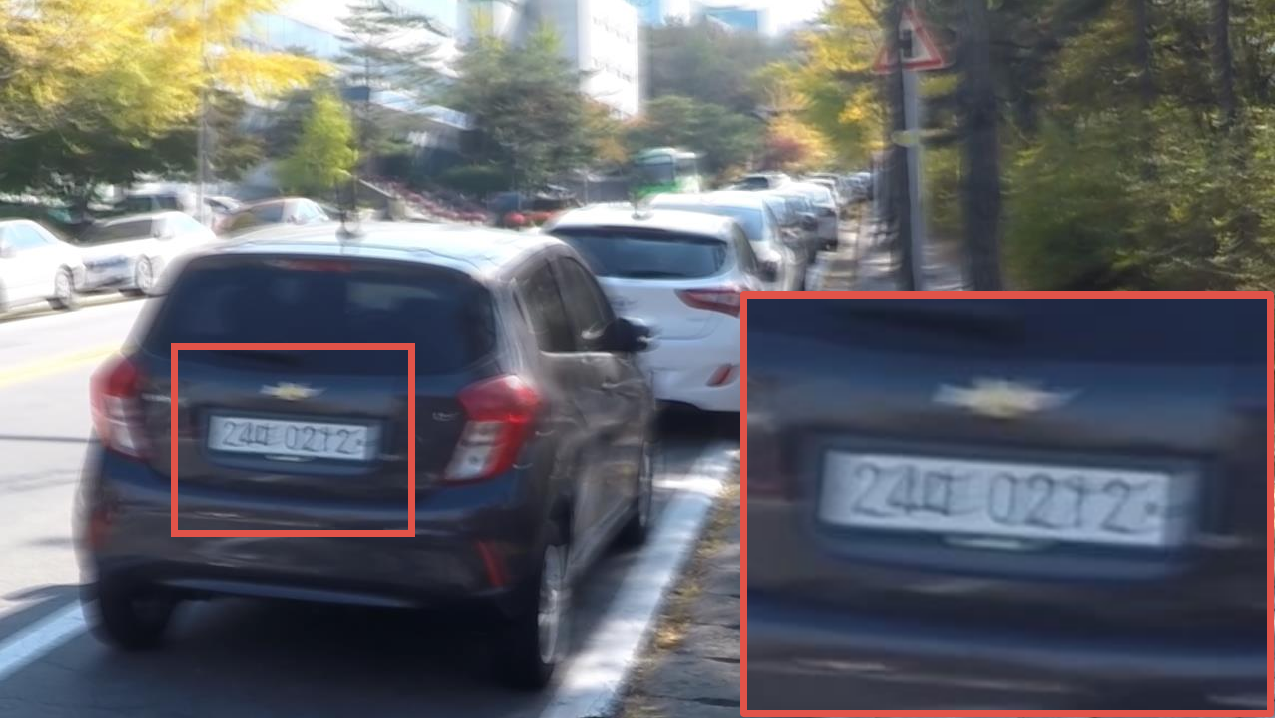}
    \includegraphics[width=0.155\textwidth]{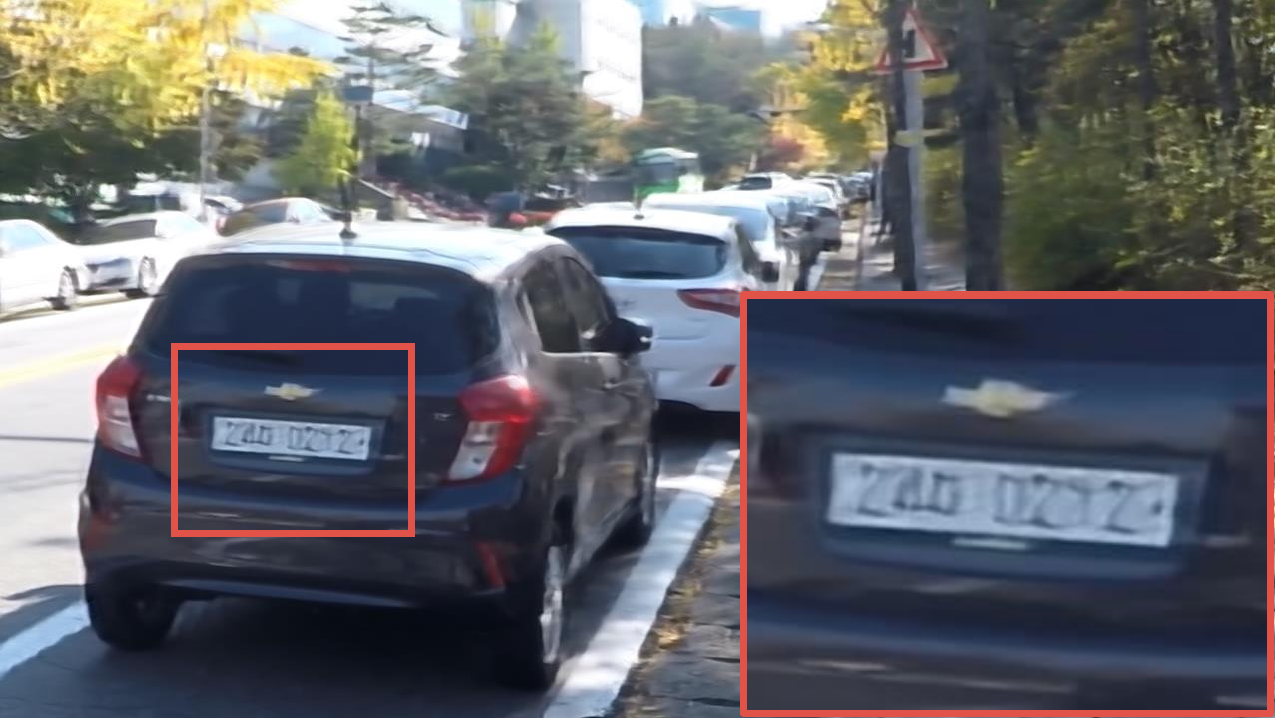}
    \includegraphics[width=0.155\textwidth]{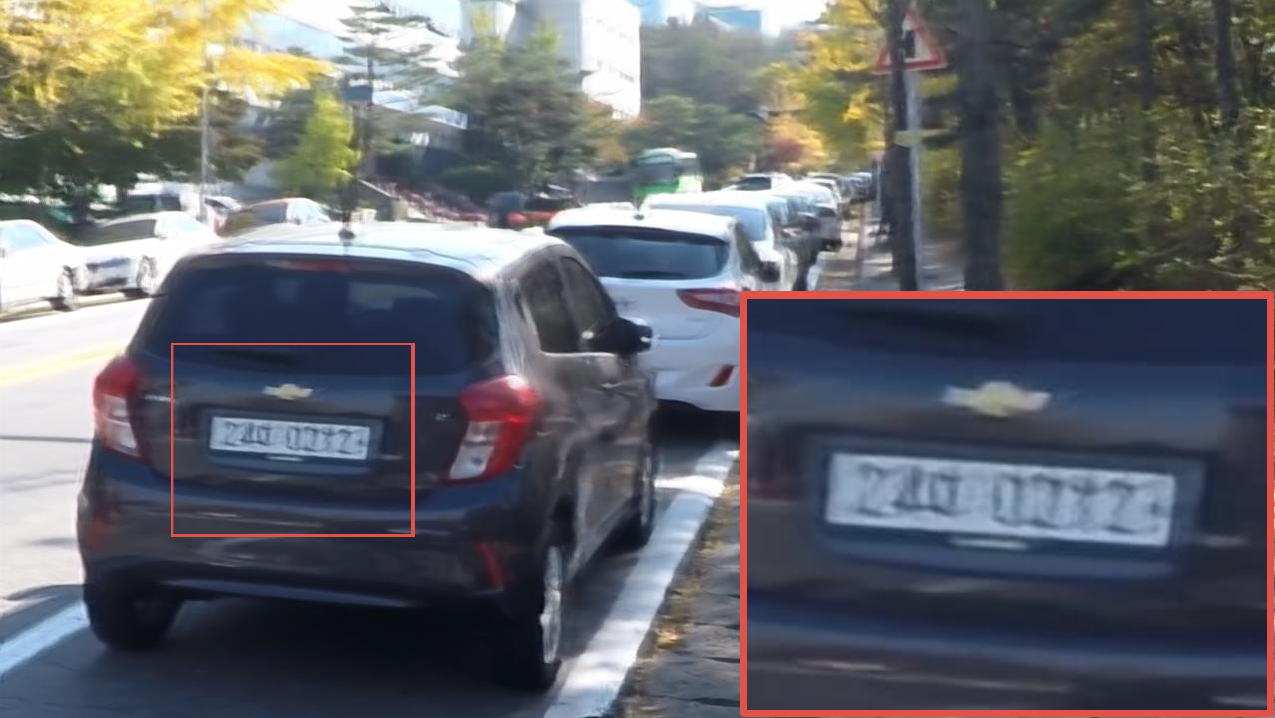}
    \includegraphics[width=0.155\textwidth]{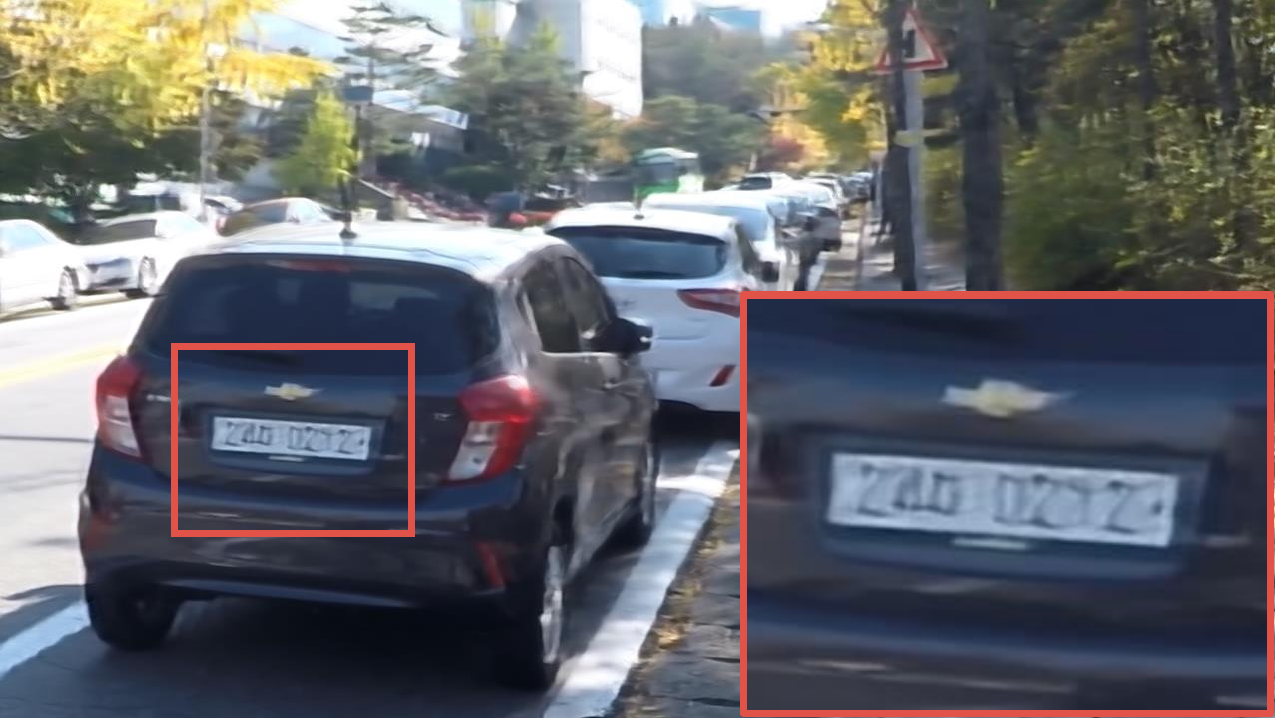}
    \includegraphics[width=0.155\textwidth]{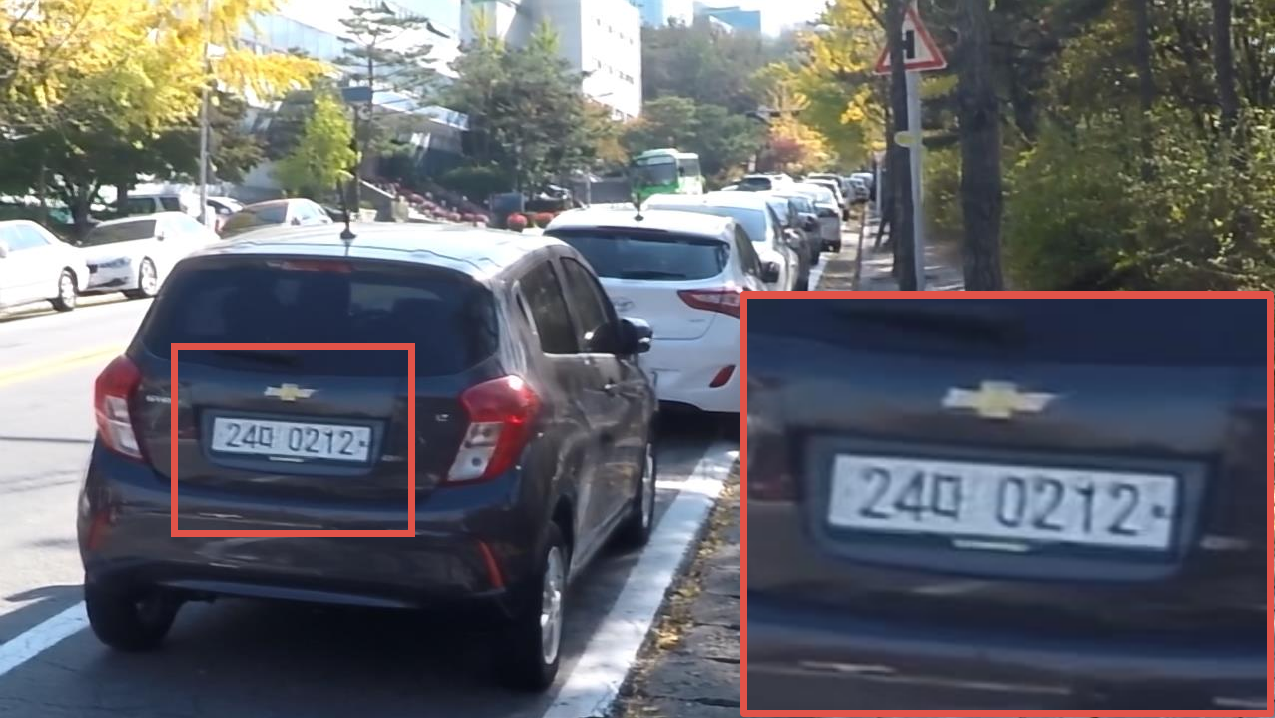}
    \includegraphics[width=0.155\textwidth]{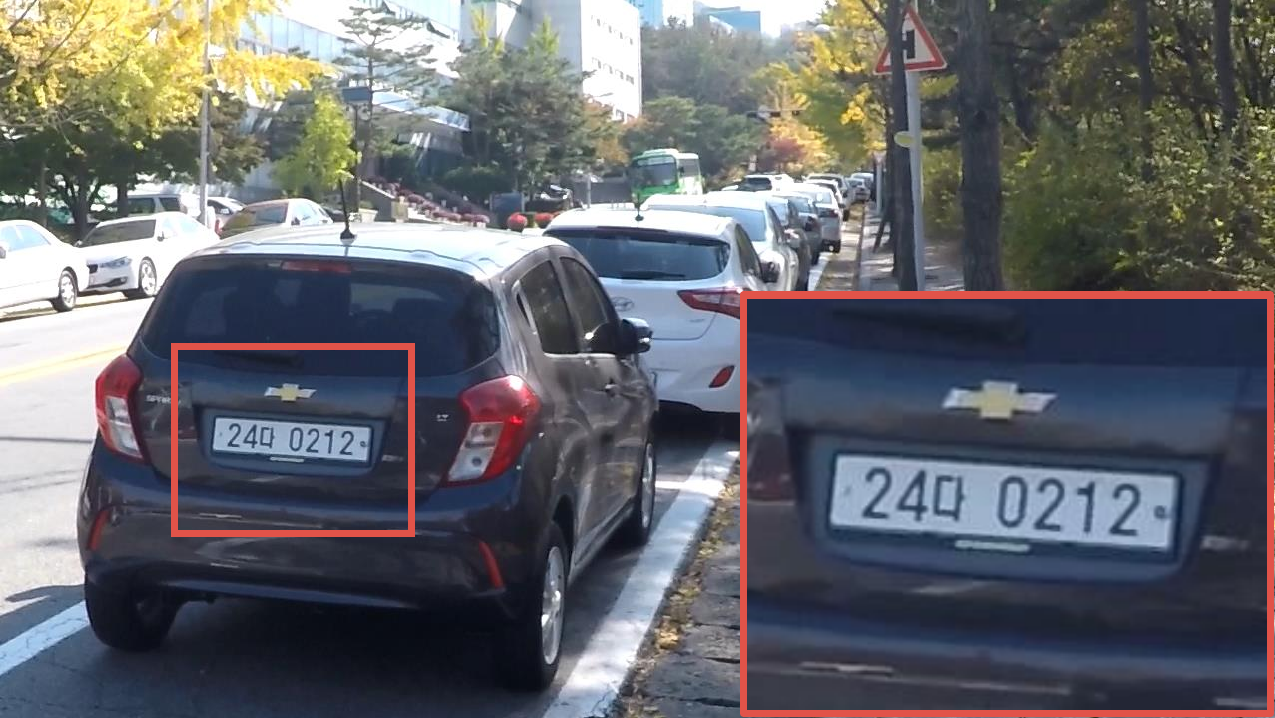}
    \\
    
    \raisebox{0.15\height}{\makebox[0.02\textwidth]{\rotatebox{90}{\makecell{\footnotesize LoLv1~\cite{wei2018loldataset}}}}}
    \includegraphics[width=0.155\textwidth]{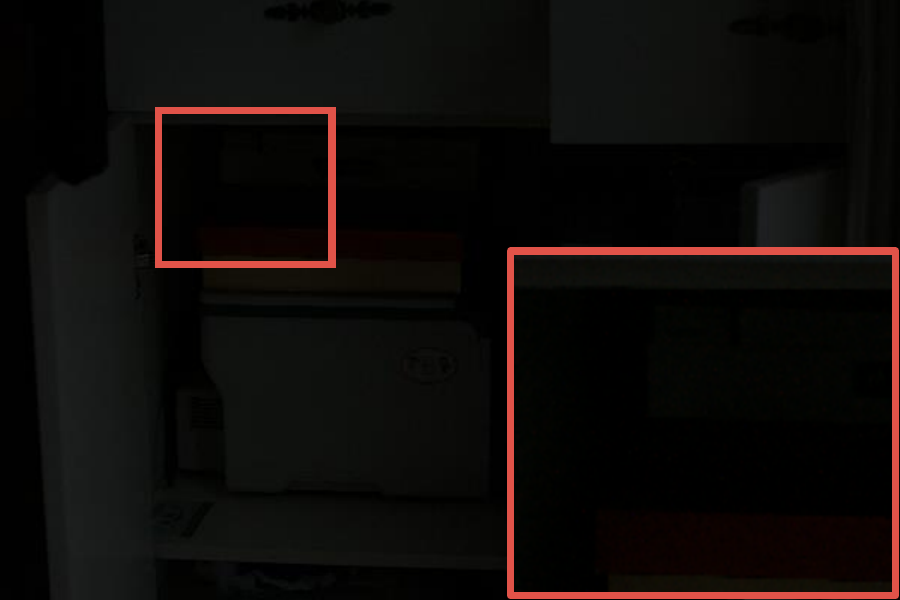}
    \includegraphics[width=0.155\textwidth]{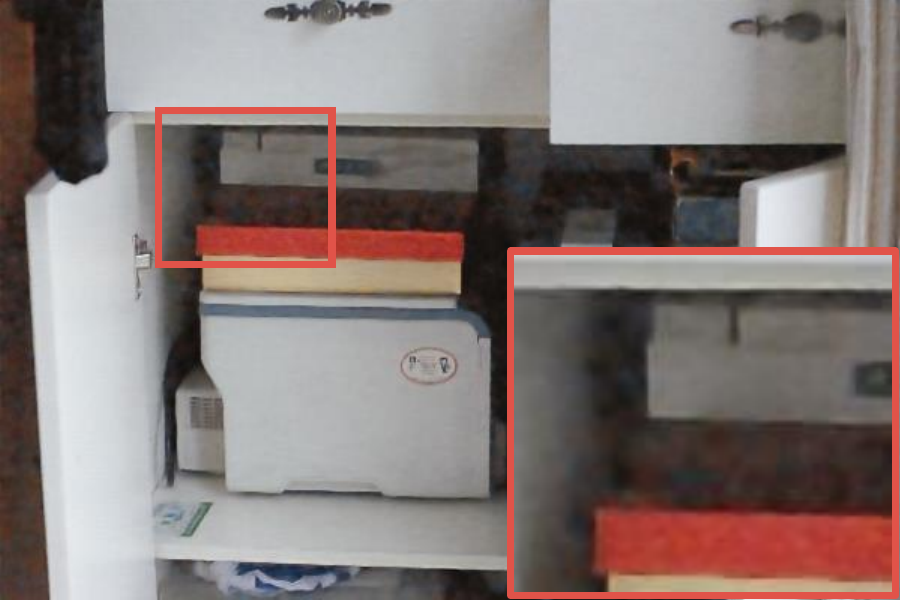}
    \includegraphics[width=0.155\textwidth]{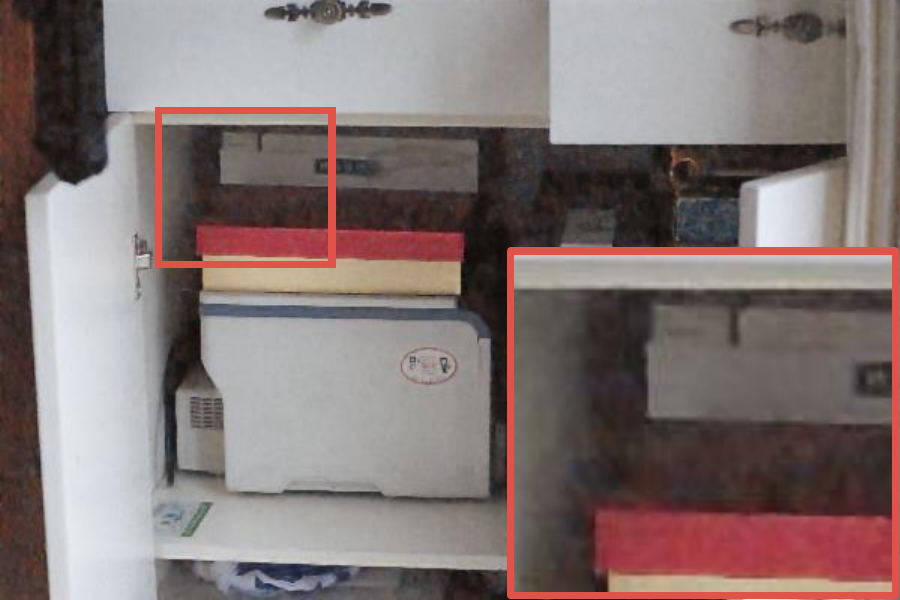}
    \includegraphics[width=0.155\textwidth]{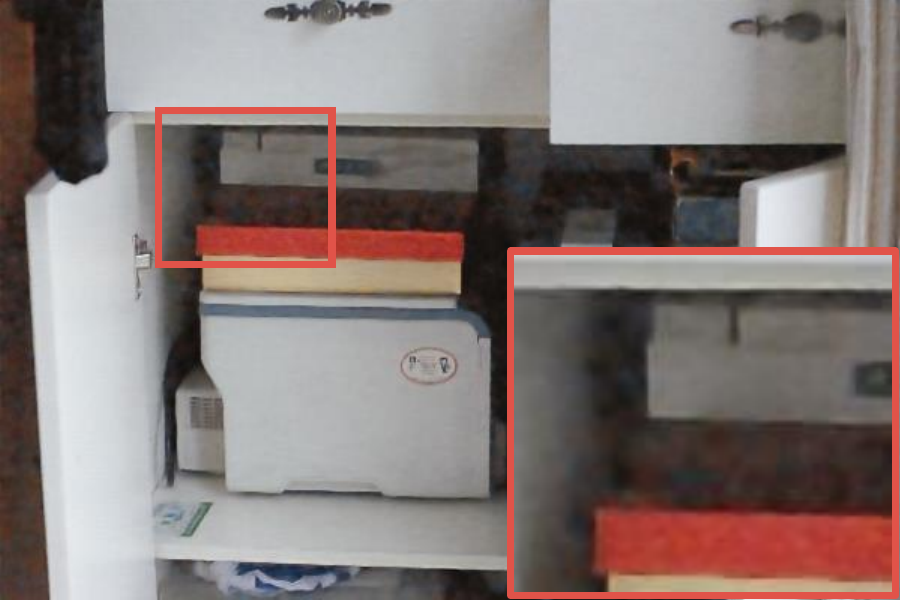}
    \includegraphics[width=0.155\textwidth]{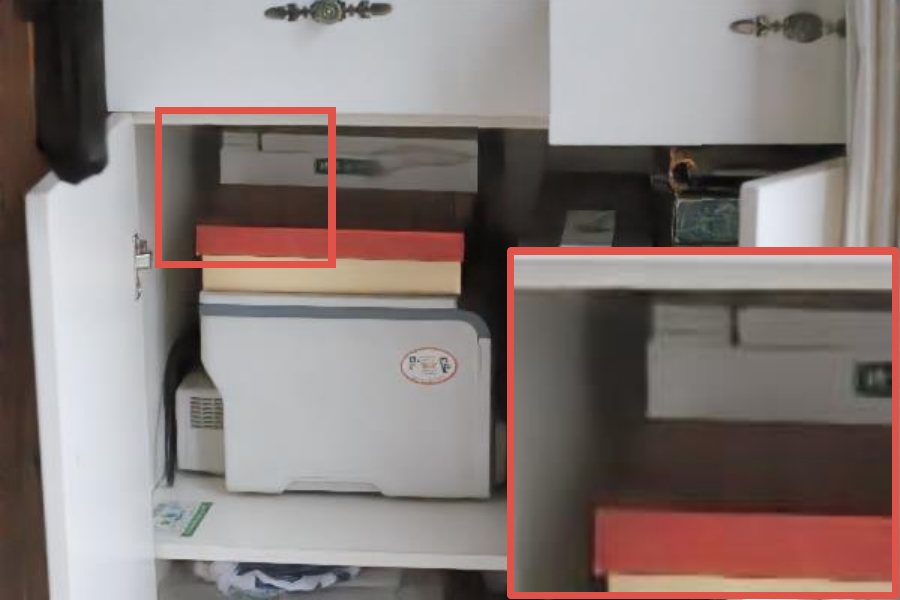}
    \includegraphics[width=0.155\textwidth]{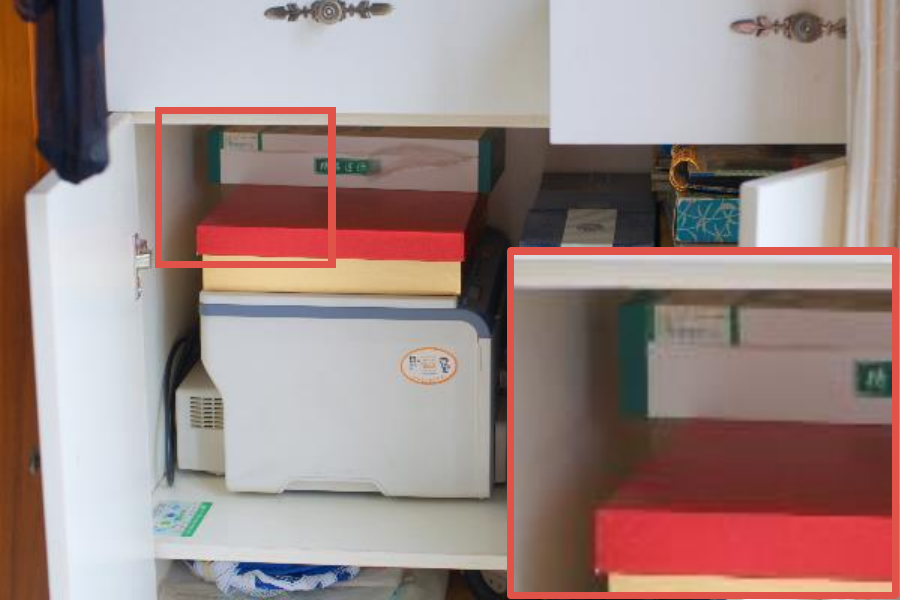}
    \\
    
    \raisebox{0.025\height}{\makebox[0.02\textwidth]{\rotatebox{90}{\makecell{\footnotesize Rain100H~\cite{yang2017deep}}}}}
    \includegraphics[width=0.155\textwidth]{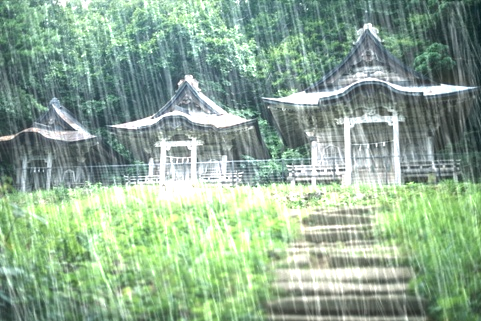}
    \includegraphics[width=0.155\textwidth]{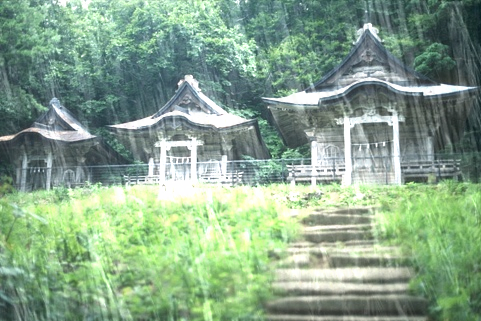}
    \includegraphics[width=0.155\textwidth]{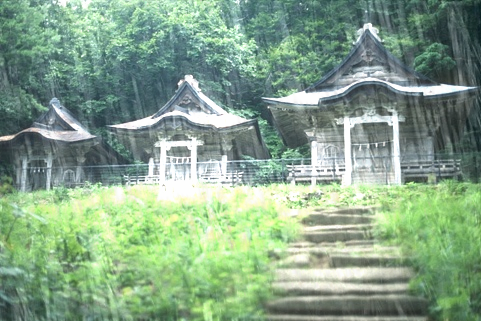}
    \includegraphics[width=0.155\textwidth]{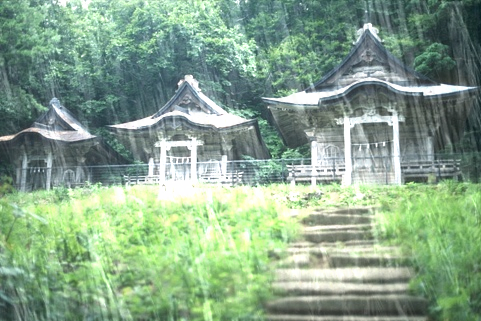}
    \includegraphics[width=0.155\textwidth]{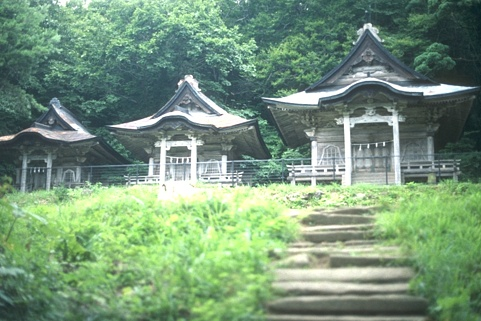}
    \includegraphics[width=0.155\textwidth]{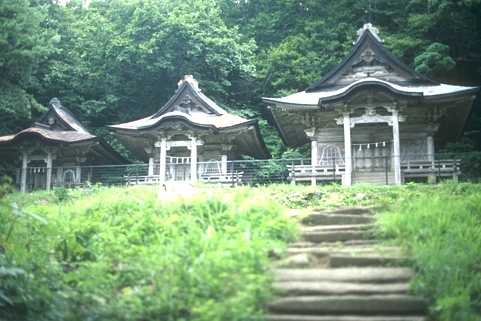}
    \\ 
    \makebox[0.02\textwidth]{}
    \makebox[0.155\textwidth]{\small Input}
    \makebox[0.155\textwidth]{\small Restormer~\cite{zamir2022restormer}}
    \makebox[0.155\textwidth]{\small PromptIR~\cite{potlapalli2024promptir}}
    \makebox[0.155\textwidth]{\small AdaIR~\cite{cui2025adair}}
    \makebox[0.155\textwidth]{\small Ours-OH}
    \makebox[0.155\textwidth]{\small Ground truth}
    \vspace{-1.5mm}
    \caption{Qualitative results for single degradation removal, including deblurring on the GoPro~\cite{Nah2017gopro} dataset, denoising on the LoLv1~\cite{wei2018loldataset} dataset, and deraining on the Rain100H~\cite{yang2017deep} dataset.} 
    \label{fig:qualitative}
\end{figure*}

\vspace{-2mm}
\paragraph{3-Degradation blind IR.} Following Li et al.~\cite{li2022airnet}, we evaluate our approach on a three-task blind IR setup, comparing it to specialized all-in-one methods for deraining, dehazing, and denoising. Compared to the five-task setup, we omit deblurring and low-light enhancement while introducing two additional noise levels: $\sigma{=}15$ and $\sigma{=}50$ on the BSD68~\cite{martin2001database} dataset. Two LoRA adapters and the lightweight estimator are trained for these new noise levels. The baseline model remains the same, trained on the five-task setup. Results in Table \ref{tab:3tasks} demonstrate that our approach outperforms all state-of-the-art methods on average while maintaining consistent performance across all degradation types. In detail, we achieve 0.52 dB average PSNR over AdaIR~\cite{cui2025adair}, even when our baseline was trained on the five-degradation setup.

\begin{figure*}[t] \centering
    \raisebox{0.3\height}{\makebox[0.02\textwidth]{\rotatebox{90}{\makecell{\footnotesize JPEG~\cite{sheikh2006statistical}}}}}
    \includegraphics[width=0.155\textwidth]{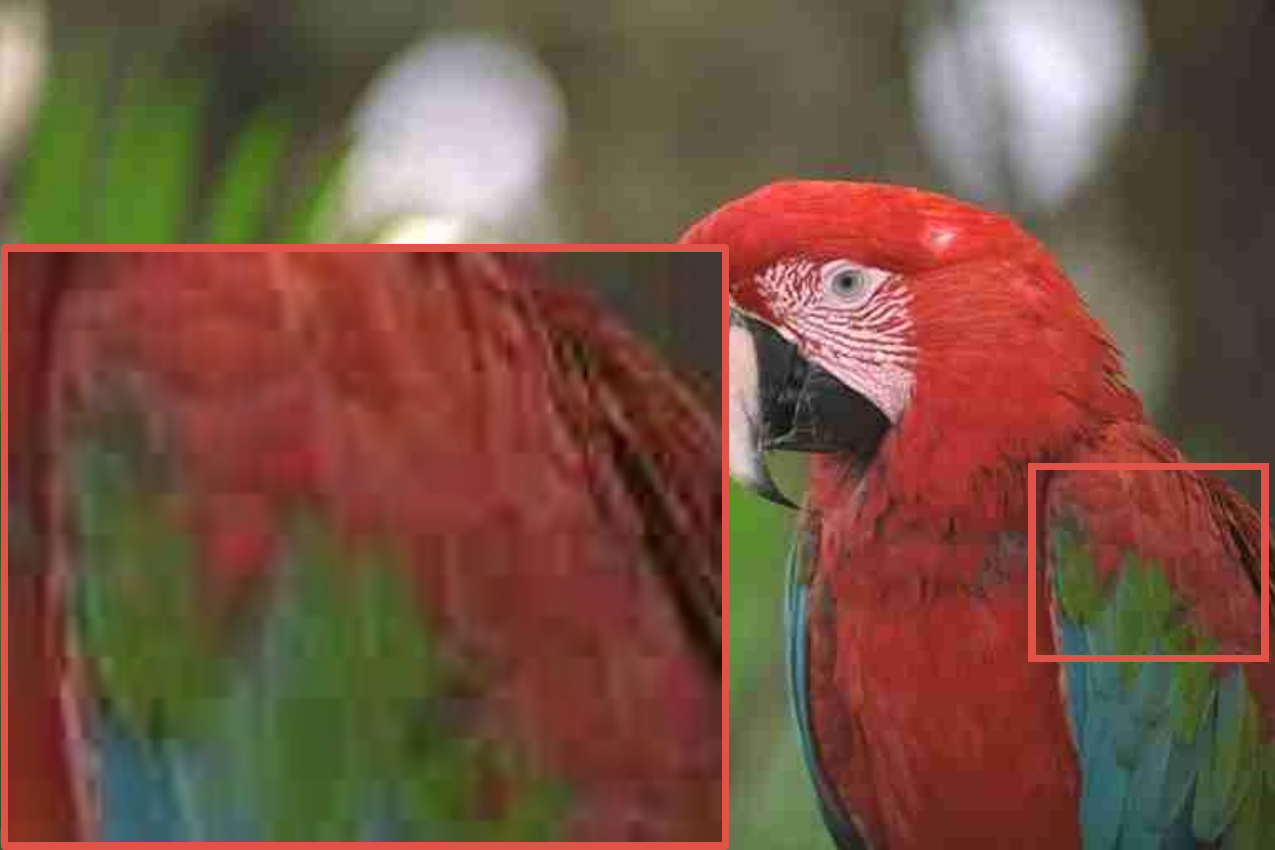}
    \includegraphics[width=0.155\textwidth]{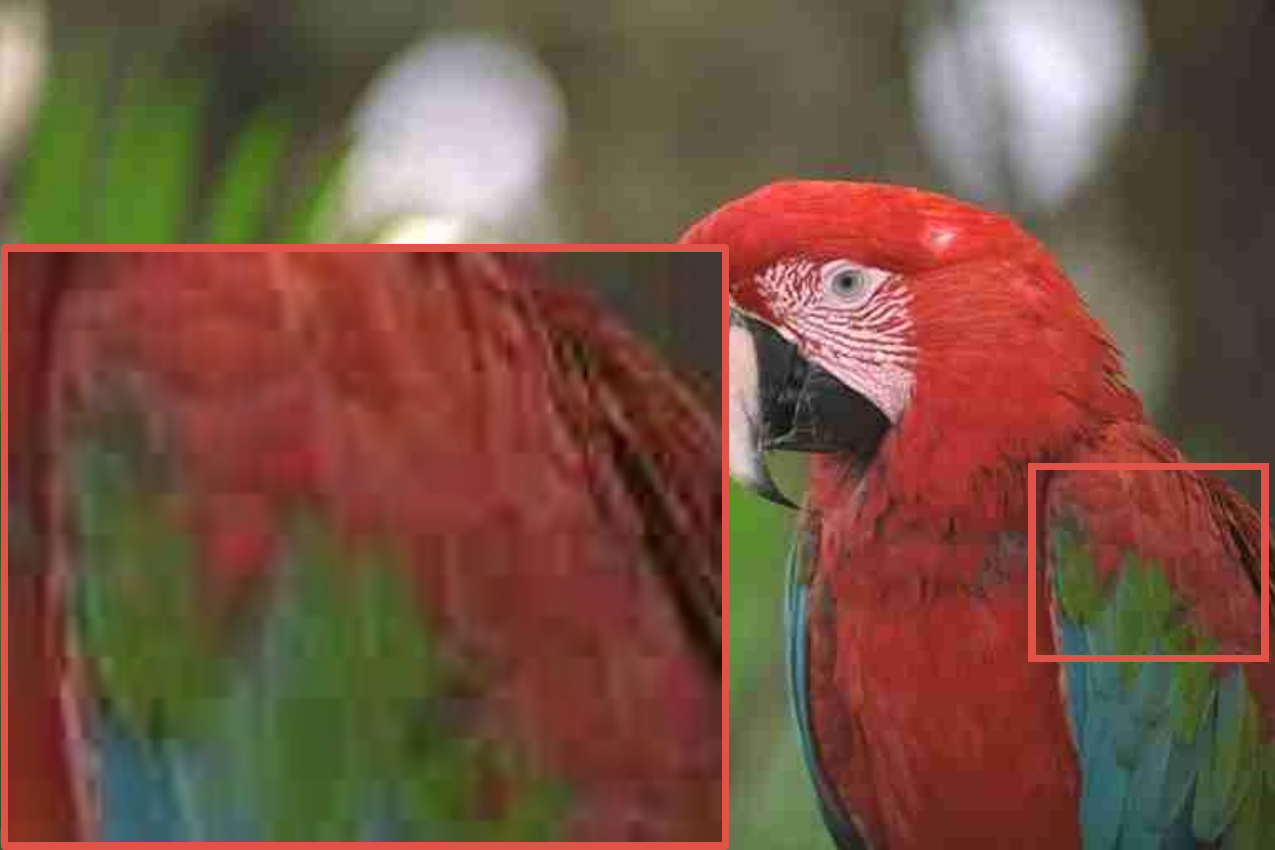}
    \includegraphics[width=0.155\textwidth]{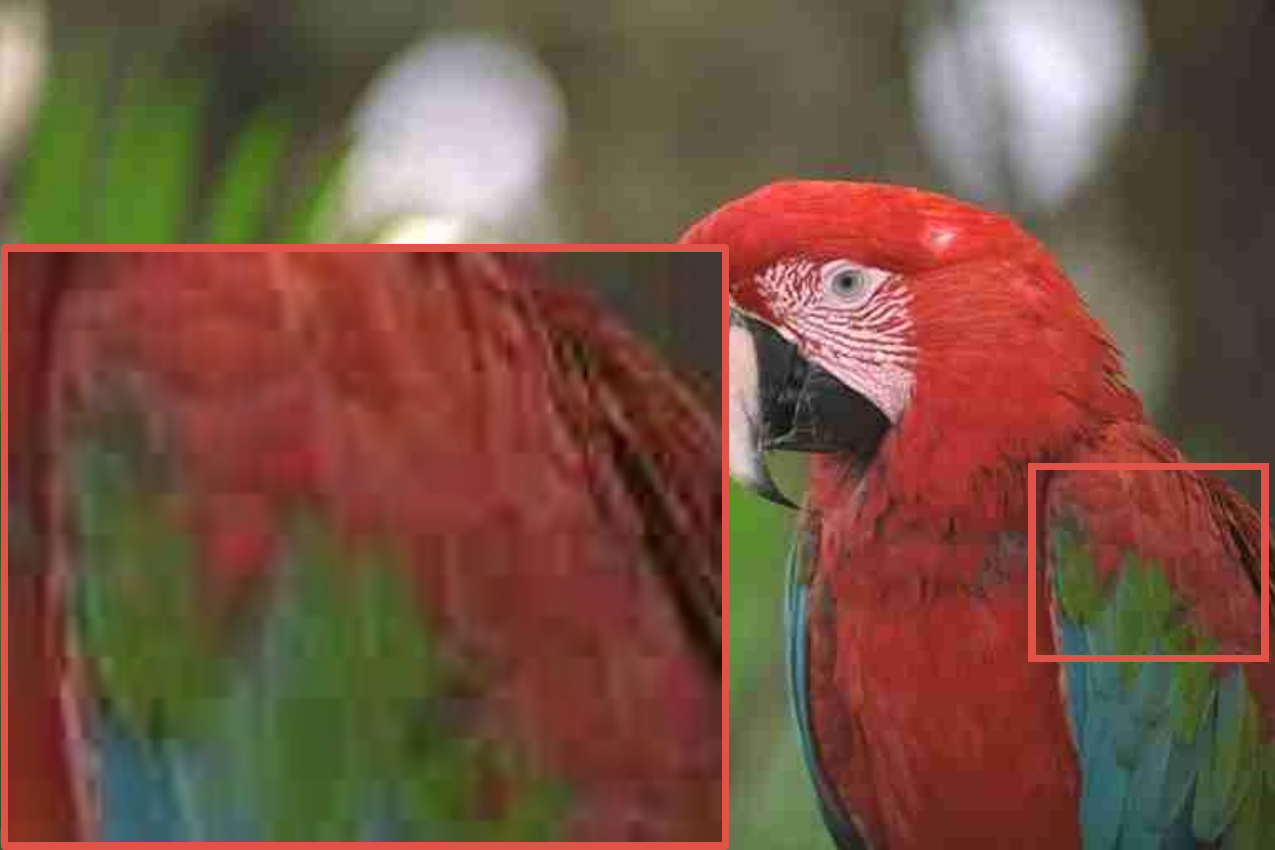}
    \includegraphics[width=0.155\textwidth]{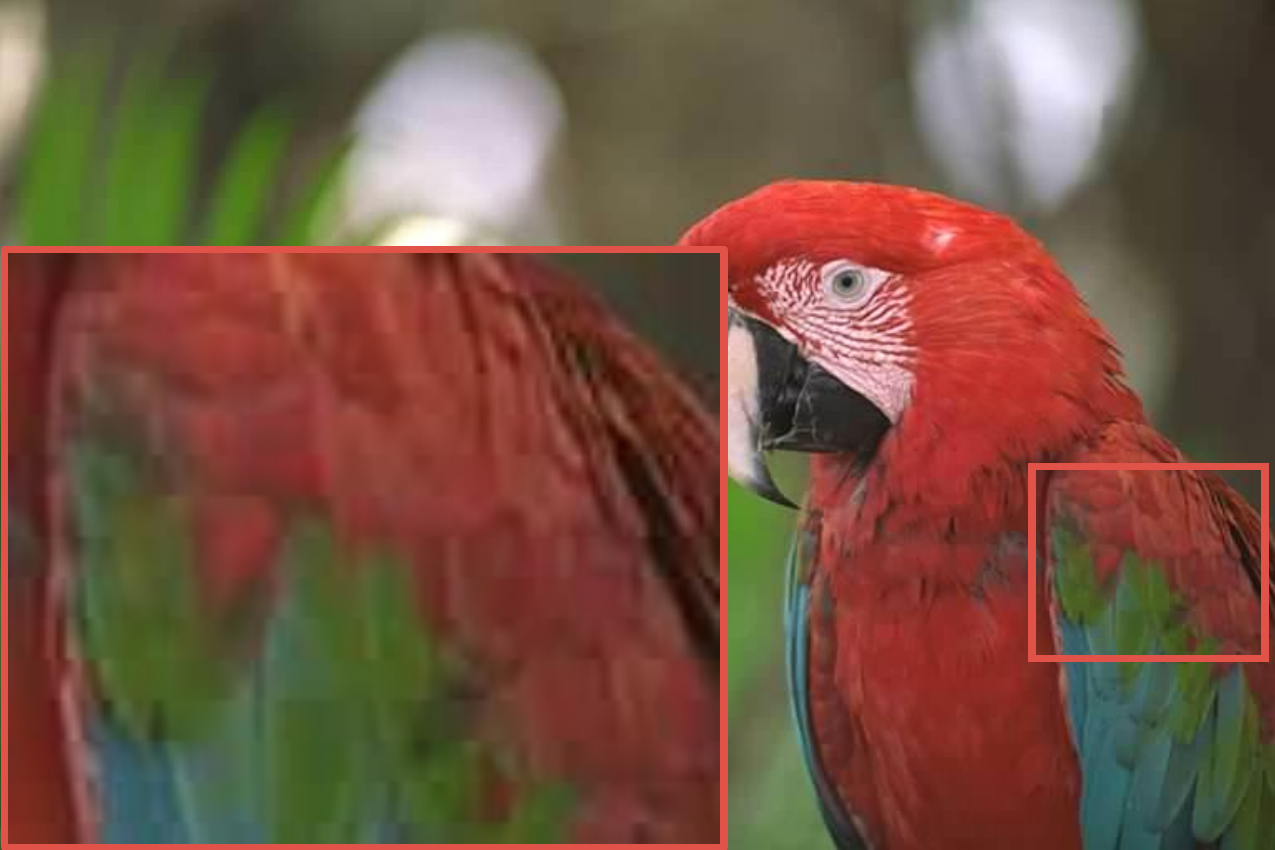}
    \includegraphics[width=0.155\textwidth]{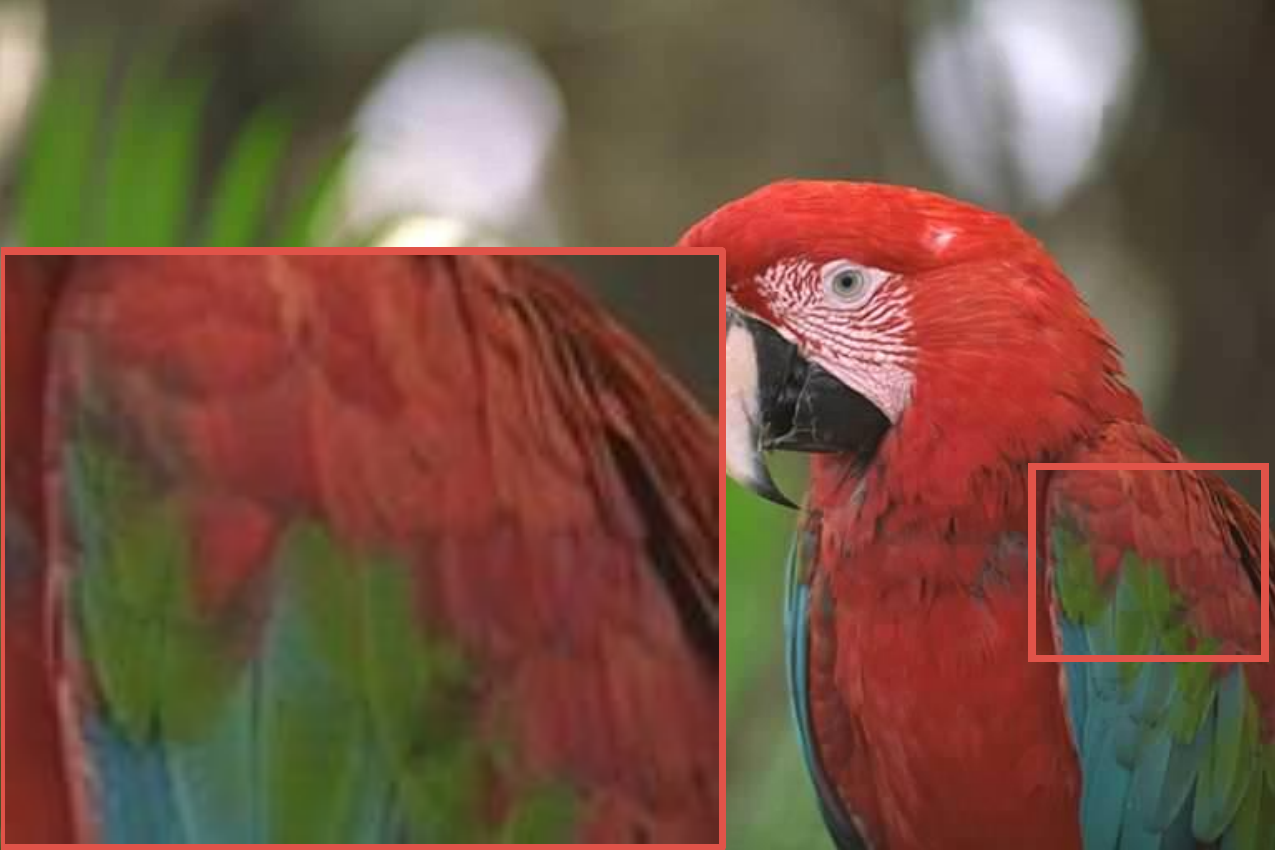}
    \includegraphics[width=0.155\textwidth]{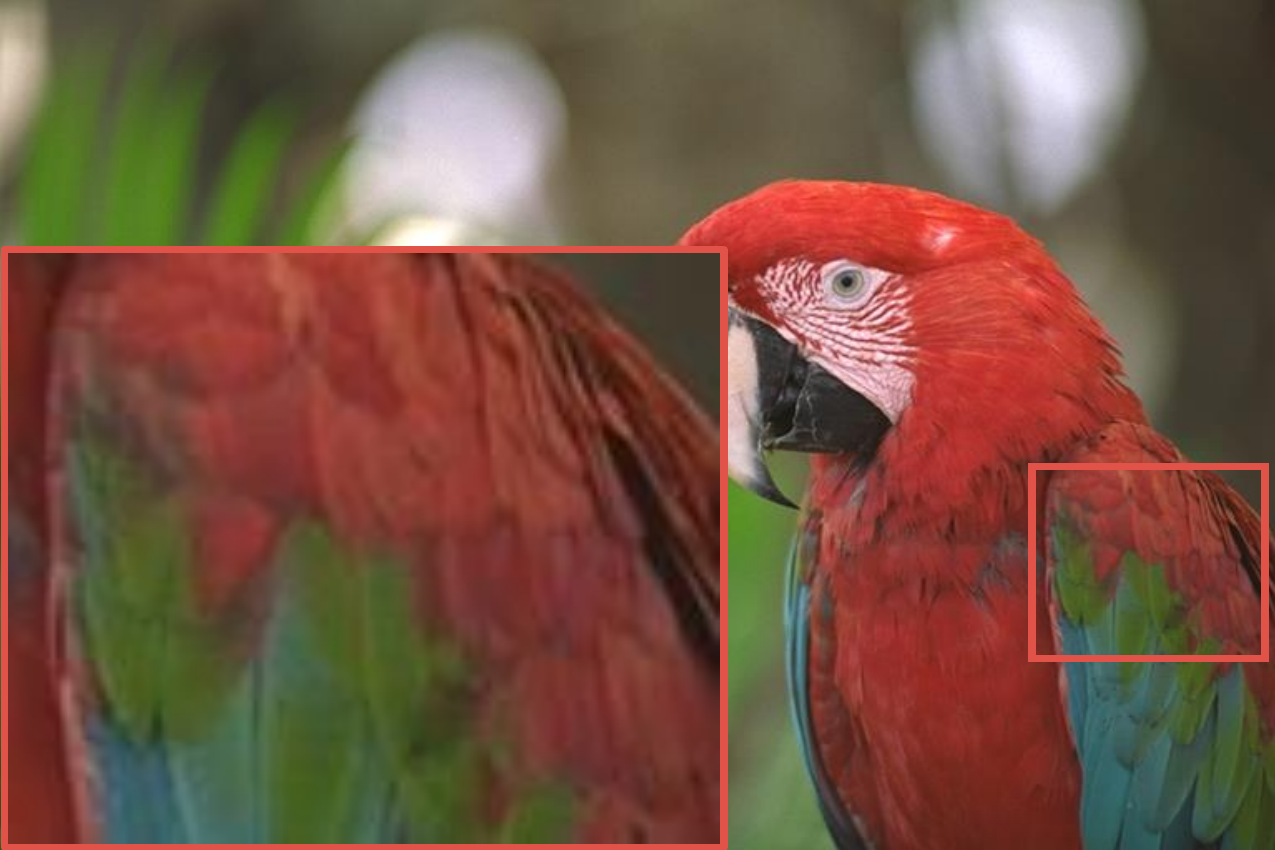}
    \\
    \raisebox{0.1\height}{\makebox[0.02\textwidth]{\rotatebox{90}{\makecell{\footnotesize 4-to-8 bits~\cite{sheikh2006statistical}}}}}
    \includegraphics[width=0.155\textwidth]{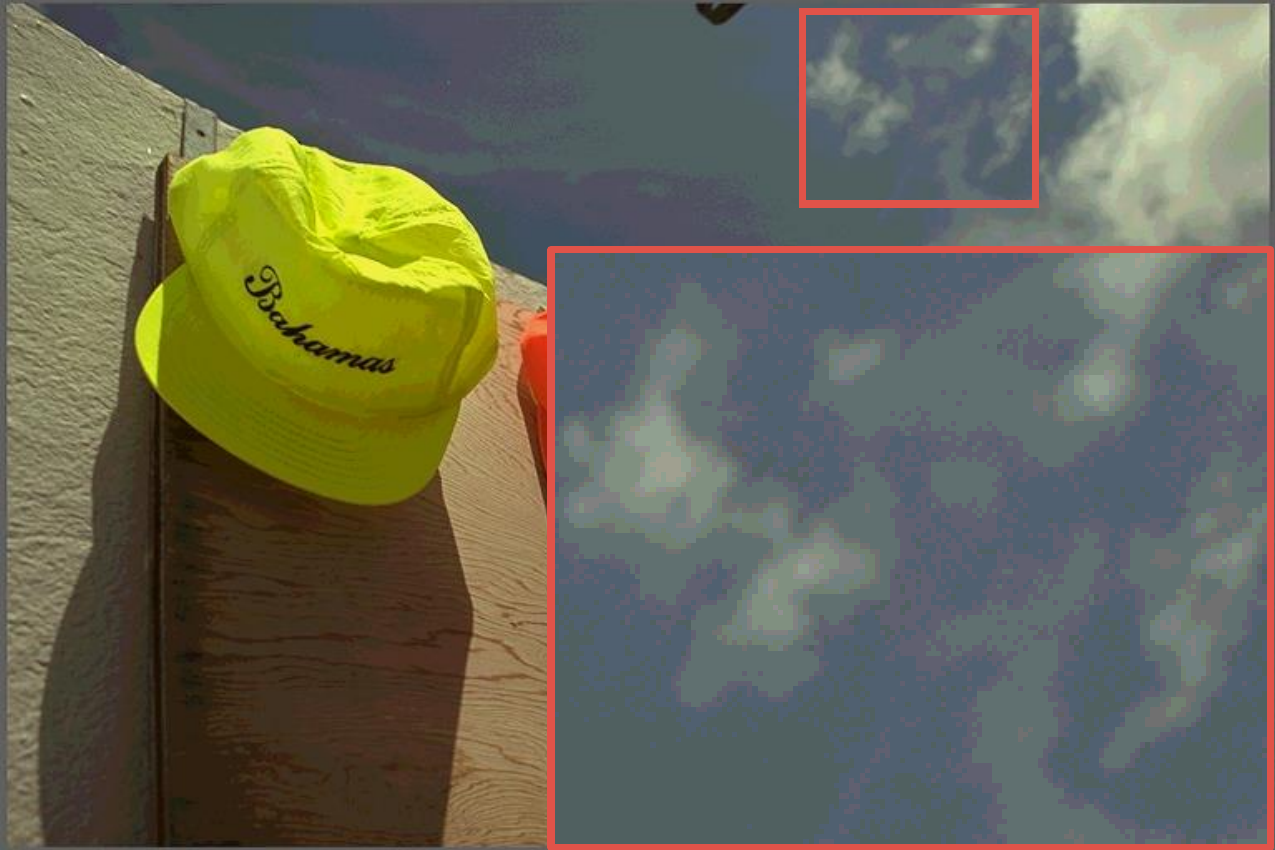}
    \includegraphics[width=0.155\textwidth]{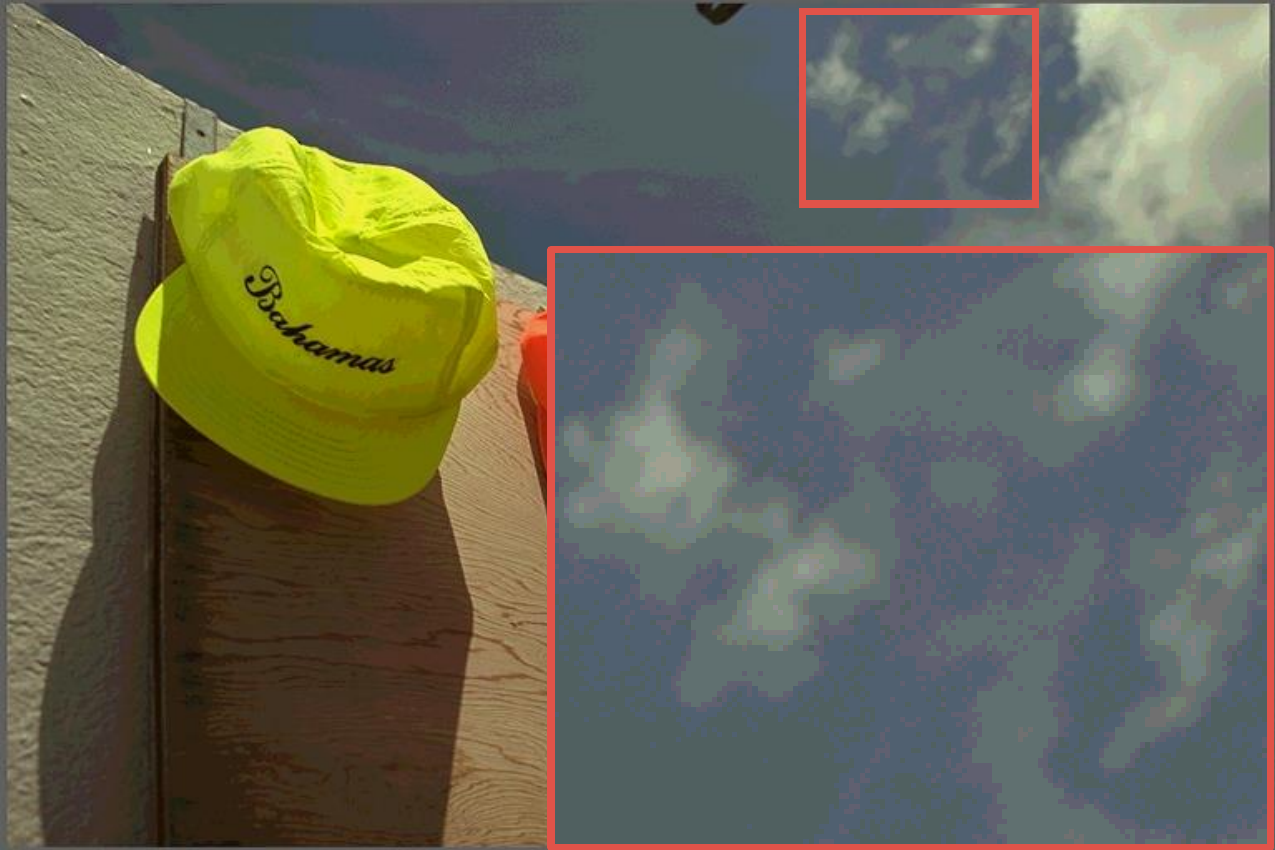}
    \includegraphics[width=0.155\textwidth]{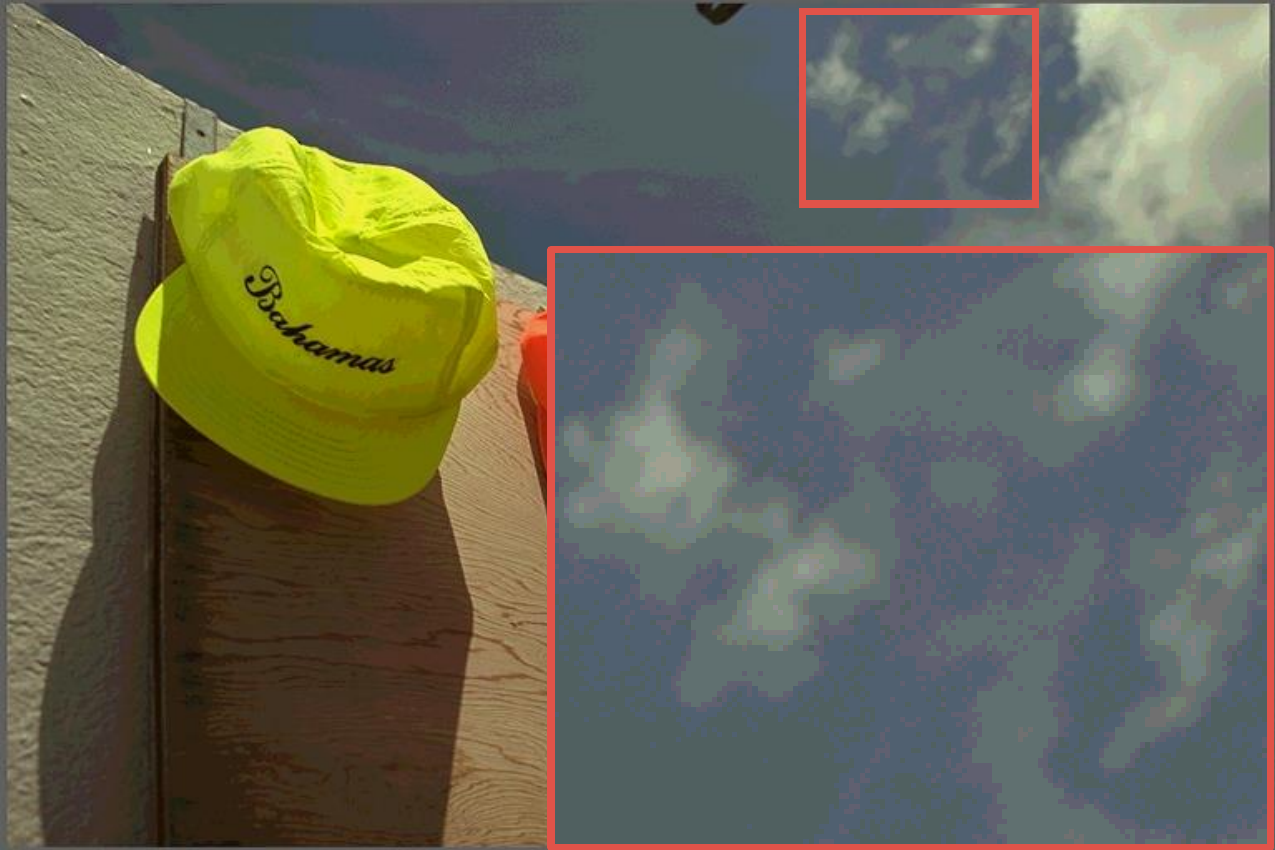}
    \includegraphics[width=0.155\textwidth]{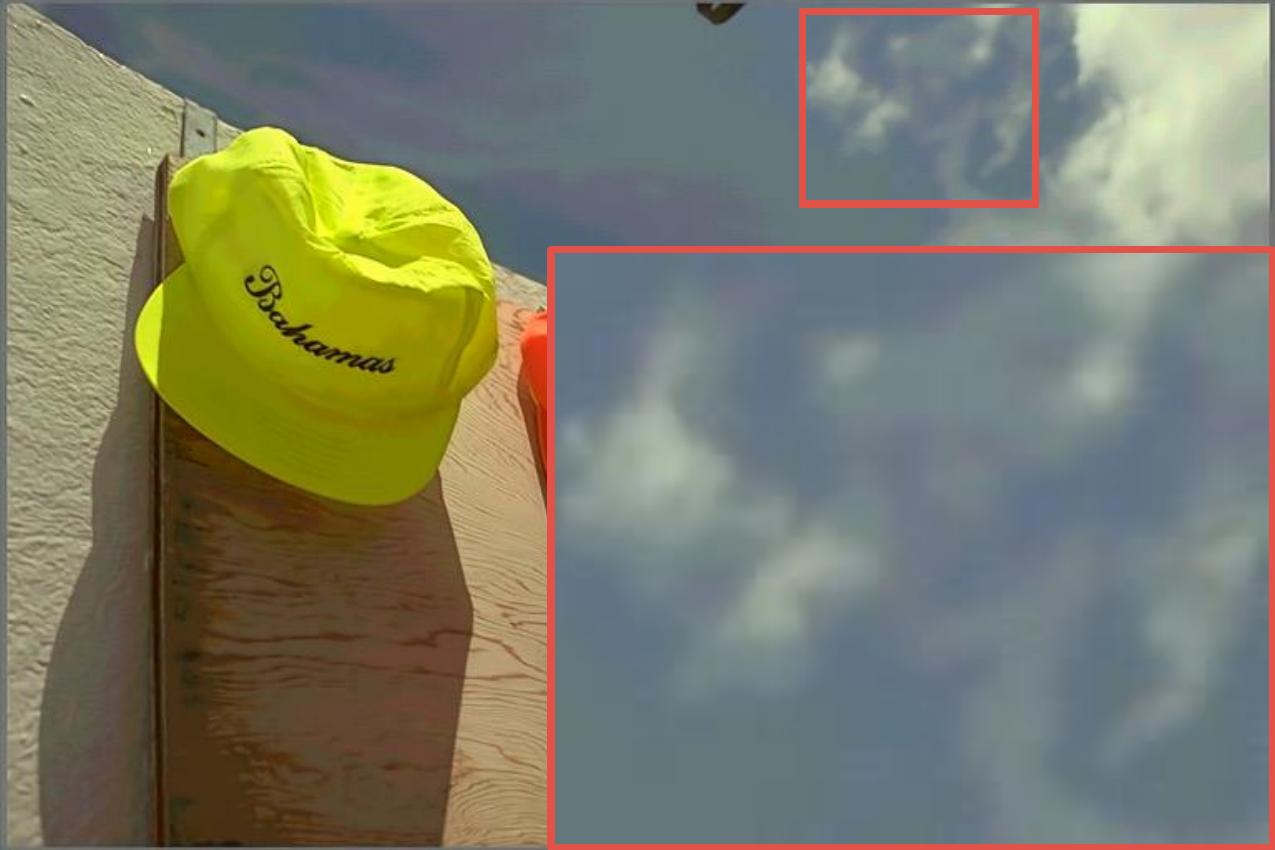}
    \includegraphics[width=0.155\textwidth]{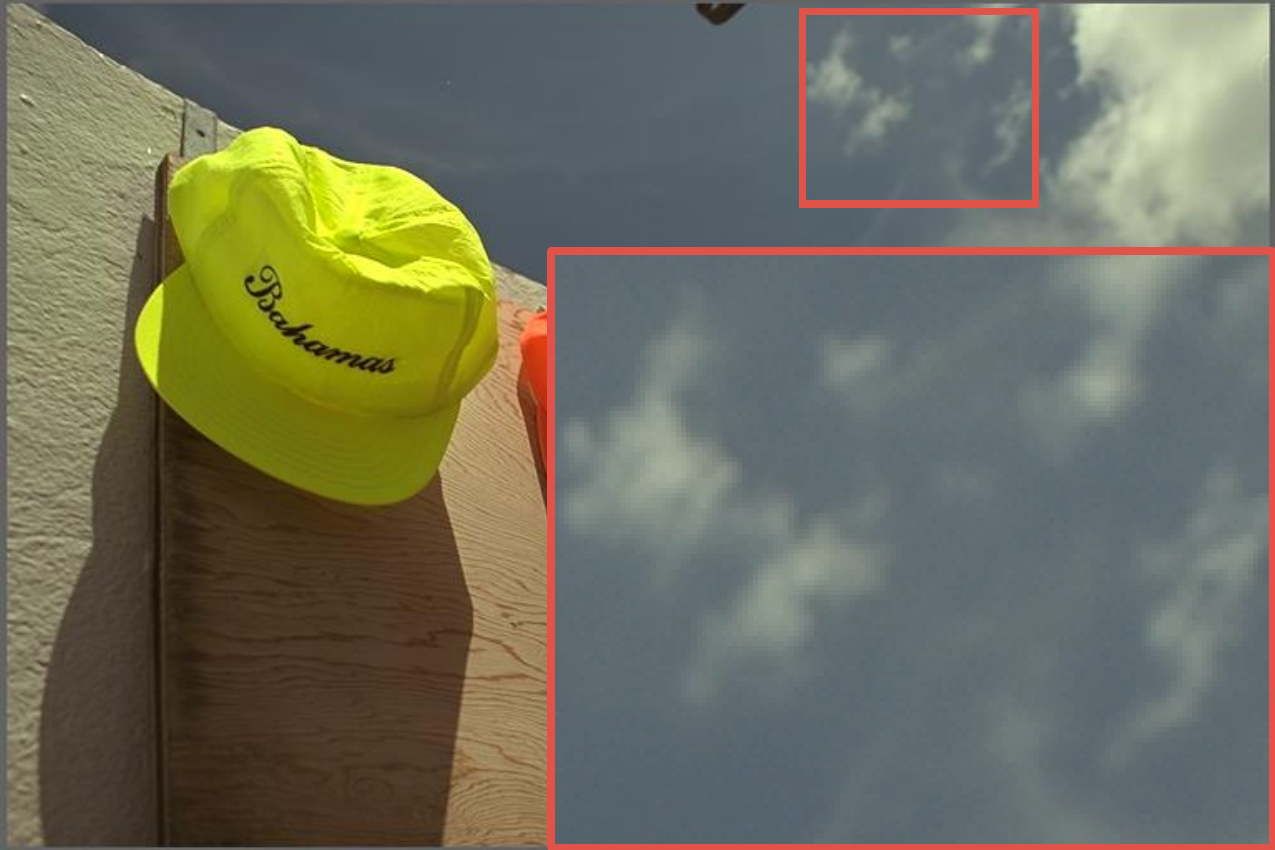}
    \includegraphics[width=0.155\textwidth]{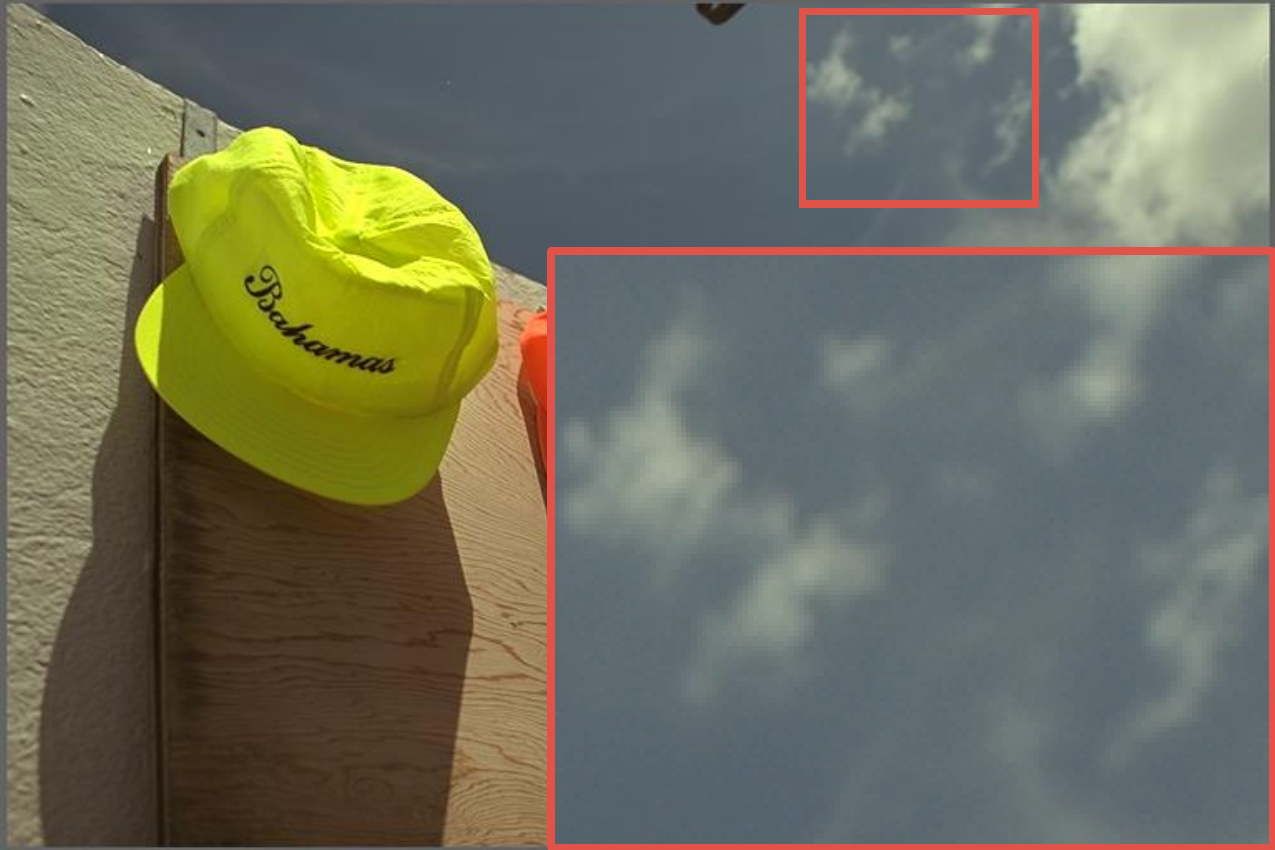}
    \\
    \makebox[0.02\textwidth]{}
    \makebox[0.155\textwidth]{\small Input}
    \makebox[0.155\textwidth]{\small PromptIR~\cite{potlapalli2024promptir}}
    \makebox[0.155\textwidth]{\small AdaIR~\cite{cui2025adair}}
    \makebox[0.155\textwidth]{\small Ours-SW}
    \makebox[0.155\textwidth]{\small Ours-SW retrained}
    \makebox[0.155\textwidth]{\small Ground truth}

    \vspace{-1.5mm}
    \caption{Qualitative results for unseen IR tasks, including JPEG artifact removal and 4-to-8 bit reconstruction. PromptIR~\cite{potlapalli2024promptir}, AdaIR~\cite{cui2025adair}, and Ours are not trained for these tasks, while Ours-retrained has a specified LoRA in an 8-degradation setup.} 
    \label{fig:qualitative-jpeg}
    \vspace{-3mm}

\end{figure*}

\vspace{-2mm}
\paragraph{Additional test sets.}~To assess generalization, we evaluate our model on three datasets not used during training: Rain100H~\cite{yang2017deep} for heavy deraining, HIDE~\cite{shen2019hide} for human-centric deblurring, and LoLv2-Real~\cite{yang2021sparse} for low-light image enhancement. Table~\ref{tab:unseen_datasets} presents the results for the methods trained in the five-task setup. Our method outperforms all competitors across all datasets and metrics, achieving substantial improvements of 3.98 dB over DiffUIR~\cite{zheng2024selective} on Rain100H, 1.65 dB over X-Restormer~\cite{chen2023x-restormer} on HIDE, and 1.39 dB over PromptIR~\cite{potlapalli2024promptir} on LoLv2-Real. These results highlight the effectiveness and robustness of our approach, which benefits from pretraining on synthetic degradations, enabling it to generalize across diverse natural images and degradation scenarios.
\vspace{-2mm}
\paragraph{Unseen IR tasks.}~Although our model is trained on five degradation types, we assess its adaptability to three additional IR tasks: JPEG artifact removal and bit-depth reconstruction using the Live1 dataset~\cite{sheikh2006statistical}, and desnowing using CityScapes-Snow-Medium~\cite{zhang2021snowcityscapes}. The results in Table~\ref{tab:unseen_tasks} demonstrate that our method consistently outperforms competing approaches. Moreover, by leveraging its modular scheme, we train independent LoRA adapters for each new task, blending them with the five-task baseline (denoted with $^*$ in Table~\ref{tab:unseen_tasks}). This strategy allows for refining the model with minimal training (3.6M parameters per task) while preserving previously acquired knowledge, enabled by our robust baseline model and three-phase training procedure. In contrast, other methods either require retraining the entire model from scratch or fine-tuning it without ensuring that prior knowledge is retained.

\begin{figure*}[ht!] \centering
    \raisebox{0.15\height}{\makebox[0.02\textwidth]{\rotatebox{90}{\makecell{\small Blur \& N}}}}
    \includegraphics[width=0.155\textwidth]{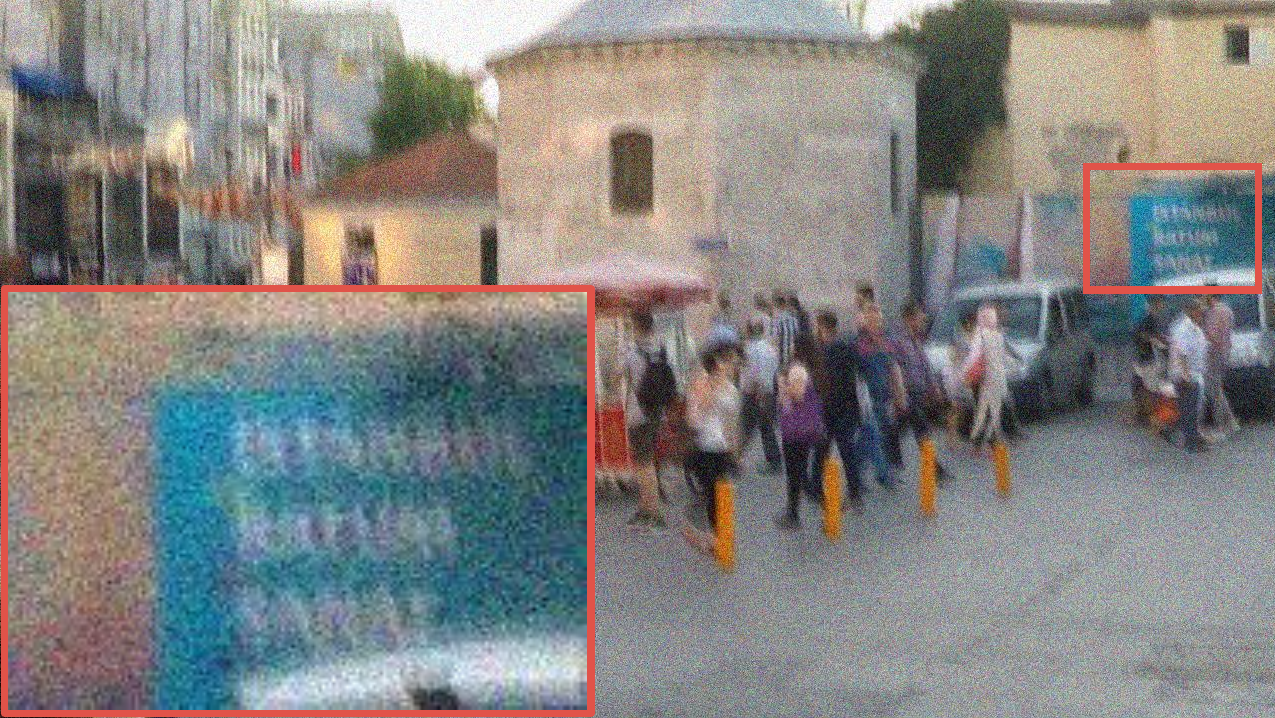}
    \includegraphics[width=0.155\textwidth]{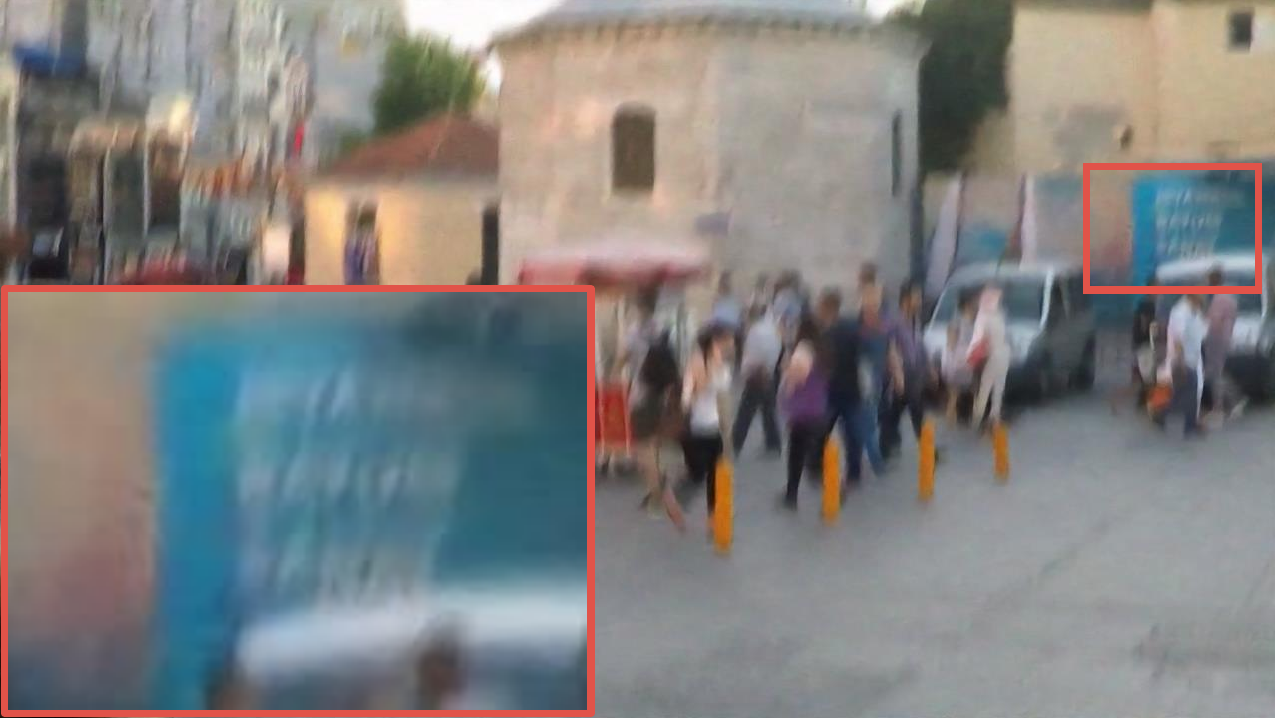}
    \includegraphics[width=0.155\textwidth]{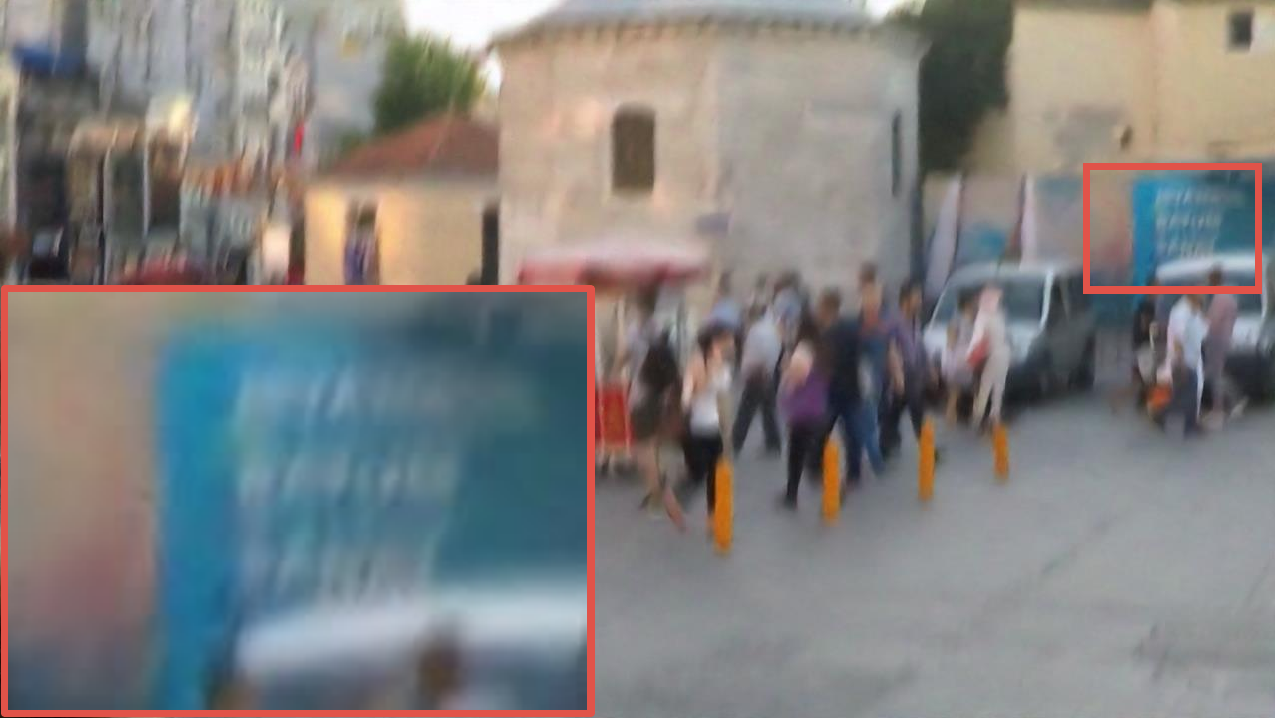}
    \includegraphics[width=0.155\textwidth]{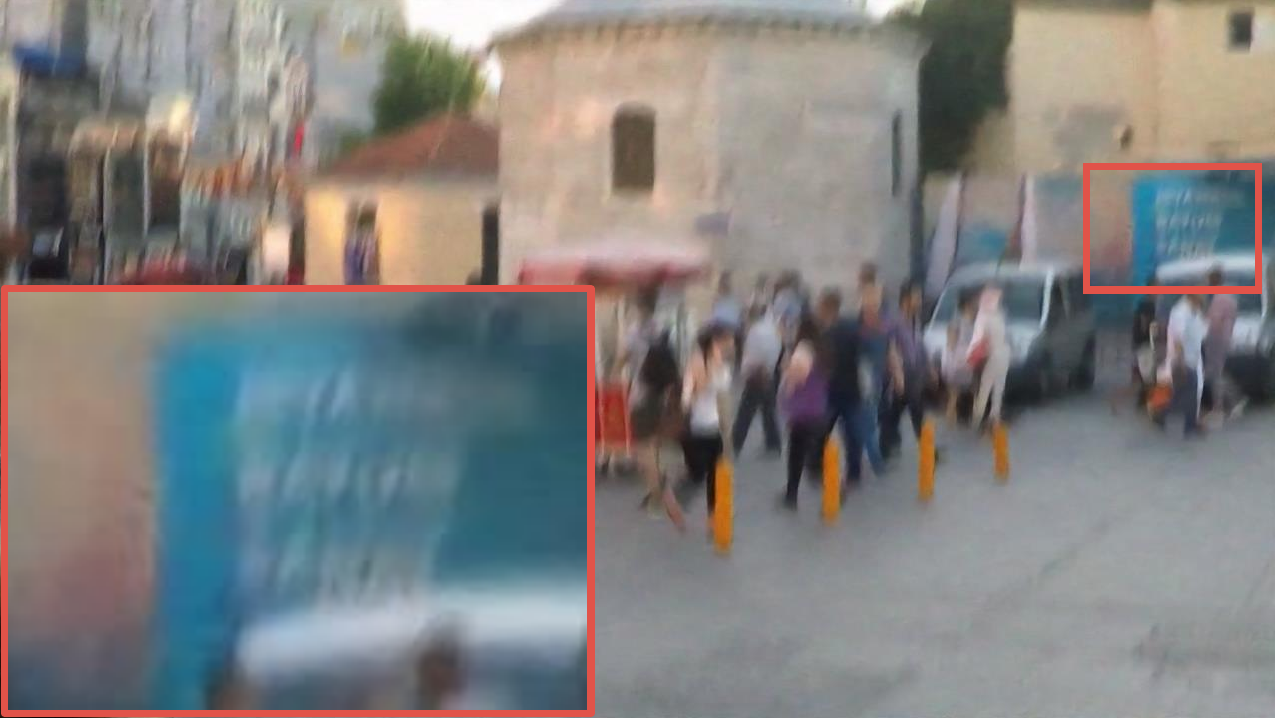}
    \includegraphics[width=0.155\textwidth]{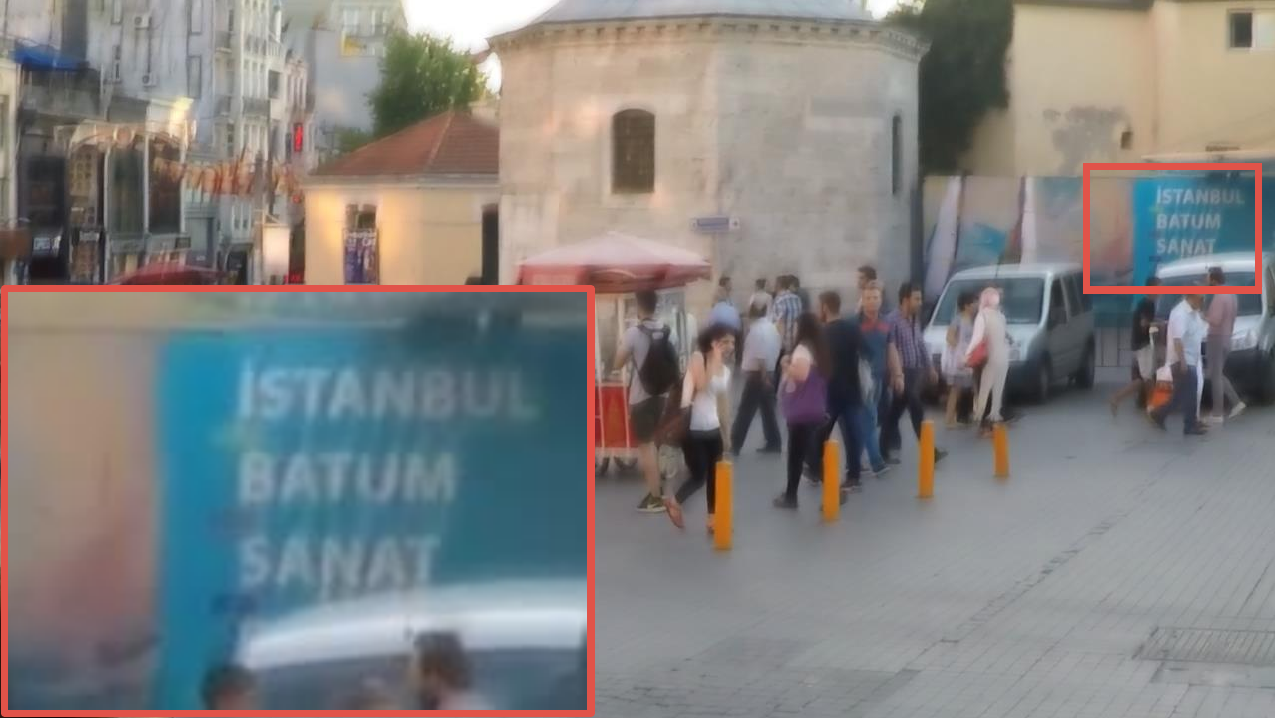}
    \includegraphics[width=0.155\textwidth]{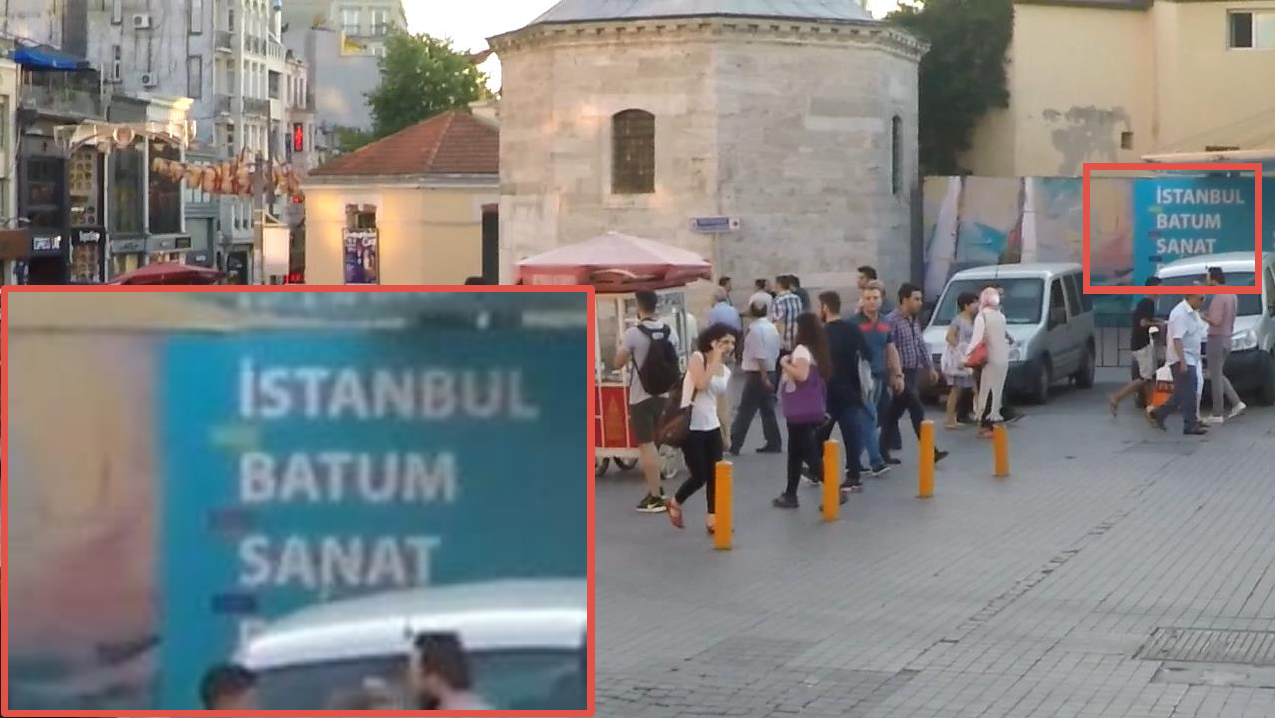}
    \\
    \raisebox{0.05\height}{\makebox[0.02\textwidth]{\rotatebox{90}{\makecell{\small Haze \& Snow}}}}
    \includegraphics[width=0.155\textwidth]{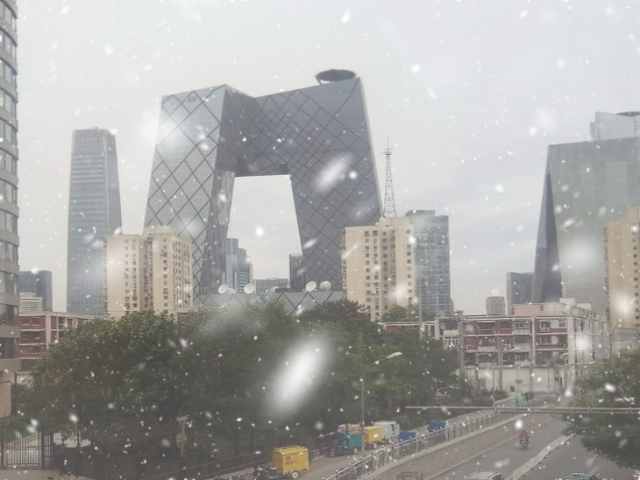}
    \includegraphics[width=0.155\textwidth]{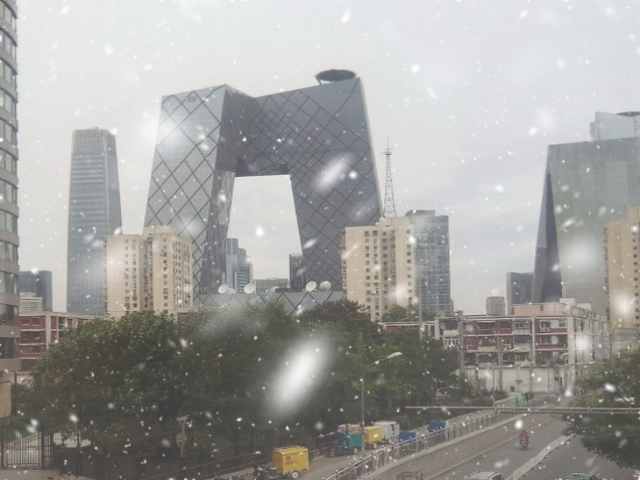}
    \includegraphics[width=0.155\textwidth]{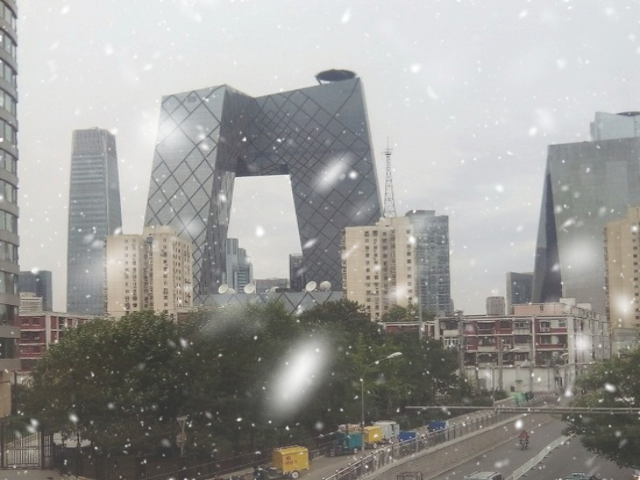}
    \includegraphics[width=0.155\textwidth]{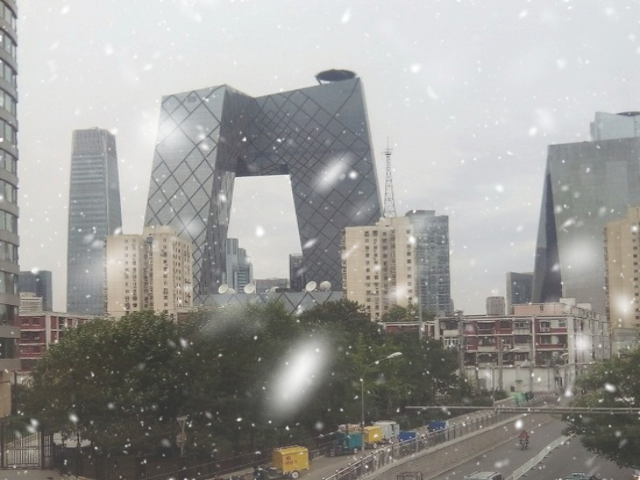}
    \includegraphics[width=0.155\textwidth]{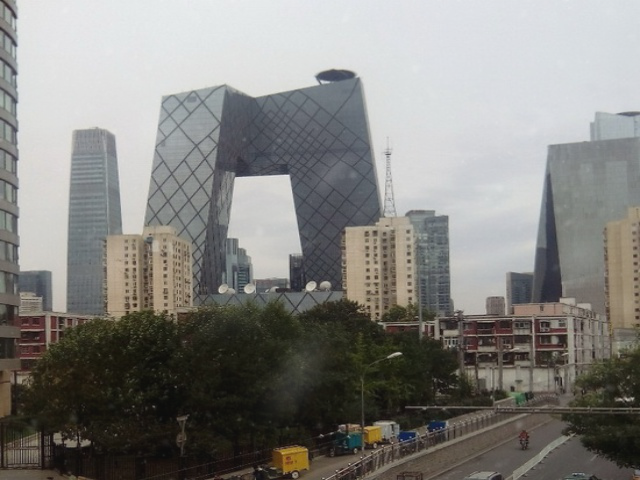}
    \includegraphics[width=0.155\textwidth]{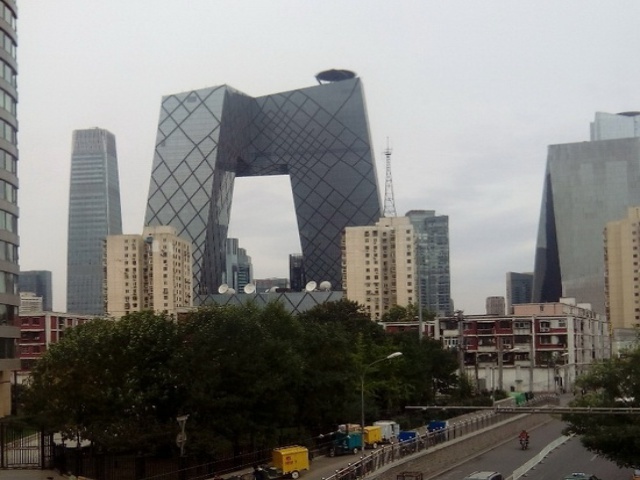}
    \\
    \makebox[0.02\textwidth]{}
    \makebox[0.155\textwidth]{\small Input}
    \makebox[0.155\textwidth]{\small Restormer~\cite{zamir2022restormer}}
    \makebox[0.155\textwidth]{\small PromptIR~\cite{potlapalli2024promptir}}
    \makebox[0.155\textwidth]{\small AdaIR~\cite{cui2025adair}}
    \makebox[0.155\textwidth]{\small Ours-SW}
    \makebox[0.155\textwidth]{\small Ground truth}

    \vspace{-1.5mm}
    \caption{Qualitative results for two examples of mixed degradations. The first row depicts an image with blur and noise from the GoPro dataset~\cite{Nah2017gopro}, while the second row shows an image with haze and snow from the SRRS dataset~\cite{chen2020srrs}. Columns display the input image, results from Restormer~\cite{zamir2022restormer}, PromptIR~\cite{potlapalli2024promptir}, our method, and the ground truth, respectively.}
    \vspace{-2mm}

    \label{fig:qualitative-mixed}
\end{figure*}

\paragraph{Mixed degradations.}~We also evaluate our approach on three mixed-degradation scenarios: blur with noise ($\sigma{=}25$) using GoPro~\cite{Nah2017gopro}, blur with JPEG artifacts using REDS~\cite{nah2019reds}, and haze with snow using SRRS~\cite{chen2020srrs}. Notably, JPEG artifacts and snow are unseen during training. As shown in Table~\ref{tab:mixed_degradations}, our model outperforms all competitors by at least 1 dB PSNR in all three cases. Addressing blur and noise simultaneously is particularly challenging due to their opposing frequency characteristics. Our specialize-then-merge scheme allows the lightweight estimator to efficiently blend the task-specific adaptors, offering greater flexibility than end-to-end models trained on fixed degradation types.



\vspace{-2mm}
\paragraph{Qualitative results.}~In Figure~\ref{fig:qualitative}, we present examples featuring three types of single degradations from the five-task setup. Our approach effectively enhances the license plate in the first row, reduces noise in low-light conditions in the second row, and removes heavy rain streaks in the third. In Figure~\ref{fig:qualitative-jpeg}, we show unseen IR tasks including JPEG artifact removal and 4-to-8-bit reconstruction using the Live1 dataset~\cite{sheikh2006statistical}. Our method successfully removes JPEG artifacts from the parrot’s plumage and artifacts caused by bit-depth reduction from the cloud and blue sky. By training a new adapter for these tasks (ours-SW retrained), we achieve superior results with minimal additional training. Finally, in Figure~\ref{fig:qualitative-mixed}, we demonstrate our method’s performance under mixed degradations. The first row shows that our method achieves the best reconstruction of the text in the image, while the second row illustrates our model’s effectiveness in removing haze and snowflakes.

\begin{table}[t!]
    \centering
    \caption{\small Quantitative results on datasets with mixed degradations.}
    \vspace{-1mm}
    \setlength{\tabcolsep}{3.2pt} 
    \small
    \begin{tabular}{lcccccc}
    \toprule
    \multirow{2}{*}{PSNR/SSIM} & \multicolumn{2}{c}{Blur\&Noise} & \multicolumn{2}{c}{Blur\&JPEG} & \multicolumn{2}{c}{Haze\&Snow} \\
    \cmidrule{2-7}
    & \multicolumn{2}{c}{GoPro} & \multicolumn{2}{c}{REDS} & \multicolumn{2}{c}{SRRS} \\
    \midrule
    IDR~\cite{zhang2023idr} & 21.98 & .683 & 23.02 & .681 & 20.51 & .789\\
    X-Restormer~\cite{chen2023x-restormer} & 22.67 & .669 & 23.98 & .710 & 20.76 & .805 \\
    DiffUIR~\cite{zheng2024selective} & 22.71 & .670 & 24.00 & .711 & 20.86 & .802 \\
    Restormer~\cite{zamir2022restormer} & 22.35 & .662 & 23.24 & .698 & 20.76 & .800\\
    AdaIR~\cite{cui2025adair} & 22.45 & .663 & 23.16 & .689 & 20.77 & .802 \\
    PromptIR~\cite{potlapalli2024promptir} & 22.89 & .671 & 23.92 & .705 & 20.94 & .803\\
    \midrule
    Ours OH & \second{24.30} & \second{.743} & \second{24.81} & \second{.717} & \second{21.48} & \second{.834} \\
    Ours SW & \best{25.14} & \best{.750} & \best{24.97} & \best{.719} & \best{22.09} & \best{.839} \\
    \bottomrule
    \end{tabular}
    \label{tab:mixed_degradations}
\end{table}

\begin{table}[t]
    \centering
    \setlength{\tabcolsep}{2pt} 

    \caption{\small Ablation studies on pretraining strategies for our method (a), and PromptIR~\cite{potlapalli2024promptir} (b). We report the five-task setup average.}
    \small
    \vspace{-1mm}
    \begin{minipage}{0.48\columnwidth} 
        \centering
        \begin{tabular}{lccc}
        \toprule
        (a) ABAIR & PSNR & SSIM\\ 
        \midrule
        IR datasets & 29.50 & .892 \\
        GLD+2 global & 27.68 & .887 \\
        GLD+1 global & 30.63 & .913 \\
        GLD+CutMix & 31.11 & .920 \\
        \hspace{2mm} + Aux. seg. & 31.21 & .921 \\
        \bottomrule
        \end{tabular}
    \end{minipage}
    \hspace{2pt}
    \begin{minipage}{0.48\columnwidth} 
        \centering
        \begin{tabular}{lcc}
        \toprule
        (b) & PSNR & SSIM\\
        \midrule
        PromptIR~\cite{potlapalli2024promptir} & 28.58 & .885 \\
        RAM~\cite{qin2024restore} & 28.79 & .889 \\
        Art~\cite{wu2024harmony} & 28.83 & .889 \\
        GLD+1 global & 29.27 & .902\\
        GLD+CutMix & 29.58 & .908\\
        \bottomrule
        \end{tabular}
    \end{minipage}%
    
    \vspace{-4mm}
    \label{tab:synthetic_ablation}
\end{table}

\vspace{-4mm}
\paragraph{Ablation studies.}~Our method consistently outperforms previous state-of-the-art approaches across seen, unseen, and mixed degradations, as demonstrated in all the tables. To further assess its effectiveness, we conduct additional experiments regarding the impact of our Phase I pretraining strategy. Table~\ref{tab:synthetic_ablation} reports the results averaged over the five-task setup.

First, we evaluate the impact of the synthetic pretraining to the final performance of our method. Table~\ref{tab:synthetic_ablation} (a) compares different pretraining setups: standard IR datasets as pretraining ({\it IR datasets}), Google Landmarks dataset with one and two simultaneous global synthetic degradations applied across the entire image, ({\it GLD+1 global} and {\it GLD+2 global}), respectively, as well as incorporating the Degradation CutMix ({\it GLD+Cutmix}) and the auxiliary segmentation head with its corresponding loss ({\it Aux. seg.}). Next, we compare our synthetic pretraining approach with the pretraining strategy from RAM~\cite{qin2024restore} and the optimization technique from Art~\cite{wu2024harmony} on the PromptIR~\cite{potlapalli2024promptir} model. We select PromptIR as a well-established baseline for evaluation. The results of these comparisons are presented in Table~\ref{tab:synthetic_ablation} (b).

From Table~\ref{tab:synthetic_ablation}, we derive four key observations: (i)~Our synthetic pretraining substantially outperforms other pretraining and optimization techniques. (ii)~Our three-phase approach is effective, as even without using synthetic pretraining ({\it IR datasets}), our method surpasses the baseline PromptIR model. (iii)~Training with multiple simultaneous degradations ({\it GLD+2 global}) negatively impacts performance, likely due to excessive degradation in Phase I, which hinders the learning of robust baseline. (iv)~Both Degradation CutMix and the auxiliary segmentation head contribute substantially to performance improvements. These findings highlight the importance of both synthetic pretraining and the three-phase training scheme in achieving optimal results. Additional results and ablation studies for Phases II and III, including analyses of different adapter types, ranks, and alternative strategies for blending the adapters, are provided in the Supplementary Material.

\vspace{-1mm}
\section{Conclusion}
In this work, we introduce an adaptive blind all-in-one IR (ABAIR) model aiming towards practical IR. We design a specialize-then-merge scheme with dedicated adapters for strongly handling specific distortions and a flexible architecture for dealing with unseen and mixed degradations. We first developed a pretraining pipeline featuring multiple synthetic degradations over a large dataset that boosts the model's generalization. Second, we derived compact per-task adapters that robustly adapt to specific degradations. Third, we developed a lightweight degradation estimator that identifies varying degradations to blend the respective adapters. Our method is also capable of efficiently incorporate new degradations by training a small fraction of parameters. Our model largely outperforms the state-of-the-art on five- and three-task IR setups, and shows improved generalization to unseen datasets and IR tasks.

\section*{Acknowledgements}
DSL, LH, and JVC were supported by Grant PID2021-128178OB-I00 funded by MCIN/AEI/10.13039/ 501100011033 and by ERDF "A way of making Europe", by the  Departament de Recerca i Universitats from Generalitat de Catalunya with reference 2021SGR01499, and by the Generalitat de Catalunya CERCA Program. DSL also acknowledges the FPI grant from Spanish Ministry of Science and Innovation (PRE2022-101525). LH was also supported by the Ramon y Cajal grant RYC2019-027020-I. SS was supported by the HORIZON MSCA Postdoctoral Fellowships funded by the European Union (project number 101152858)

\section{Supplementary Material}
\appendix

We provide additional material to supplement our main submission. Specifically, we cover:

\begin{enumerate}[label=\Alph*.]
    \item Details on our baseline architecture.
    \item Description of the generation process for synthetic degradations, including rain, haze, noise, blur, and low-light conditions. Additional dataset examples.
    \item Additional ablation experiments on LoRA’s rank and other decomposition methods.
    \item Details on our estimator architecture and additional analysis of the estimator's performance.
    \item Additional qualitative results on known datasets, unseen tasks, and images with mixed degradations.
\end{enumerate}
    
\section{Baseline Architecture Details}
\begin{figure*}[t!]
    \centering
    \includegraphics[width=\linewidth]{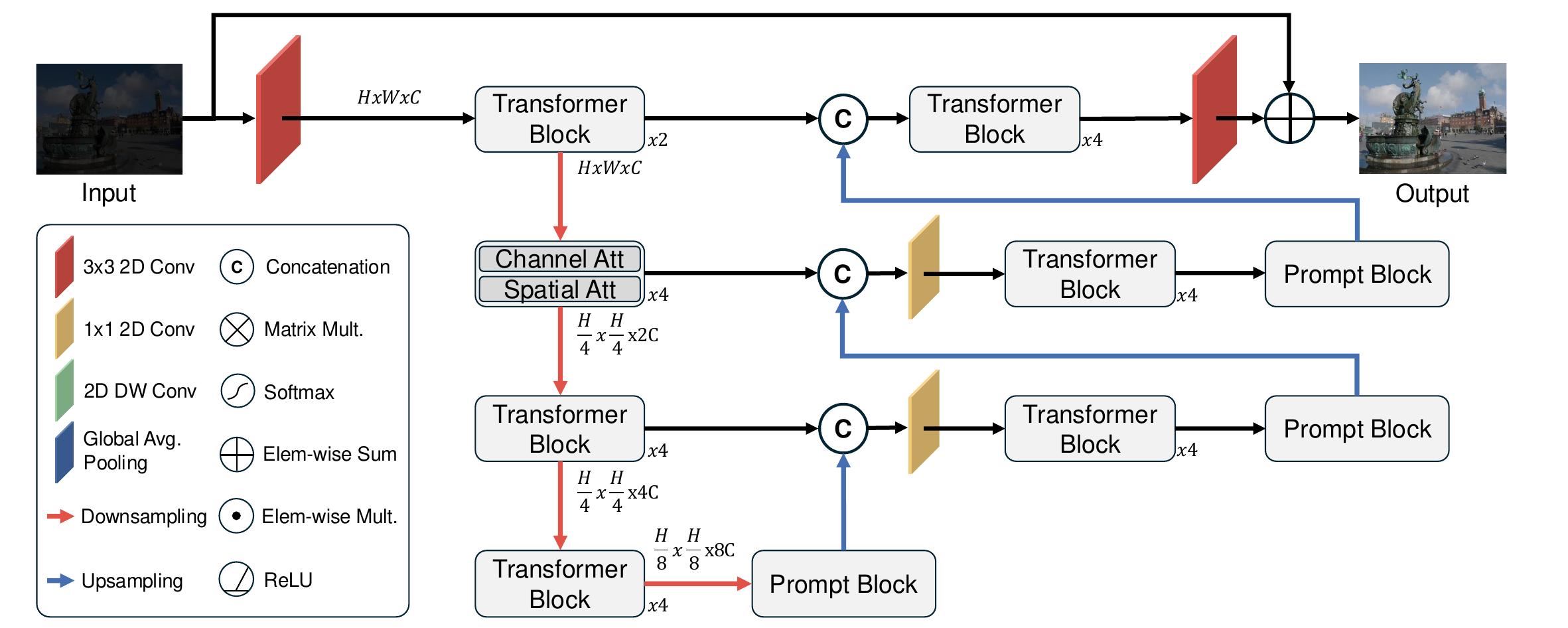}
    \caption{Overview of our baseline model. The input degraded image is processed through a four-level U-shaped network with transformer blocks. In the upsampling path, prompt blocks are integrated to assist the model in capturing degradation-specific information.}
    \label{fig:baseline}
\end{figure*}

We present an Adaptive Blind All-in-One Image Restoration (ABAIR) method, designed to bridge the gap between IR techniques and their application in practical complex scenarios. Our approach follows a three-phase scheme. The first phase involves pre-training an IR baseline using natural images with synthetic degradations. In this section, we describe this baseline architecture. ~\Cref{fig:baseline} shows the details of our baseline model. Our baseline model adopts the Restormer~\cite{zamir2022restormer} architecture, a transformer-based UNet-like framework. Given a degraded image, the model first applies a convolutional layer to extract low-level features of size $H \times W \times C$, where $H$ and $W$ are the spatial dimensions, and $C{=}48$ in all our experiments. These features are then processed through a four-level encoder-decoder structure composed of transformer blocks, with pixel unshuffling and shuffling~\cite{shi2016real} used for downsampling and upsampling, respectively. Finally, a convolutional layer generates the residual image, which is added to the degraded input image to produce the restored output.

\begin{figure}[t]
    \centering
    \includegraphics[width=\linewidth]{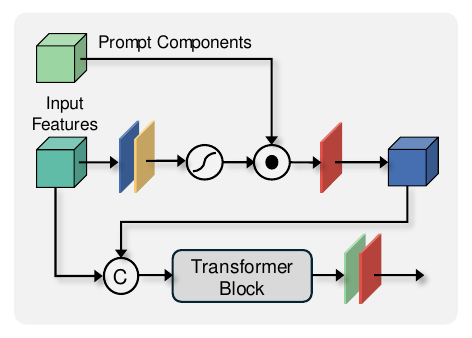}
    \caption{Overview of our Prompt Block. The input features, derived from the output of a transformer block, are modulated through a weighted element-wise multiplication with a set of model parameters. These modulated features are then further refined using an additional transformer block, producing the enhanced output features of the Prompt Block.}
    \label{fig:prompt_block}
\end{figure}

Each transformer block comprises a channel self-attention module followed by a spatial self-attention module. For channel self-attention, we adopt the implementation from Zamir et al.~\cite{zamir2022restormer}, while for spatial attention, we use the overlapping cross-attention mechanism proposed by Chen et al.~\cite{chen2023activating}. This combination effectively addresses the limitations of Restormer’s U-shaped architecture, particularly its difficulty in reconstructing high-frequency details~\cite{chen2023x-restormer}. Additionally, inspired by Potlapalli et al.~\cite{potlapalli2024promptir}, we integrate a Prompt Block between the transformer blocks in the upsampling path. This block facilitates the architecture in identifying degradation-specific features in the input image by combining extracted features with a set of model parameters. The design of our Prompt Block is depicted in~\cref{fig:prompt_block}.

\section{Synthetic Degradations Generation}
\begin{figure*}[t] \centering
    \makebox[0.19\textwidth]{\small $d{=}0.005$, $l{=}25$, $\theta{=}70^{\circ}$,}
    \makebox[0.19\textwidth]{\small min${=}0.2$, max${=}0.7$}
    \makebox[0.19\textwidth]{}
    \makebox[0.19\textwidth]{}
    \makebox[0.19\textwidth]{}
    \makebox[0.19\textwidth]{\small $s{=}1$, and $w{=}0.75$}
    \makebox[0.19\textwidth]{\small $A{=}140$}
    \makebox[0.19\textwidth]{\small $\sigma{=}15$}
    \makebox[0.19\textwidth]{\small $d{=}9$ and $\Theta{=}0^{\circ}$}
    \makebox[0.19\textwidth]{\small $c{=}0.25$ and $\sigma{=}0.5$}

    \vspace{1mm}
    
    \includegraphics[width=0.19\textwidth]{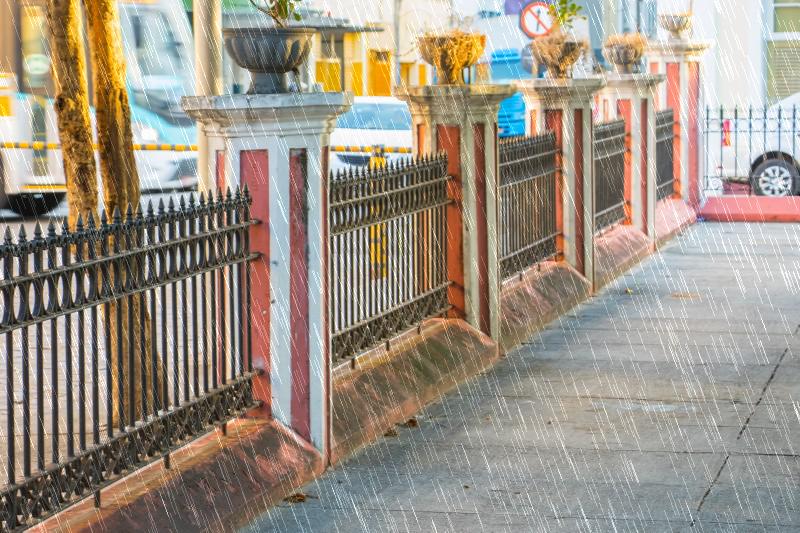}
    \includegraphics[width=0.19\textwidth]{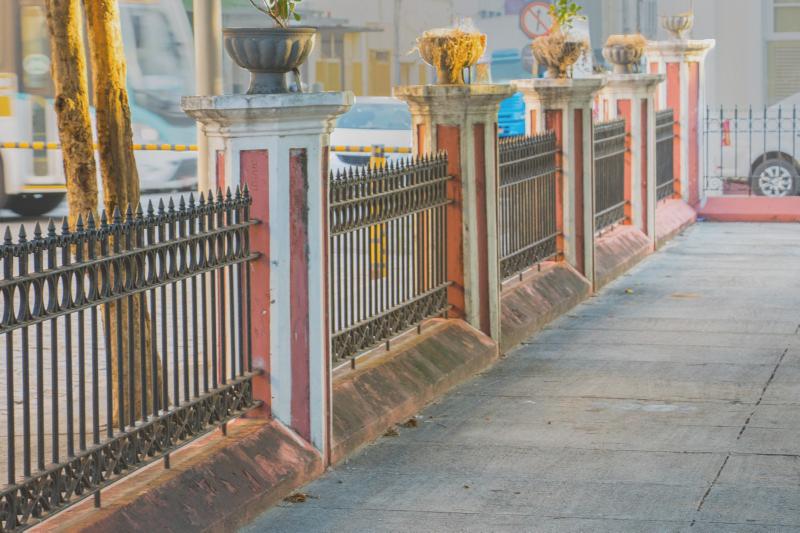}
    \includegraphics[width=0.19\textwidth]{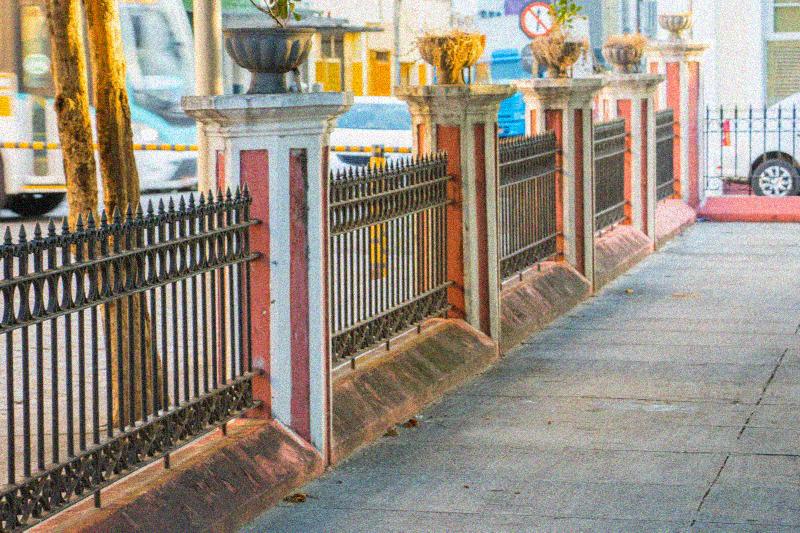}
    \includegraphics[width=0.19\textwidth]{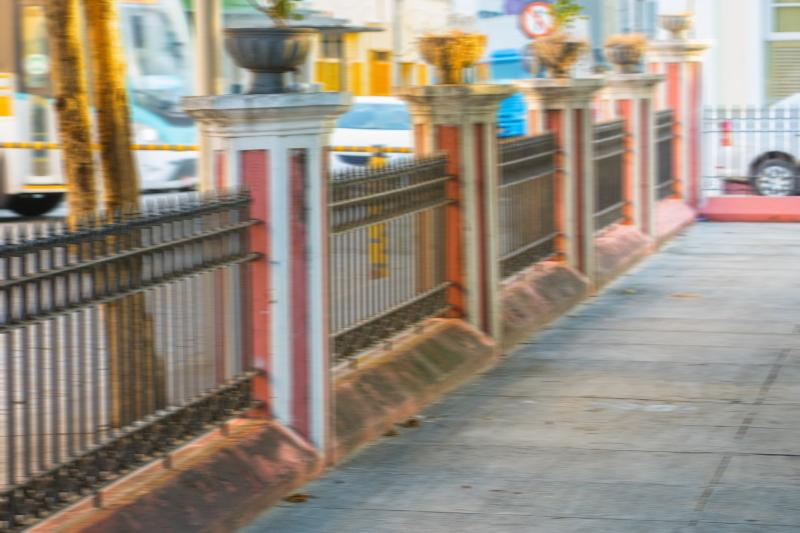}
    \includegraphics[width=0.19\textwidth]{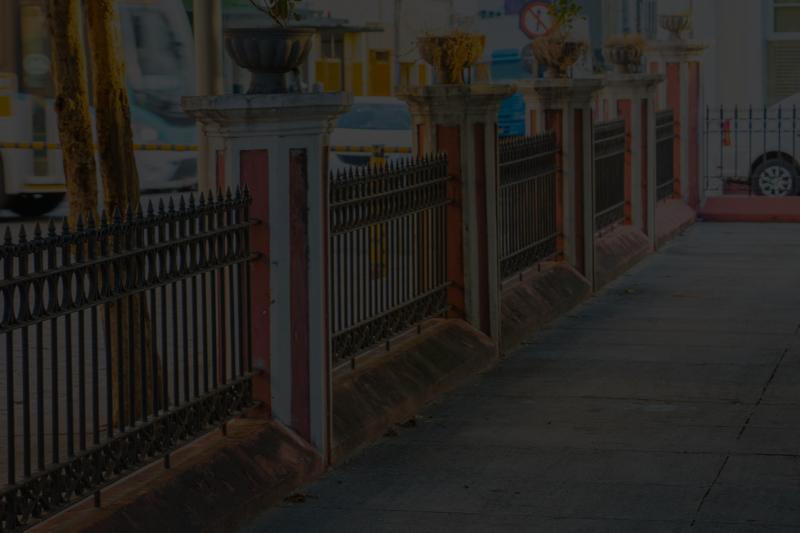}
    \\

    \makebox[0.19\textwidth]{\small $d{=}0.01$, $l{=}30$, $\theta{=}90^{\circ}$,}
    \makebox[0.19\textwidth]{\small min${=}0.3$, max${=}0.8$}
    \makebox[0.19\textwidth]{}
    \makebox[0.19\textwidth]{}
    \makebox[0.19\textwidth]{}
    \makebox[0.19\textwidth]{\small $s{=}2$, and $w{=}1$}
    \makebox[0.19\textwidth]{\small $A{=}160$}
    \makebox[0.19\textwidth]{\small $\sigma{=}25$}
    \makebox[0.19\textwidth]{\small $d{=}25$ and $\Theta{=}45^{\circ}$}
    \makebox[0.19\textwidth]{\small $c{=}0.5$ and $\sigma{=}1$}

    \vspace{1mm}
    
    \includegraphics[width=0.19\textwidth]{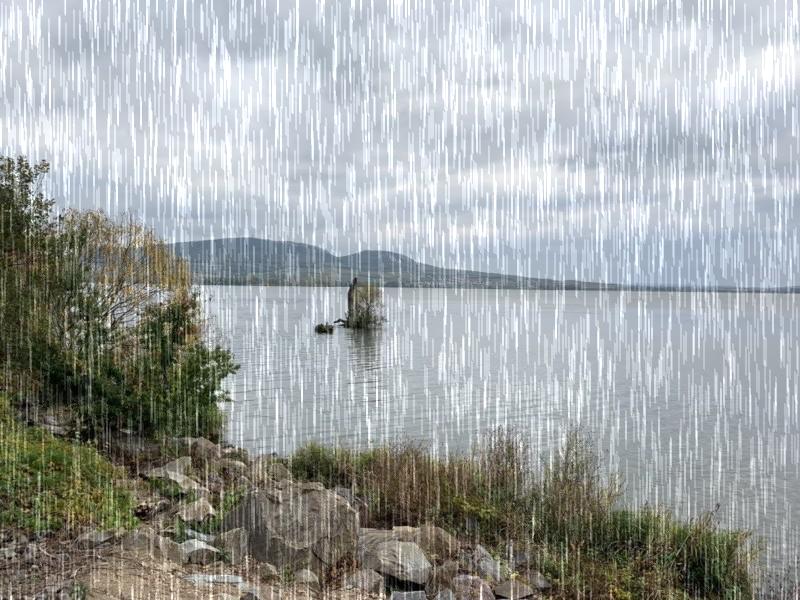}
    \includegraphics[width=0.19\textwidth]{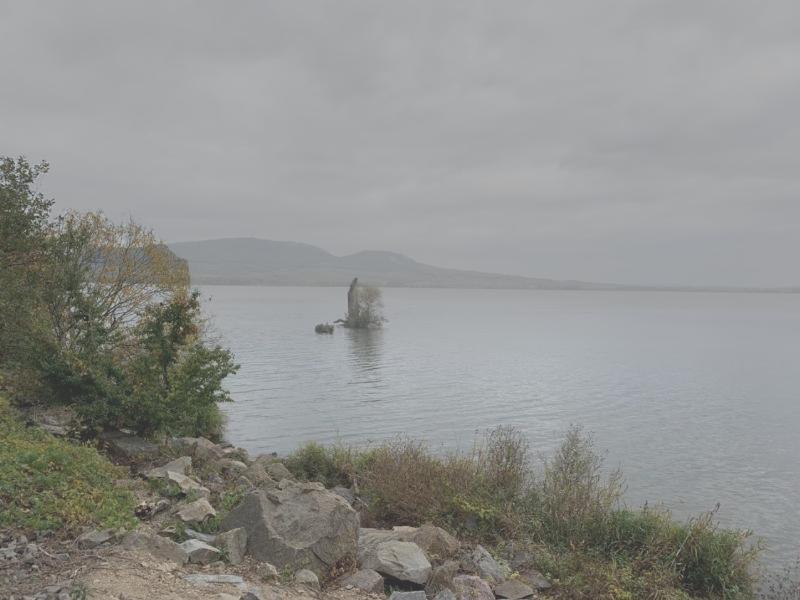}
    \includegraphics[width=0.19\textwidth]{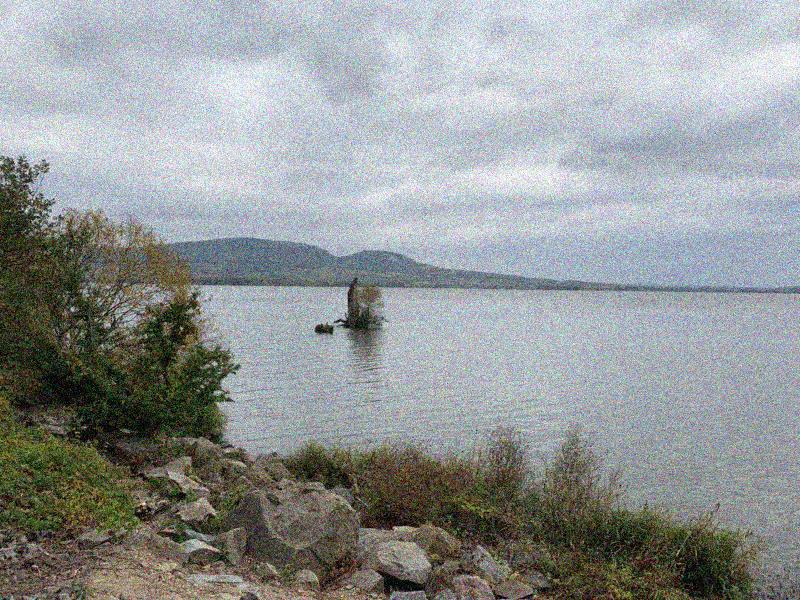}
    \includegraphics[width=0.19\textwidth]{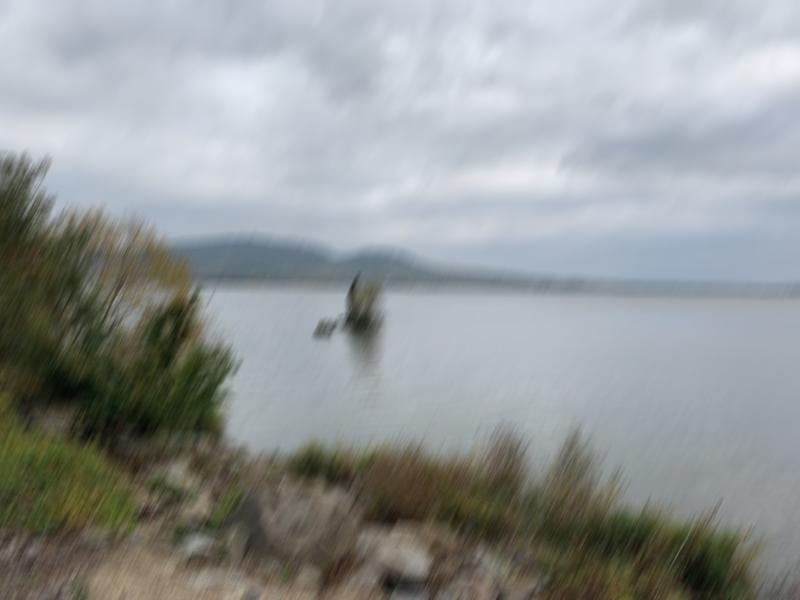}
    \includegraphics[width=0.19\textwidth]{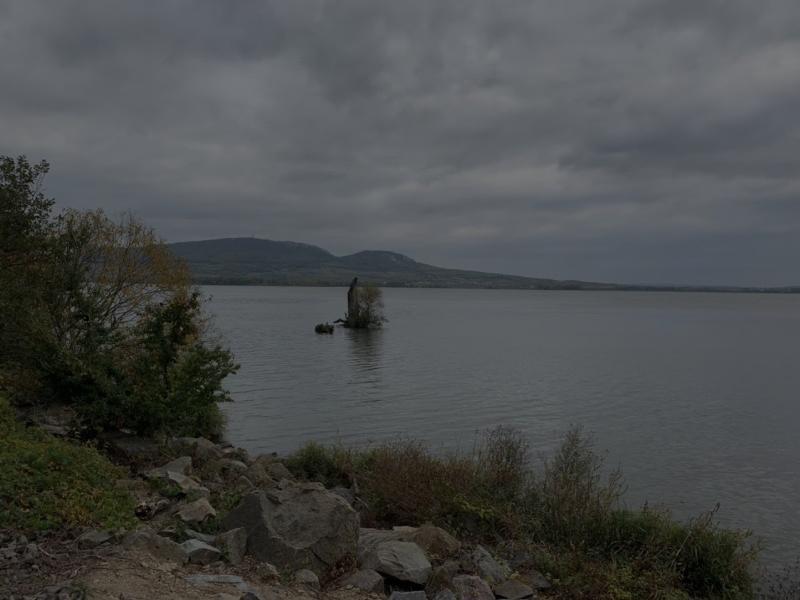}
    \\

    \makebox[0.19\textwidth]{\small $d{=}0.01$, $l{=}30$, $\theta{=}90^{\circ}$,}
    \makebox[0.19\textwidth]{\small min${=}0.3$, max${=}0.8$}
    \makebox[0.19\textwidth]{}
    \makebox[0.19\textwidth]{}
    \makebox[0.19\textwidth]{}
    \makebox[0.19\textwidth]{\small $s{=}3$, and $w{=}1$}
    \makebox[0.19\textwidth]{\small $A{=}180$}
    \makebox[0.19\textwidth]{\small $\sigma{=}25$}
    \makebox[0.19\textwidth]{\small $d{=}25$ and $\Theta{=}90^{\circ}$}
    \makebox[0.19\textwidth]{\small $c{=}0.5$ and $\sigma{=}1.5$}

    \vspace{1mm}

    \includegraphics[width=0.19\textwidth]{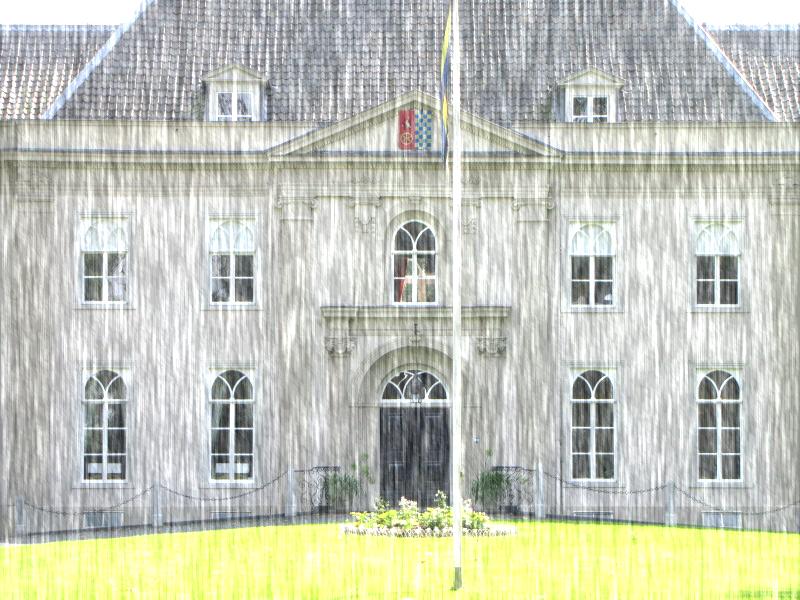}
    \includegraphics[width=0.19\textwidth]{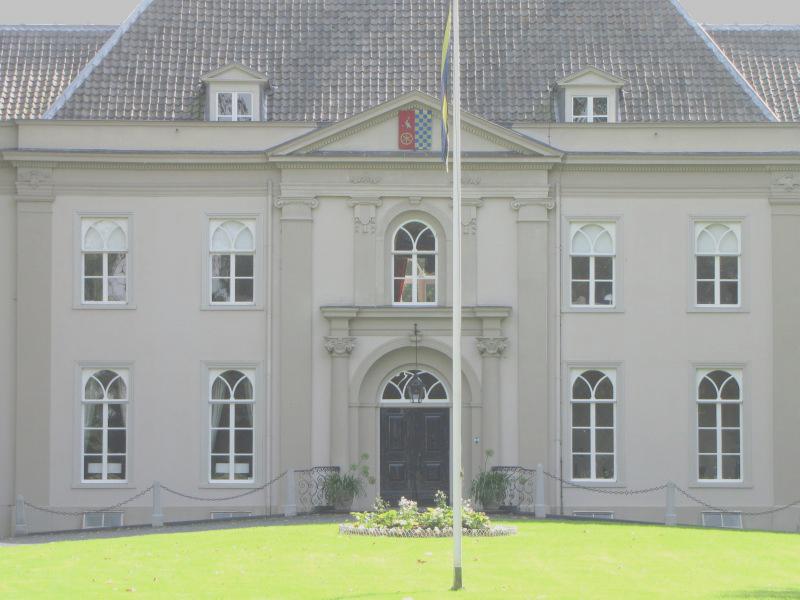}
    \includegraphics[width=0.19\textwidth]{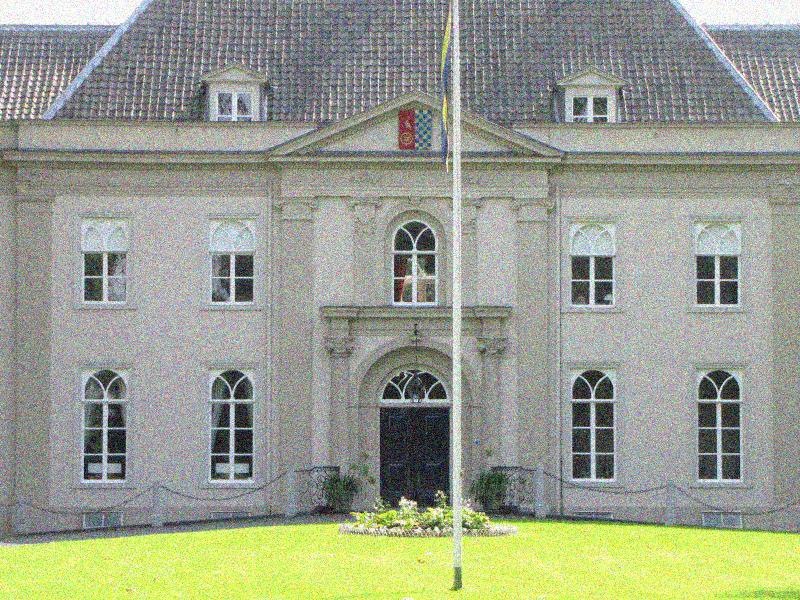}
    \includegraphics[width=0.19\textwidth]{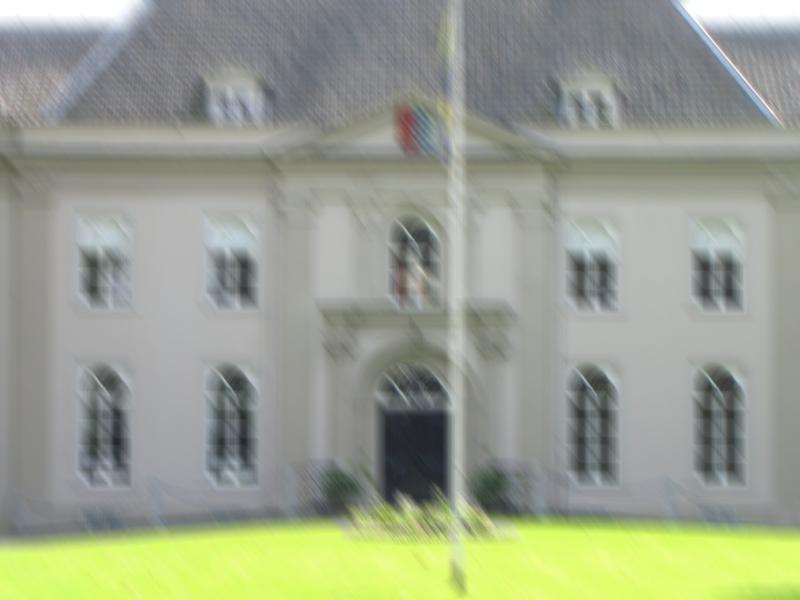}
    \includegraphics[width=0.19\textwidth]{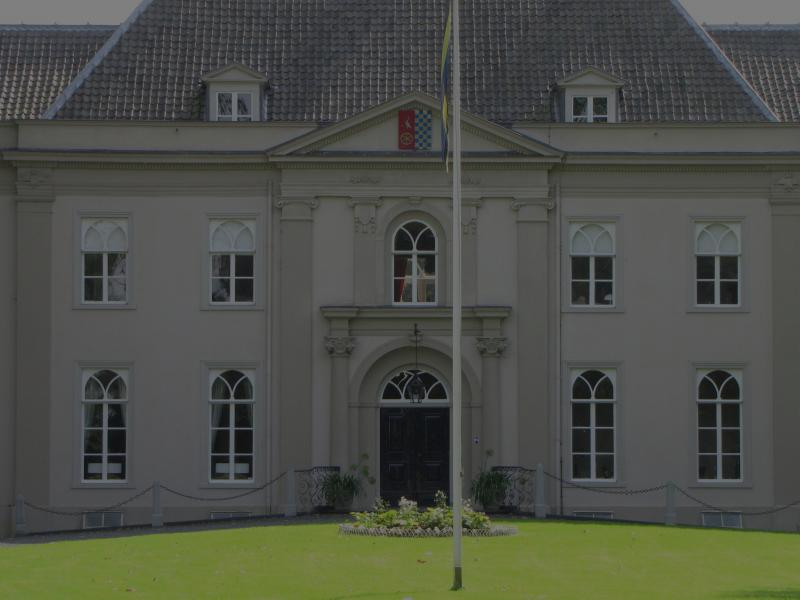}
    \\

    \makebox[0.19\textwidth]{\small $d{=}0.02$, $l{=}35$, $\theta{=}110^{\circ}$,}
    \makebox[0.19\textwidth]{\small min${=}0.4$, max${=}0.9$}
    \makebox[0.19\textwidth]{}
    \makebox[0.19\textwidth]{}
    \makebox[0.19\textwidth]{}
    \makebox[0.19\textwidth]{\small $s{=}1$, and $w{=}1$}
    \makebox[0.19\textwidth]{\small $A{=}200$}
    \makebox[0.19\textwidth]{\small $\sigma{=}50$}
    \makebox[0.19\textwidth]{\small $d{=}35$ and $\Theta{=}135^{\circ}$}
    \makebox[0.19\textwidth]{\small $c{=}0.25$ and $\sigma{=}1.5$}

    \vspace{1mm}

    \includegraphics[width=0.19\textwidth]{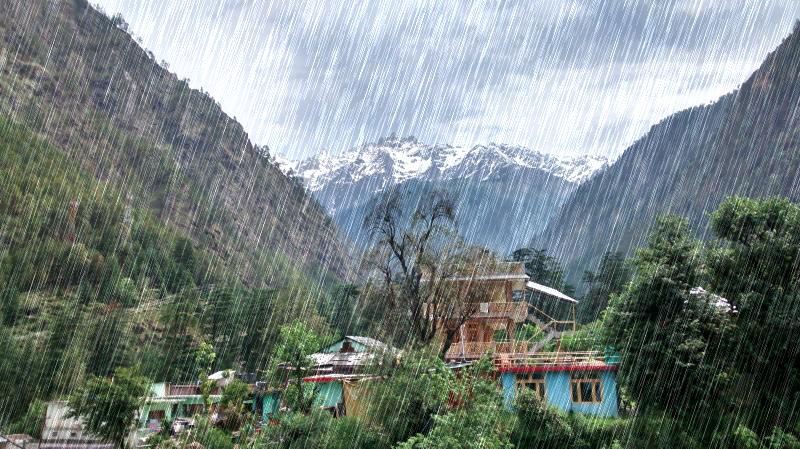}
    \includegraphics[width=0.19\textwidth]{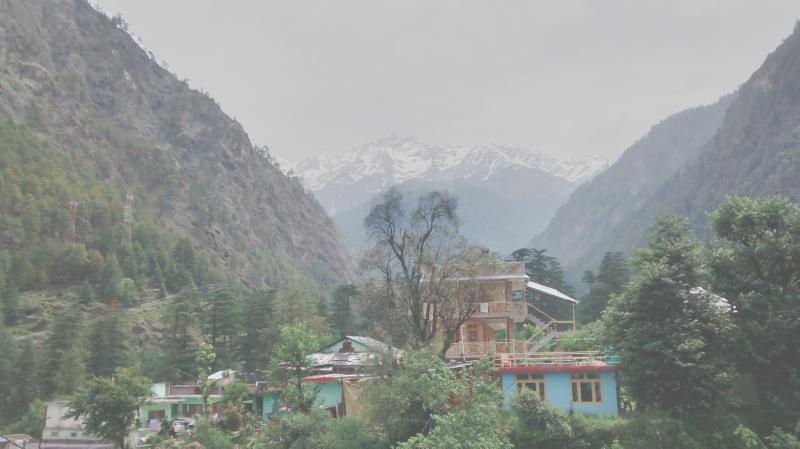}
    \includegraphics[width=0.19\textwidth]{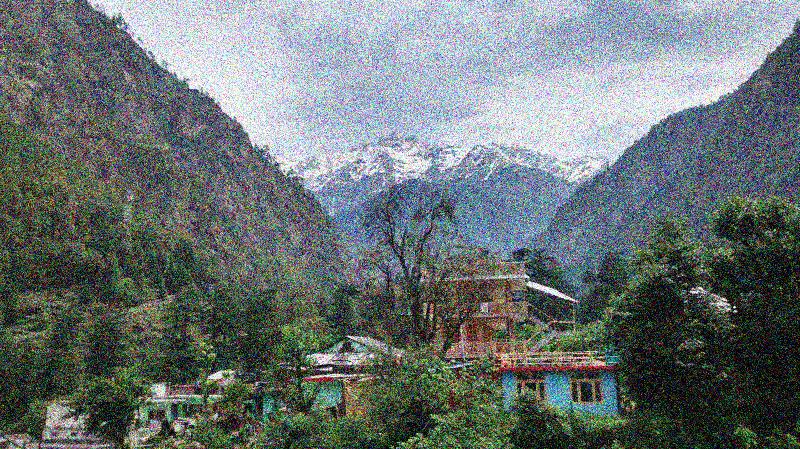}
    \includegraphics[width=0.19\textwidth]{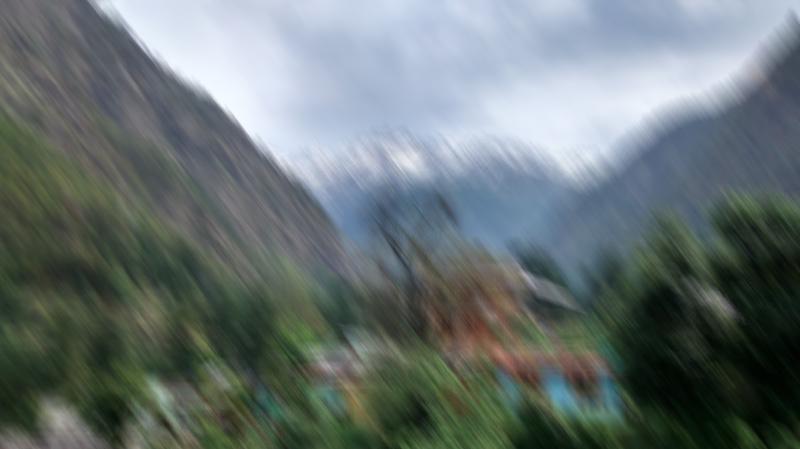}
    \includegraphics[width=0.19\textwidth]{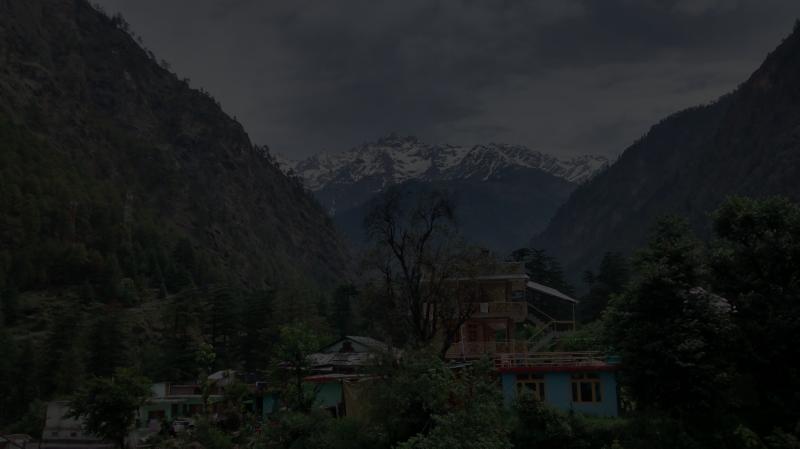}
    \\
    
    \makebox[0.19\textwidth]{Rain}
    \makebox[0.19\textwidth]{Haze}
    \makebox[0.19\textwidth]{Noise}
    \makebox[0.19\textwidth]{Blur}
    \makebox[0.19\textwidth]{Low-Light}
    \caption{Additional examples showing our synthetic degradation generation for rain, haze, noise, blur, and low-light conditions. On top of each image, we report the specific parameters used to produce the corresponding degraded output.}
    \vspace{-2mm}
    \label{fig:synthetics_more}
\end{figure*}

The first phase of our approach involves pre-training the baseline model using natural images with synthetically generated degradations, including rain, haze, noise, blur, and low-light conditions. Unlike standard IR datasets, our pipeline introduces degradations dynamically to clean input images, offering greater flexibility by enabling the model to be trained on the same image with diverse degradation types and varying severity levels. Moreover, each degradation type is parametrized by a set of values that constrain the severity of the distortion. During training, these parameters are randomly selected for each forward pass. Specifically, we utilize 450K images from the Google Landmarks dataset~\cite{weyand2020google} --- those with NIMA score higher than 4.90 and with short-edge resolution larger than 400, providing a diverse set of real scenes. \Cref{fig:synthetics_more} show examples for the different degradations under different parameter settings. In what follows, we detail the generation process for each type of synthetic degradation.

\paragraph{Rain:}~Capturing paired degraded and clean images with rain is inherently challenging, as environmental conditions often vary when capturing the same scene under rainy and clear weather. Consequently, deraining datasets typically simulate rain by creating a set of predefined rain stroke masks, which are then added to input images. However, these datasets often include a limited number of masks~\cite{yang2017deep}, leading to potential overfitting on the specific patterns present in the dataset. In our case, aiming to derive a more general set of masks, we consider five adjustable parameters: density, length, angle, drop size, and blending weight.

The process for generating a rainy image $I_{rain}$ begins by creating an empty mask $M$ of the same size as the input image $I_{input}$, which will hold the rain-stroke patterns. The density parameter $d$ determines the number of raindrops, computed as a fraction of the total image pixels. Random coordinates for the starting positions of the drops are generated, ensuring they stay within bounds to accommodate the specified drop size $s$. These coordinates are then used to populate $M$ with raindrops. Next, a motion blur kernel $K$ is constructed to simulate the appearance of rain streaks, based on the specified rain length $l$ and angle $\theta$. 
The mask $M$ is convolved with $K$, creating streaks that mimic natural rain patterns. Finally, the rain streaks are normalized and expanded into three RGB channels to match the input image dimensions. These streaks are blended with the original image $I_{input}$ using a weight $w$, resulting in the final rain-augmented image $I_{\text{rain}}$. All the input parameters are randomly selected for each forward pass. Mathematically ---omitting normalizations for clarity,
\begin{equation} \label{eq:rain}
    I_{rain} = w I_{input} + (1-w) (K * M).
\end{equation}
The value ranges are rain density $d$:$[0.005, 0.02]$, rain length $l$: $[25, 35]$, rain angle $\theta$: $[70, 110]$, raindrop size $s$: $[1, 3]$, and mask weight $[0.75, 1]$. 

\paragraph{Haze:}~As with rainy images, capturing paired degraded and clean images under hazy conditions is nearly impossible due to the variability of environmental factors. Inspired by prior works~\cite{li2019benchmarking,galdran2016EVID}, we model haze degradation using the Kochsmieder model~\cite{harald1924theorie}, which describes how the visibility of distant objects diminishes, vanishing into the horizon as a function of their distance from the observer. We can formulate it as:

\begin{equation} \label{eq:haze}
    I_{haze} = T \cdot I_{input} + (1-T) \cdot A,
\end{equation}
where $I_{input}$ is the input clean image, $T$ is the transmission map derived from the estimated depth map, $A$ is the haze color, and $I_{haze}$ is the resulting image with synthetic haze.

We estimate the depth map of the input image using DepthAnythingv2~\cite{yang2024depth}. This transmission map is normalized within the predefined minimum and maximum haze values to generate the transmission map $T$. The depth map is then replicated for each color channel to compute \cref{eq:haze}. To ensure variability, all parameters are randomly sampled for each forward pass. The parameter ranges are as follows:  minimum haze $[0.2, 0.4]$, and maximum haze $[0.7, 0.9]$, haze color is unique for the three color channels with values ranging $[140, 200]$.

\paragraph{Blur:}~Blur in images can arise from various sources, such as motion blur, out-of-focus blur, and lens blur, among others. In this work, we focus on simulating the motion blur due to its relevance in practical applications. The blur effect is introduced by convolving the input image $I_{input}$ with a parametrized kernel $K$, designed to model motion blur along a specific direction. The kernel is defined by its size in pixels ($d$) and its angle ($\Theta$), to simulate the directional streaks characteristics of motion blur. Mathematically,

\begin{equation}
    I_{blur} = K_{\Theta,d} * I_{input}.
\end{equation}
The kernel size ($d$) is selected from odd values within the range $[9, 35]$, while the angle ($\Theta$) specifies the orientation of the blur in degrees, ranging from $[0, 360]$.

\paragraph{Noise:}~We use the standard Additive White Gaussian Noise (AWGN) approach for noise. In short, we add to the original image a second image that follows a Gaussian distribution with mean $0$ and variance $\sigma$. Mathematically,

\begin{equation}
    I_{noise}=I_{input}+\mathcal{N}(0,\sigma).
\end{equation}

\begin{table*}[ht!]
    \centering
    \caption{Ablation study on different low-rank adapters and their rank. Results are the mean for all images. LoRA outperforms both VeRA and Conv-LORA. Lower ranks perform better.}
    \small
    \setlength{\tabcolsep}{5.3pt} 
    \begin{tabular}{lcccccccccccccc}
        
        \toprule
        \multicolumn{2}{c}{PSNR/SSIM} & \multicolumn{2}{c}{Deraining} & \multicolumn{2}{c}{Dehazing} & \multicolumn{2}{c}{Denoising} & \multicolumn{2}{c}{Deblurring} & \multicolumn{2}{c}{Low-Light} & \multicolumn{2}{c}{\multirow{2}{*}{Average}} & 
        \multirow{2}{*}{\shortstack{Adapter\\Param.}}\\
        
        \cmidrule{3-12}

        \multicolumn{1}{c}{Method} & Rank & \multicolumn{2}{c}{Rain100L} & \multicolumn{2}{c}{SOTS (Out)} & \multicolumn{2}{c}{BSD68 $_{\sigma=25}$} & \multicolumn{2}{c}{GoPro} & \multicolumn{2}{c}{LoLv1} & & & \\
        
        \midrule

        \multirow{3}{*}{LoRA~\cite{hu2021lora}} & 4 & 37.79 & .979 & 33.48 & .984 & 31.38 & .898 & 29.00 & .878 & 24.19 & .865 & 31.17 & .921 & 3.6M\\
        
        & 8 & 37.75 & .978 & 33.4 & .982 & 31.39 & .898 & 29.02 & .878 & 24.18 & .865 & 31.15 & .920 & 7.2M\\
        
        & 16 & 37.61 & .972 & 33.21 & .977 & 31.31 & .896 & 28.77 & .875 & 23.96 & .862 & 30.97 & .916 & 14.3M\\   

        \midrule

        \multirow{3}{*}{VeRA~\cite{kopiczko2024vera}} & 4 & 37.02 & .971 & 32.67 & .972 & 31.32 & .896 & 28.61 & .872 & 23.78 & .580 & 30.68 & .858 & 460K\\
        & 8 & 37.09 & .971 & 32.69 & .972 & 31.32 & .896 & 28.64 & .873 & 23.79 & .580 & 30.71 & .858 & 468K\\ 
        & 16 & 37.04 & .970 & 32.62 & .970 & 31.33 & .896 & 28.62 & .872 & 23.84 & .581 & 30.69 & .858 & 476K\\ 

        \midrule
        
        \multirow{2}{*}{Conv-LoRA~\cite{zhong2024convlora}} & 4 & 37.00 & .969 & 32.55 & .971 & 31.32 & .896 & 28.54 & .870 & 23.70 & .576 & 30.62 & .856 & 3.9M\\
        & 8 & 36.94 & .968 & 32.44 & .968 & 31.30 & .895 & 28.48 & .868 & 23.62 & .575 & 30.56 & .855 & 7.5M\\ 

        \bottomrule
    \end{tabular}
    \label{tab:adapters_ablation}
\end{table*}

\begin{table*}[ht!]
    \centering
    \caption{Ablation study on different methods for blending the five degradations task-specific LoRA~\cite{hu2021lora} adapters.}
    \vspace{-2mm}
    \small
    \setlength{\tabcolsep}{5.3pt} 
    \begin{tabular}{lcccccccccccc}
        
        \toprule
        \multicolumn{1}{c}{PSNR/SSIM} & \multicolumn{2}{c}{Deraining} & \multicolumn{2}{c}{Dehazing} & \multicolumn{2}{c}{Denoising} & \multicolumn{2}{c}{Deblurring} & \multicolumn{2}{c}{Low-Light} & \multicolumn{2}{c}{\multirow{2}{*}{Average}}\\
        
        \cmidrule{2-11}
    
        \multicolumn{1}{c}{Method}  & \multicolumn{2}{c}{Rain100L} & \multicolumn{2}{c}{SOTS (Out)} & \multicolumn{2}{c}{BSD68 $_{\sigma=25}$} & \multicolumn{2}{c}{GoPro} & \multicolumn{2}{c}{LoLv1} & & \\

        \midrule

        Sum & 19.50 & .755 & 18.80 & .745 & 18.10 & .730 & 18.25 & .728 & 17.70 & .715 & 18.67 & .736 \\
        Average & 30.54 & .939 & 20.87 & .855 & 28.98 & .785 & 21.34 & .792 & 15.49 & .673 & 23.84 & .809 \\
        
        \midrule

        Ours OH & 38.18 & .983 & 33.46 & .983 & 31.38 & .898 & 29.00 & .878 & 24.20 & .865 & 31.24 & .921 \\
        Ours SW & 38.22 & .984 & 33.48 & .984 & 31.38 & .898 & 29.00 & .878 & 24.19 & .865 & 31.25 & .922 \\

        \midrule
        Ours Oracle & 39.09 & .981 & 33.54 & .984 & 31.40 & .901 & 29.10 & .879 & 24.45 & .866 & 31.39 & .913\\

        \bottomrule
    \end{tabular}
    \vspace{-2mm}
    \label{tab:blending_methods}
\end{table*}

\paragraph{Low-Light:}~When an image is captured under low-light conditions, cameras amplify the sensor's signal to capture details, amplifying noise, and leading to grainy or speckled artifacts. Furthermore, the reduced dynamic range limits the ability to capture the range of intensities, resulting in color inconsistencies, and loss of detail in shadows and highlights. To simulate these conditions, we compress the input image histogram by a factor $c$ and add a noise with a small $\sigma$ value. This process can be expressed mathematically as:

\begin{equation}
    I_{lol} = I_{input} \cdot c + \mathcal{N}(0,\sigma).
\end{equation}
The compression factor $c$ ranges in the interval $[0.25, 0.5]$ and the $\sigma$ takes values in the interval $[0.5, 1.5]$.

\section{Additional Phase II Ablation Studies}
The second phase of our approach involves training a set of adapters --- LoRA~\cite{hu2021lora} in the main manuscript --- for each type of degradation. In this section, we extend our analysis to include other low-rank decompositions and their rank. Specifically, we evaluate VeRA~\cite{kopiczko2024vera} and Conv-LoRA~\cite{zhong2024convlora}.
The results for the five-degradation setup across the three adapter types are presented in~\cref{tab:adapters_ablation}, which also includes the number of trainable parameters for each adapter in the last column. For a fair comparison, we utilize the same baseline model weights and estimator, while training only the task-specific adapters. Among the methods, LoRA~\cite{hu2021lora} achieves the best overall performance. However, VeRA~\cite{kopiczko2024vera} provides competitive results with significantly fewer parameters, as it estimates only two vectors per layer instead of two low-rank matrices. On the other hand, Conv-LoRA~\cite{zhong2024convlora} performs worse despite having more parameters, owing to its Mixture-of-Experts approach with convolution layers in the decomposed space. Regarding the decomposition rank, we find that ranks of 4 and 8 consistently outperform a rank of 16 in terms of both accuracy and parameter efficiency. 

\section{Additional Phase III Analysis}
In the third phase of our method, we train an estimator to integrate the set of adapters with the baseline model based on the input image. The estimator architecture comprises four blocks of Conv2D layers, each followed by batch normalization, ReLU activation, and max pooling, culminating with a global average pooling layer and a linear projection. The total parameter count for the estimator is 538K.  

\begin{figure*}[htbp]
    \centering
    \includegraphics[width=\textwidth]{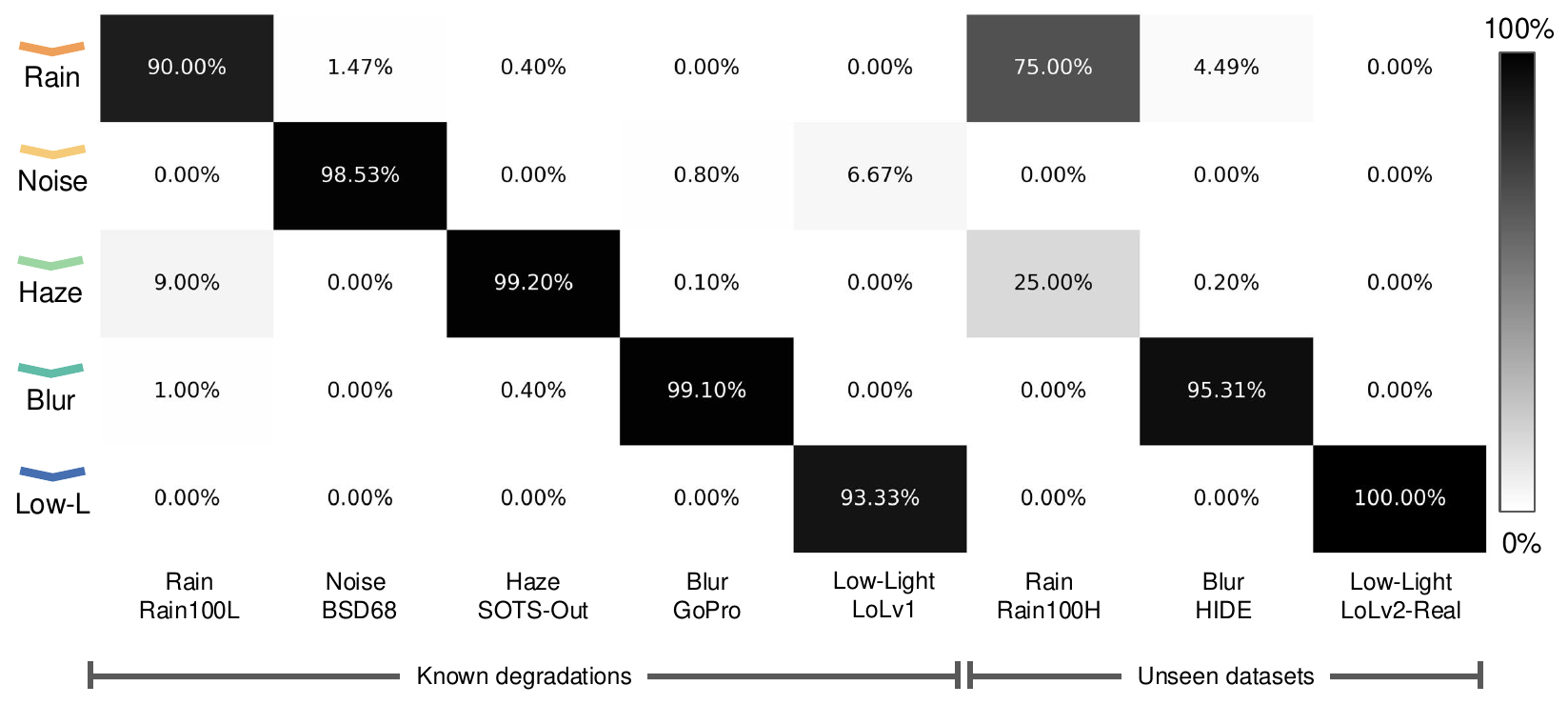}
    \caption{Confusion matrix for our estimator in test images. First five columns: known degradations with the datasets considered at training. Last three columns: unknown datasets for rain, blur, and low light. Specifically, from left-to-right columns, the datasets are: Rain100L~\cite{yang2017deep}, BSD68~\cite{martin2001database}, SOTS (outdoors)~\cite{li2019benchmarking}, GoPro~\cite{Nah2017gopro}, LoLv1~\cite{wei2018loldataset}, Rain100H~\cite{yang2017deep}, HIDE~\cite{shen2019hide}, and LoLv2-Real~\cite{Yang2020lolv2dataset}.}
    \label{fig:conf_matrix}
\end{figure*}

We present a confusion matrix in \cref{fig:conf_matrix}, showing the One-Hot predictions for known degradations ---both seen and unseen datasets. The estimator predicts known degradations with probabilities exceeding 90\% in most cases. However, for one of the unseen datasets, the Rain100H dataset~\cite{yang2017deep}, the prediction probability falls below 90\%, likely due to the severity of the rain streaks and their resemblance to haze-related degradations. Notably, even when the estimator selects an incorrect adapter, no significant artifacts or undesired effects are introduced, owing to the robustness of the large-scale pre-training. 


To highlight the importance of our estimator, we perform another ablation comparing our approach to just adding or averaging the five degradation task-specific adapters. \cref{tab:blending_methods} shows the results of this ablation. When all adapters are added, the $\Delta W$ values become excessively large for the baseline model to effectively handle them. On the other hand, averaging the adapters results in suboptimal restored images, as the model fails to specialize in any particular degradation type. Finally, Table~\ref{tab:blending_methods} also presents the results for the five-task setup assuming the lightweight estimator is perfect, i.e., an {\it oracle}. The results are only slightly better from those obtained when using our estimator, confirming that its performance is strong. Please note that these {\it Oracle} results can be interpreted as single-task performance, as the selected adapter always corresponds to the specific degradation type of the input image.

\section{Additional Qualitative Results}
\begin{figure*}[t] \centering
    \raisebox{0.35\height}{\makebox[0.02\textwidth]{\rotatebox{90}{\makecell{\small Rain~\cite{yang2017deep}}}}}
    \includegraphics[width=0.19\textwidth]{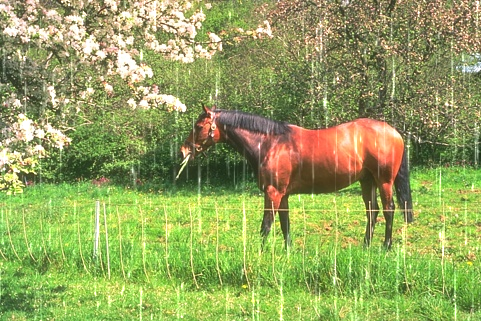}
    \includegraphics[width=0.19\textwidth]{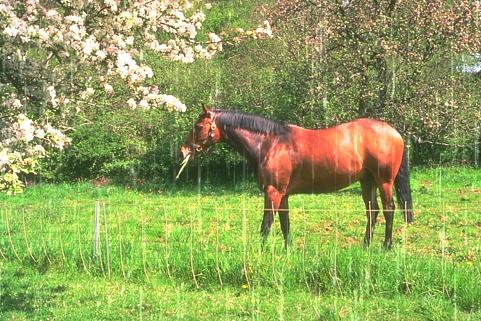}
    \includegraphics[width=0.19\textwidth]{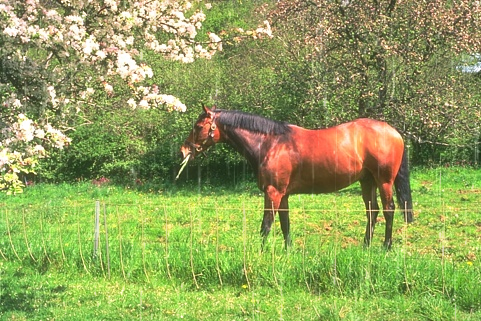}
    \includegraphics[width=0.19\textwidth]{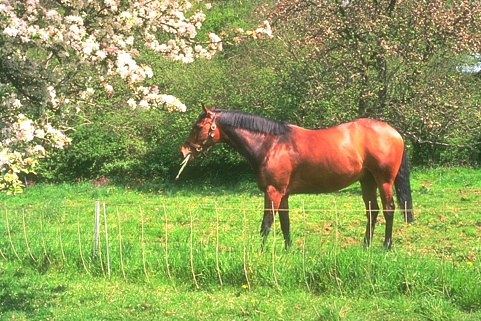}
    \includegraphics[width=0.19\textwidth]{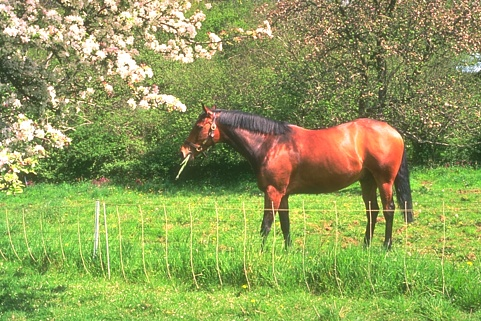}
    \\

    \vspace{0.5mm}
    
    \raisebox{1.5\height}{\makebox[0.02\textwidth]{\rotatebox{90}{\makecell{\small Haze~\cite{li2019benchmarking}}}}}
    \includegraphics[width=0.19\textwidth]{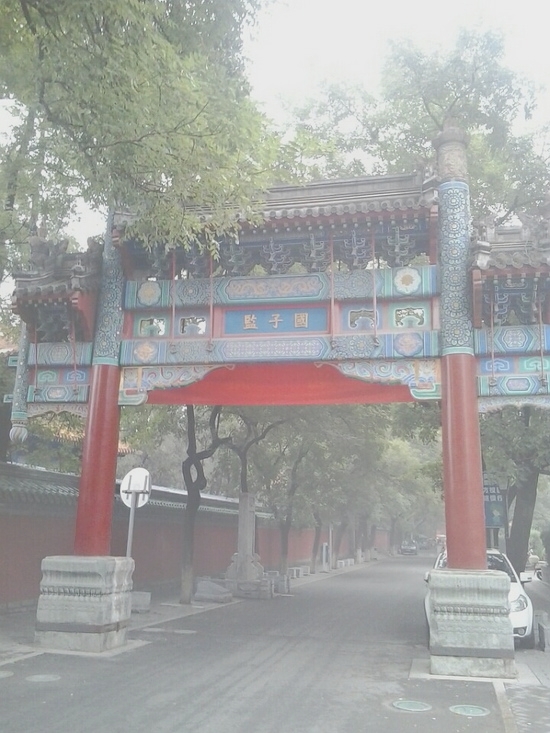}
    \includegraphics[width=0.19\textwidth]{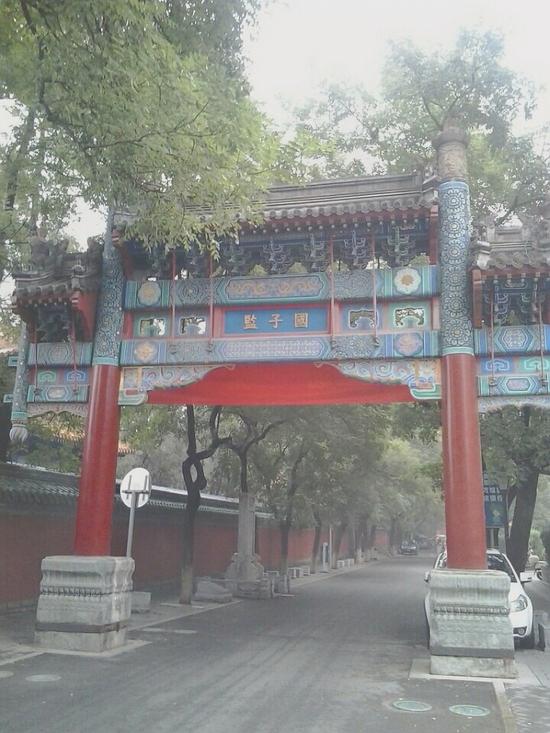}
    \includegraphics[width=0.19\textwidth]{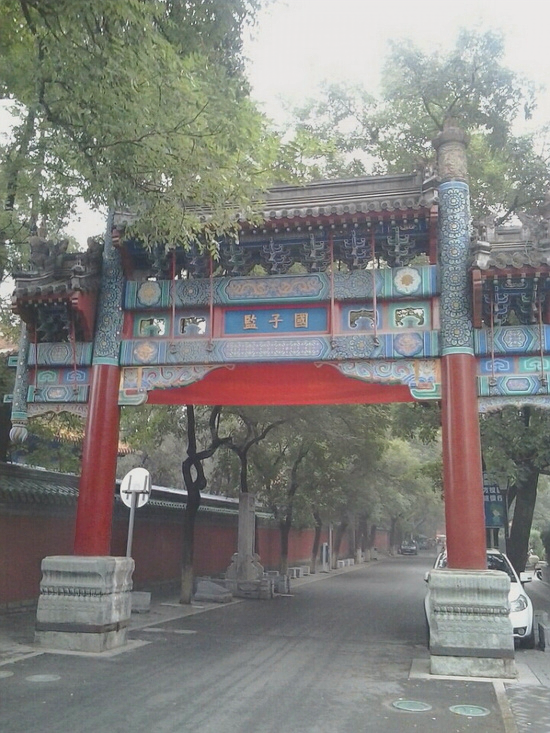}
    \includegraphics[width=0.19\textwidth]{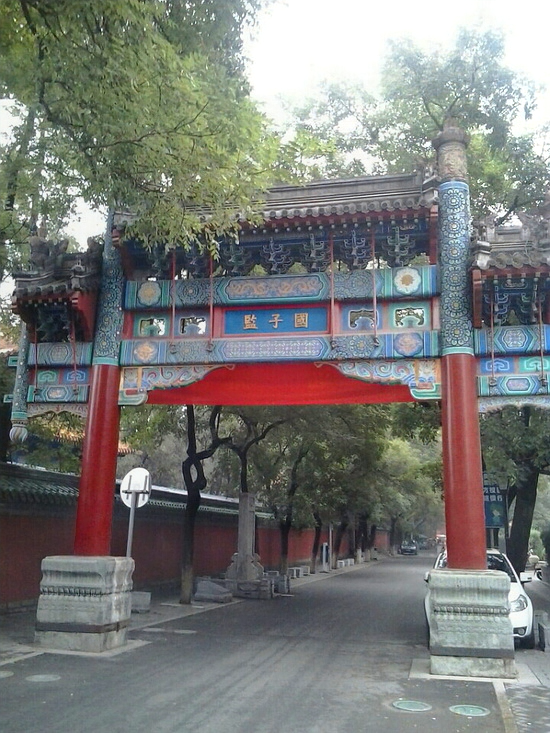}
    \includegraphics[width=0.19\textwidth]{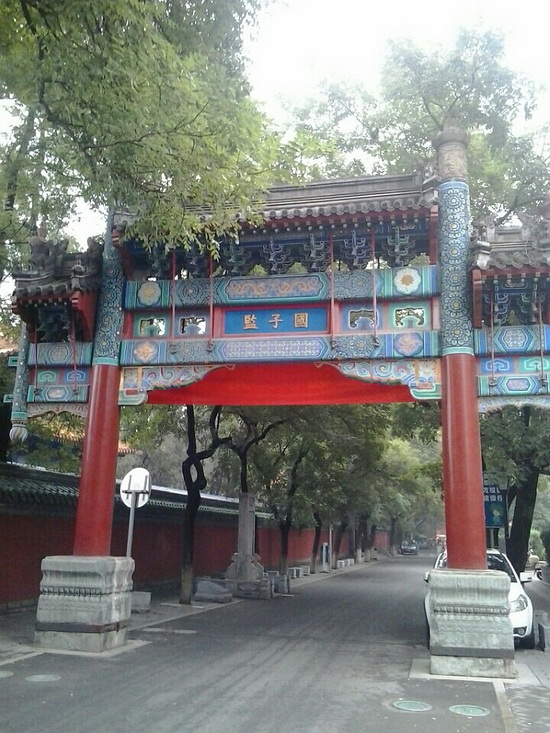}
    \\

    \vspace{0.5mm}

    \raisebox{0.3\height}{\makebox[0.02\textwidth]{\rotatebox{90}{\makecell{\small Noise~\cite{Nah2017gopro}}}}}
    \includegraphics[width=0.19\textwidth]{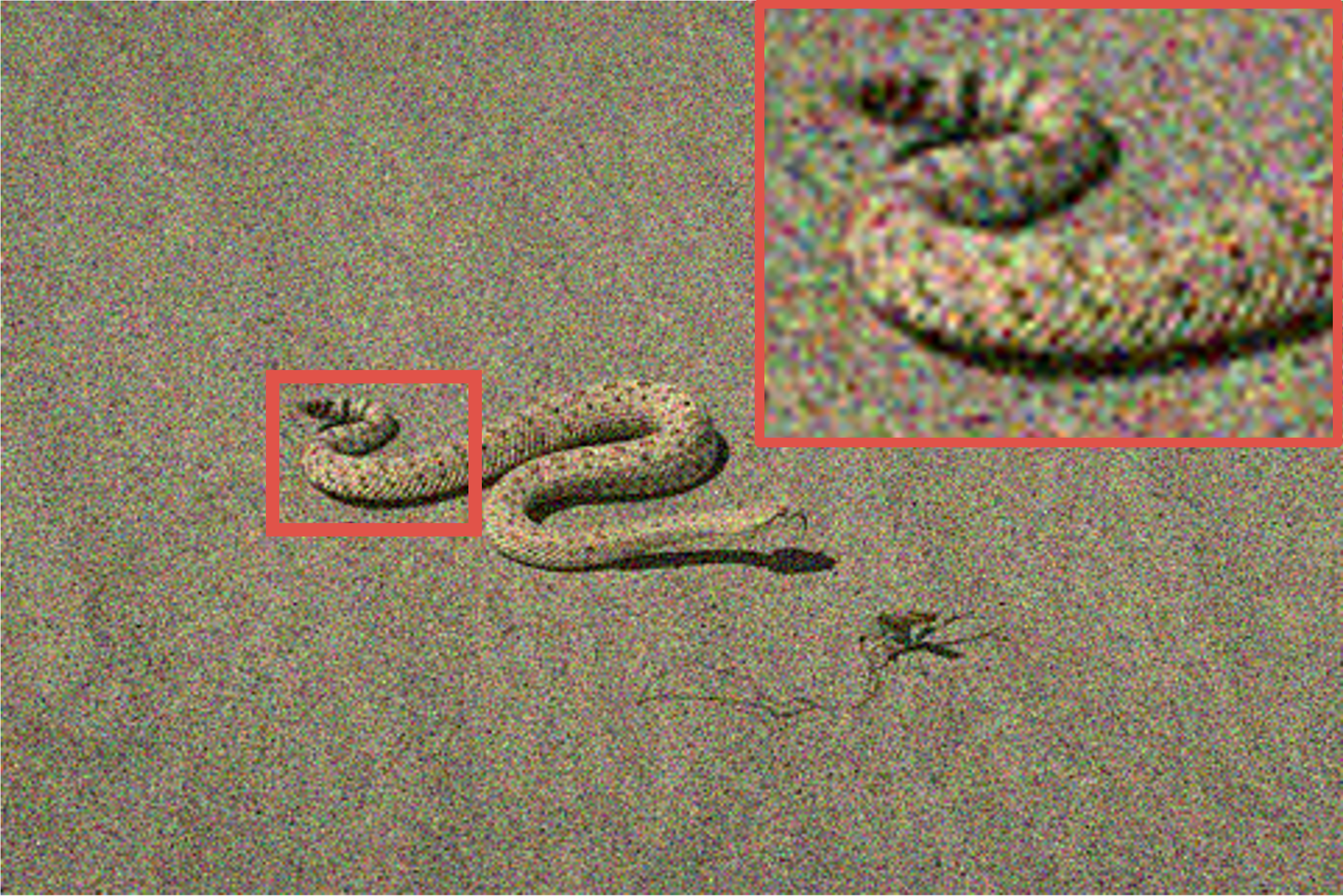}
    \includegraphics[width=0.19\textwidth]{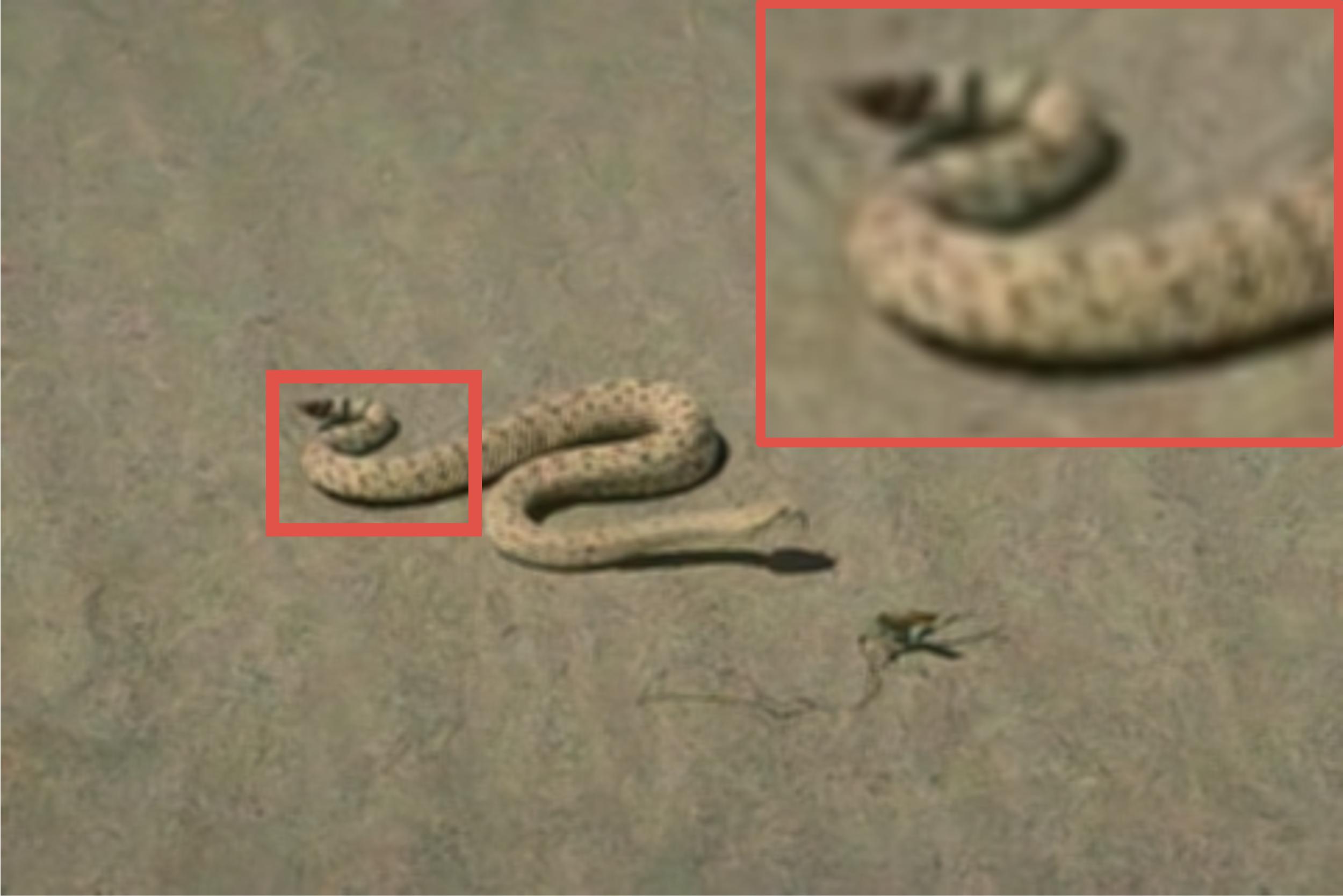}
    \includegraphics[width=0.19\textwidth]{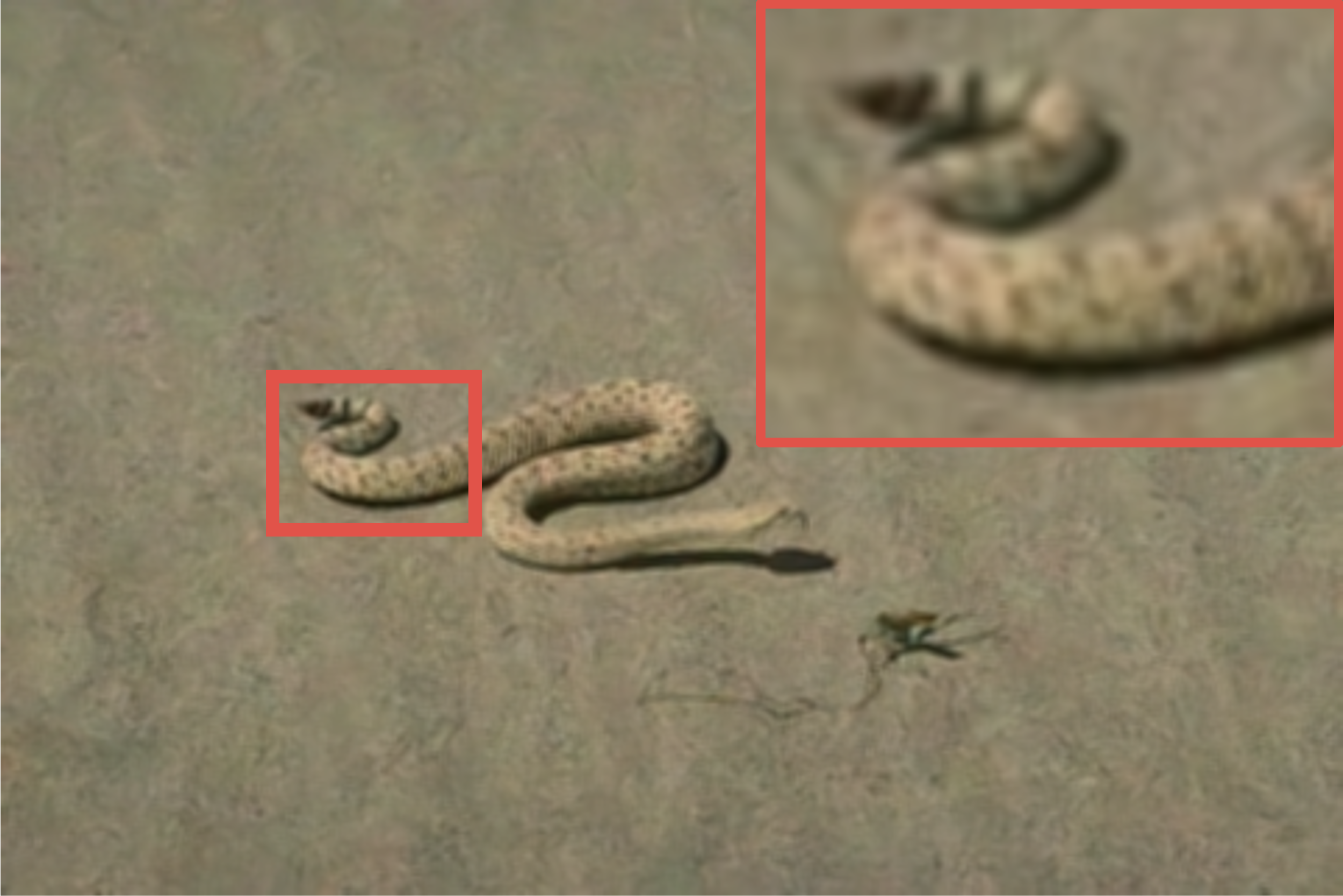}
    \includegraphics[width=0.19\textwidth]{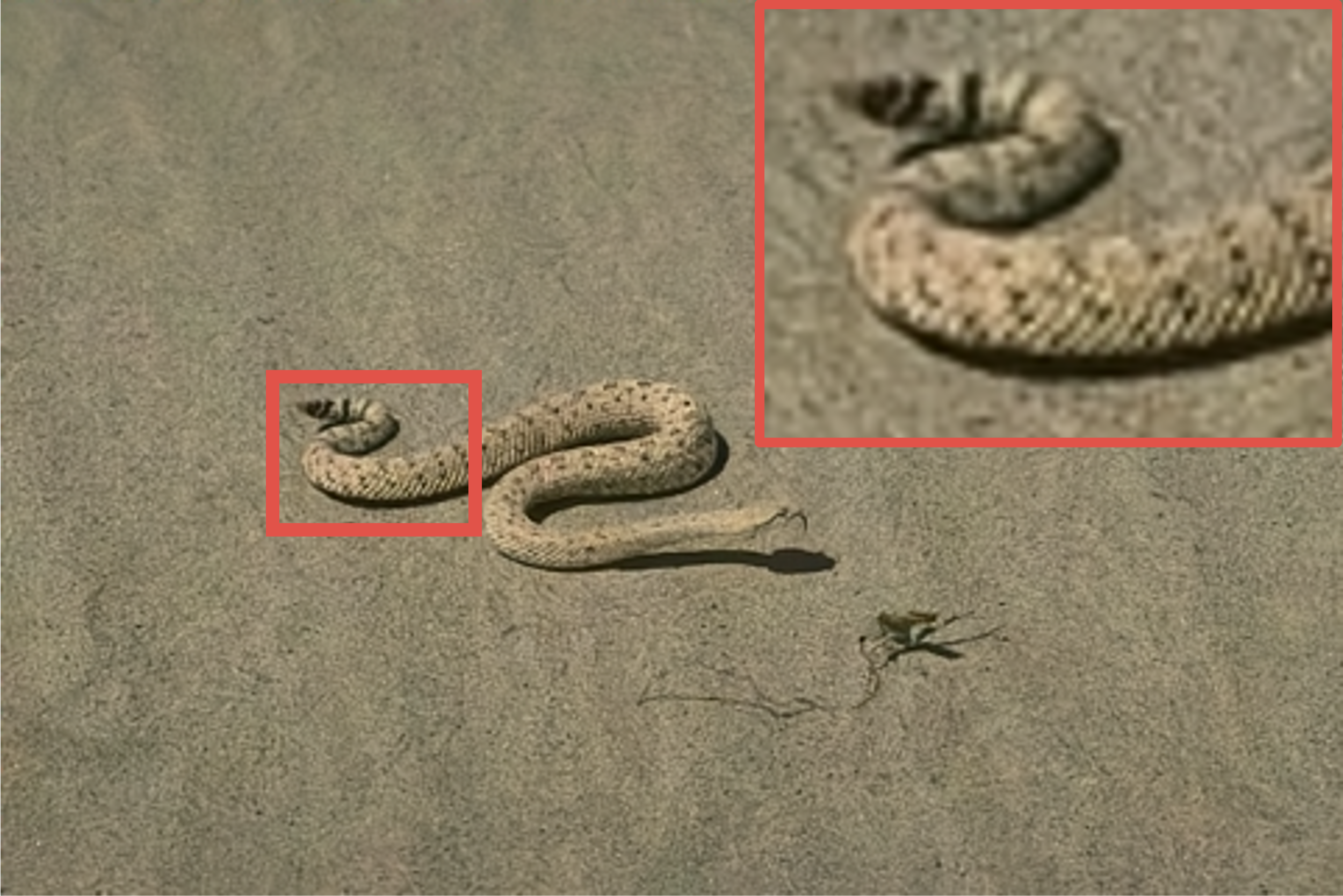}
    \includegraphics[width=0.19\textwidth]{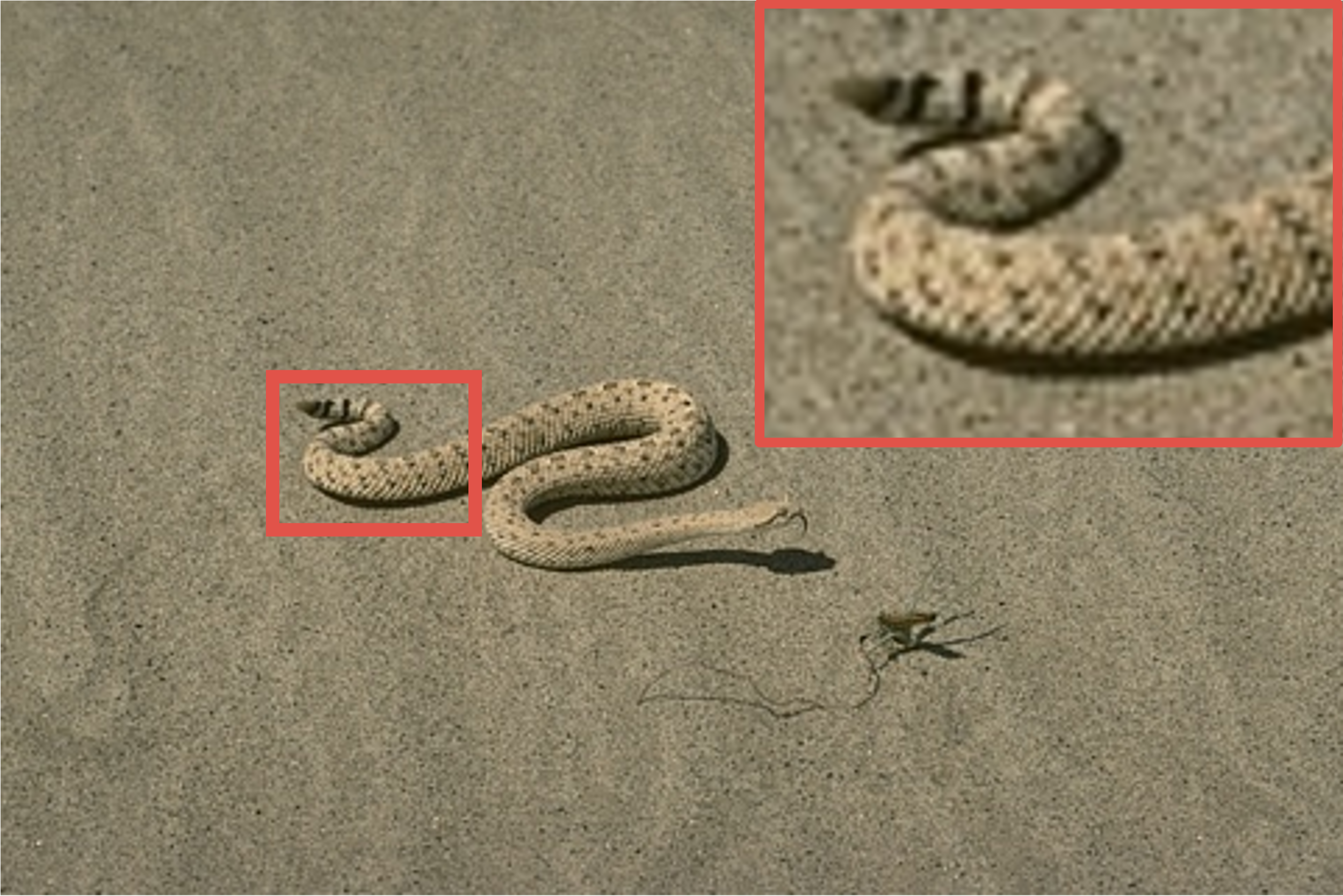}
    \\

    \vspace{0.5mm}

    \raisebox{0.2\height}{\makebox[0.02\textwidth]{\rotatebox{90}{\makecell{\small Blur~\cite{Nah2017gopro}}}}}
    \includegraphics[width=0.19\textwidth]{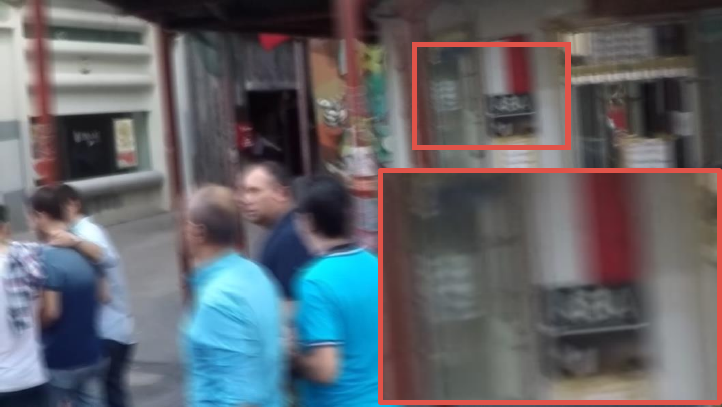}
    \includegraphics[width=0.19\textwidth]{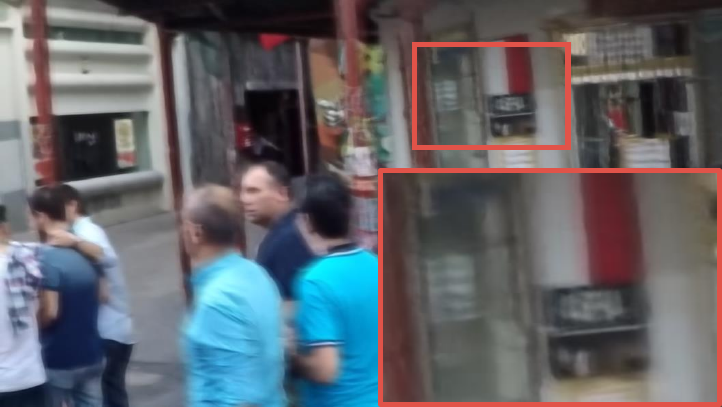}
    \includegraphics[width=0.19\textwidth]{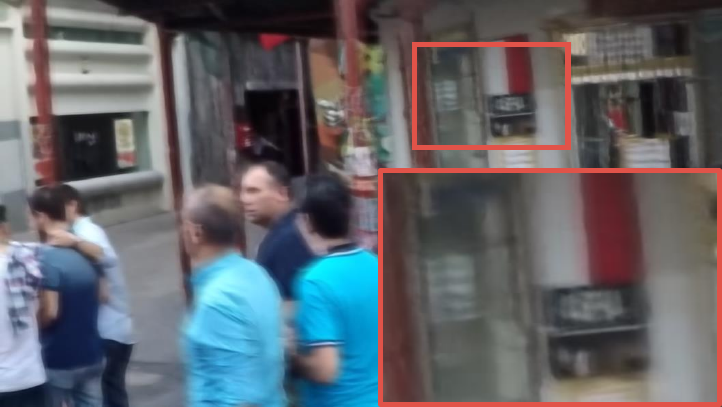}
    \includegraphics[width=0.19\textwidth]{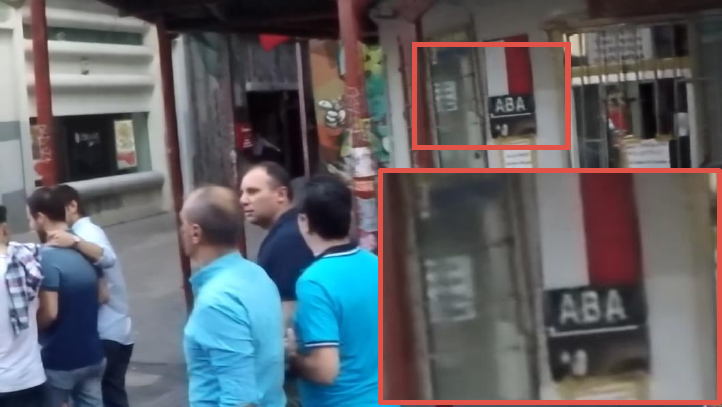}
    \includegraphics[width=0.19\textwidth]{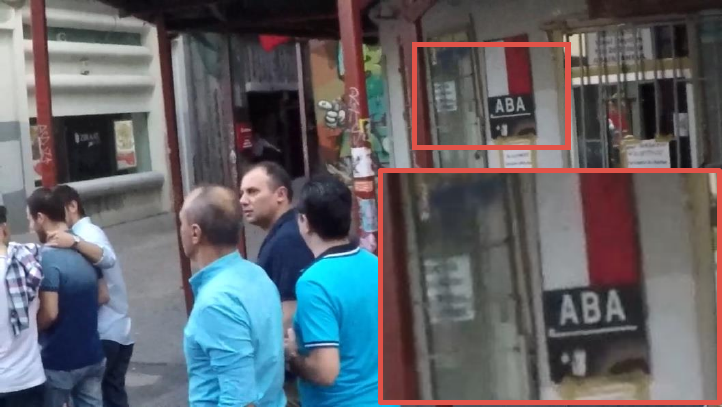}
    \\
    
    \vspace{0.5mm}
    
    \raisebox{0.05\height}{\makebox[0.02\textwidth]{\rotatebox{90}{\makecell{\small Low-Light~\cite{wei2018loldataset}}}}}
    \includegraphics[width=0.19\textwidth]{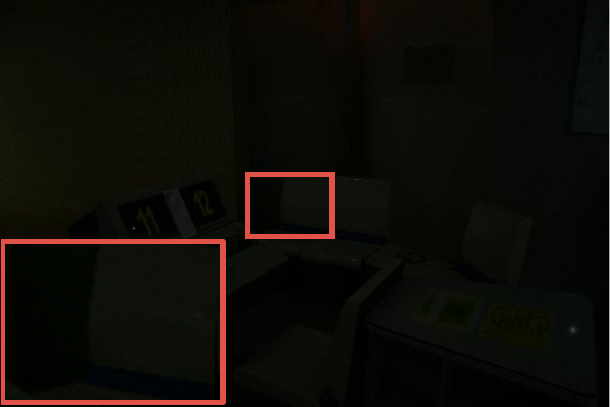}
    \includegraphics[width=0.19\textwidth]{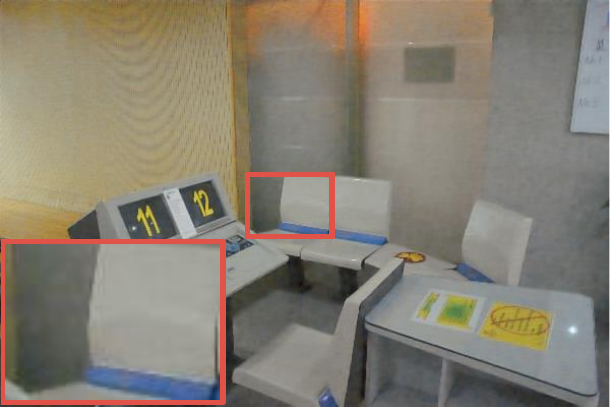}
    \includegraphics[width=0.19\textwidth]{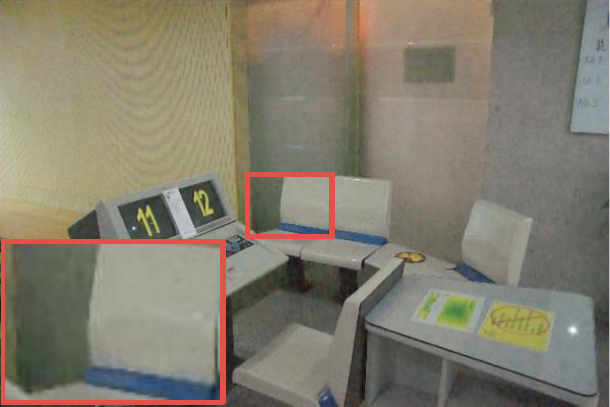}
    \includegraphics[width=0.19\textwidth]{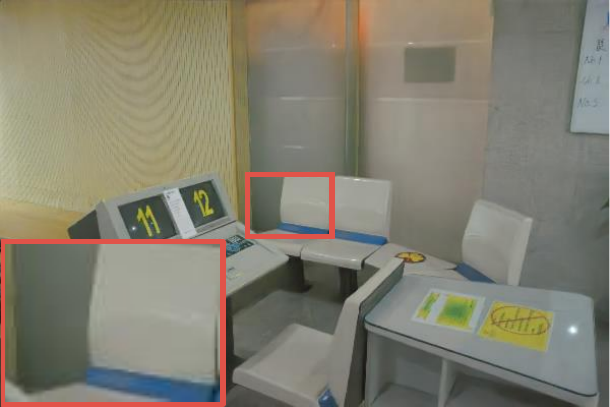}
    \includegraphics[width=0.19\textwidth]{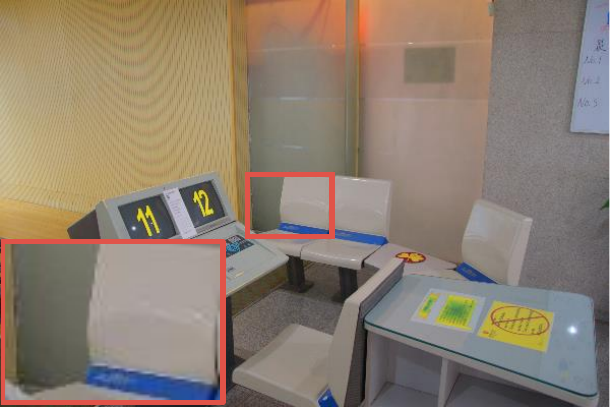}
    \\

    \vspace{0.5mm}

    \raisebox{0.8\height}{\makebox[0.02\textwidth]{\rotatebox{90}{\makecell{\small Rain100H~\cite{yang2017deep}}}}}
    \includegraphics[width=0.19\textwidth]{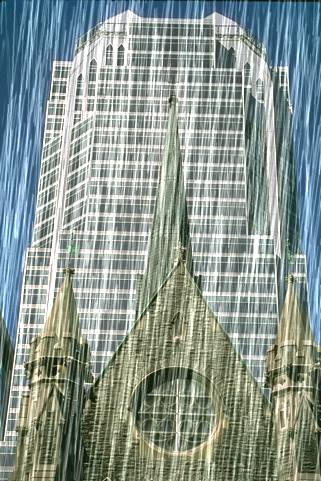}
    \includegraphics[width=0.19\textwidth]{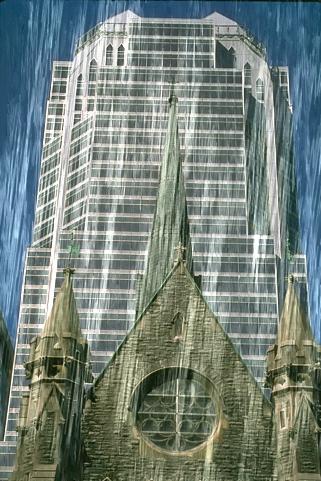}
    \includegraphics[width=0.19\textwidth]{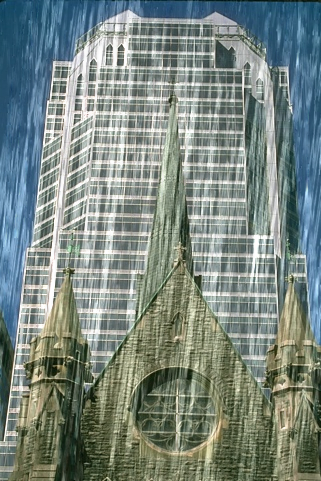}
    \includegraphics[width=0.19\textwidth]{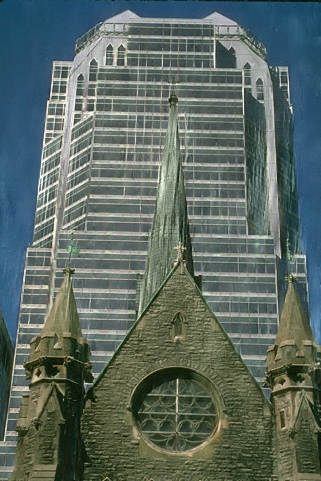}
    \includegraphics[width=0.19\textwidth]{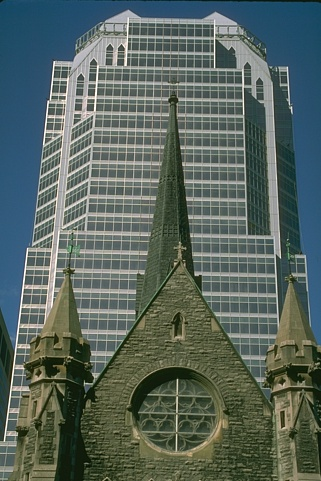}
    \\

    \vspace{0.5mm}
    
    \raisebox{0.01\height}{\makebox[0.02\textwidth]{\rotatebox{90}{\makecell{\small LoLv2-Real~\cite{yang2021sparse}}}}}
    \includegraphics[width=0.19\textwidth]{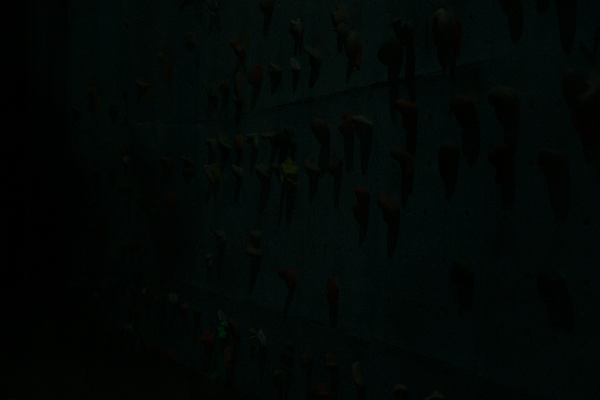}
    \includegraphics[width=0.19\textwidth]{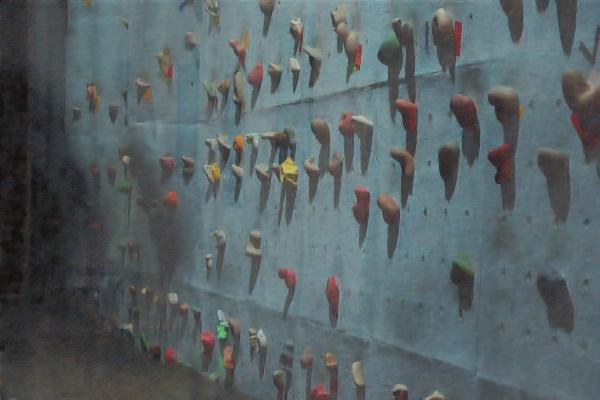}
    \includegraphics[width=0.19\textwidth]{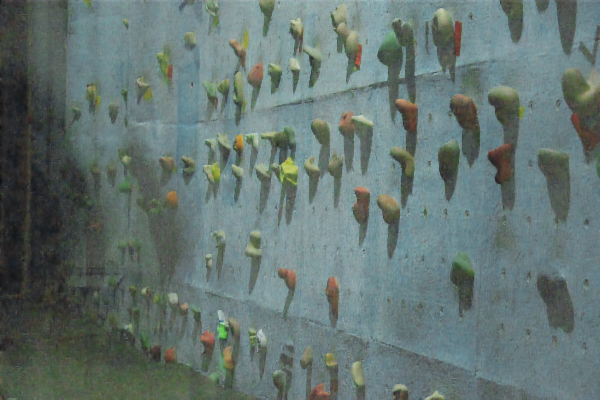}
    \includegraphics[width=0.19\textwidth]{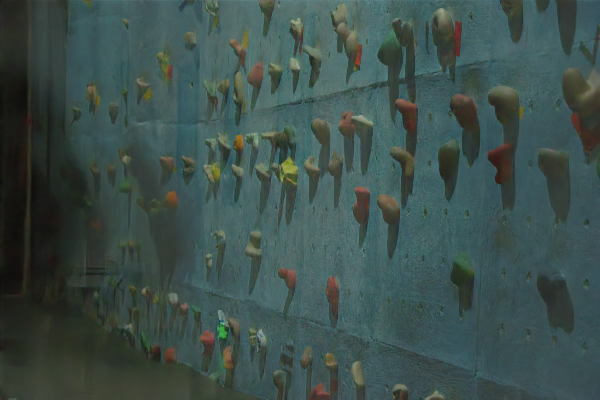}
    \includegraphics[width=0.19\textwidth]{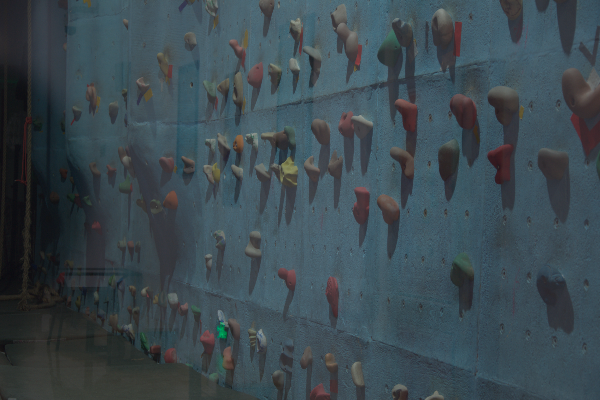}
    \\
    \makebox[0.02\textwidth]{}
    \makebox[0.19\textwidth]{\small Input}
    \makebox[0.19\textwidth]{\small Restormer~\cite{zamir2022restormer}}
    \makebox[0.19\textwidth]{\small PromptIR~\cite{potlapalli2024promptir}}
    \makebox[0.19\textwidth]{\small Ours-OH}
    \makebox[0.19\textwidth]{\small Ground truth}
    \caption{Qualitative results for known degradation removal, both seen datasets ---Rows 1 to 5--- and unseen datasets ---Rows 6 to 8.} 
    \label{fig:qualitative_known}
\end{figure*}

\begin{figure*}[t] \centering

    \raisebox{1.2\height}{\makebox[0.02\textwidth]{\rotatebox{90}{\makecell{\small JPEG~\cite{sheikh2006statistical}}}}}
    \includegraphics[width=0.97\textwidth]{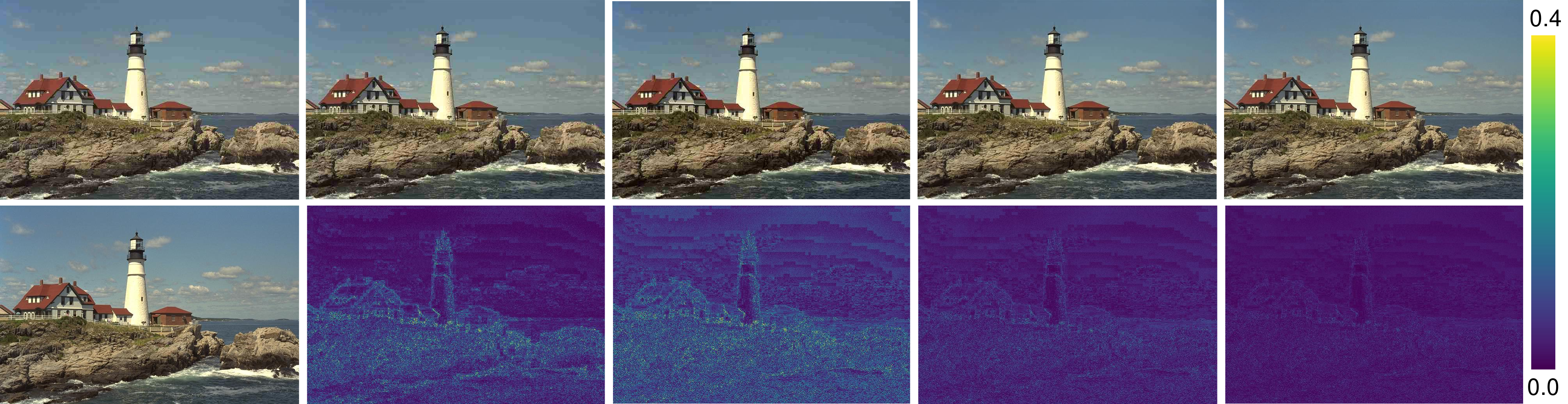}

    \vspace{2mm}
    \raisebox{.8\height}{\makebox[0.02\textwidth]{\rotatebox{90}{\makecell{\small Snow~\cite{zhang2021snowcityscapes}}}}}
    \includegraphics[width=.97\textwidth]{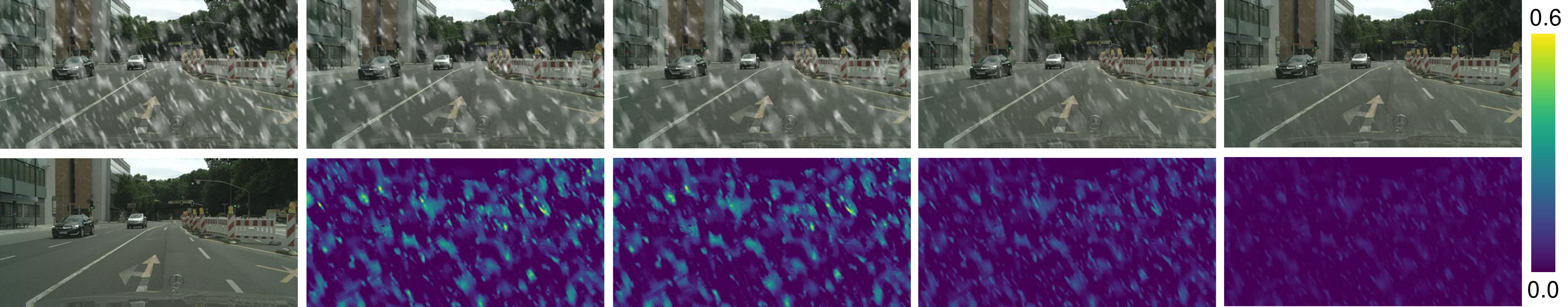}
    
    \makebox[-0.005\textwidth]{}
    \makebox[0.1825\textwidth]{\small Input/Ground truth}
    \makebox[0.1825\textwidth]{\small Restormer~\cite{zamir2022restormer}}
    \makebox[0.1825\textwidth]{\small PromptIR~\cite{potlapalli2024promptir}}
    \makebox[0.1825\textwidth]{\small Ours-SW}
    \makebox[0.1825\textwidth]{\small Ours-SW retrained}

    \caption{Qualitative results on JPEG artifact removal on Live1 dataset~\cite{sheikh2006statistical} and desnowing on CityScapes-Snow-Medium~\cite{zhang2021snowcityscapes}. We show the mean absolute error map below each image.} 
    \label{fig:qualitative_unknown}
\end{figure*}

\begin{figure*}[t] \centering

    \raisebox{0.01\height}{\makebox[0.02\textwidth]{\rotatebox{90}{\makecell{\small Blur\&JPEG~\cite{nah2019reds}}}}}
    \includegraphics[width=0.19\textwidth]{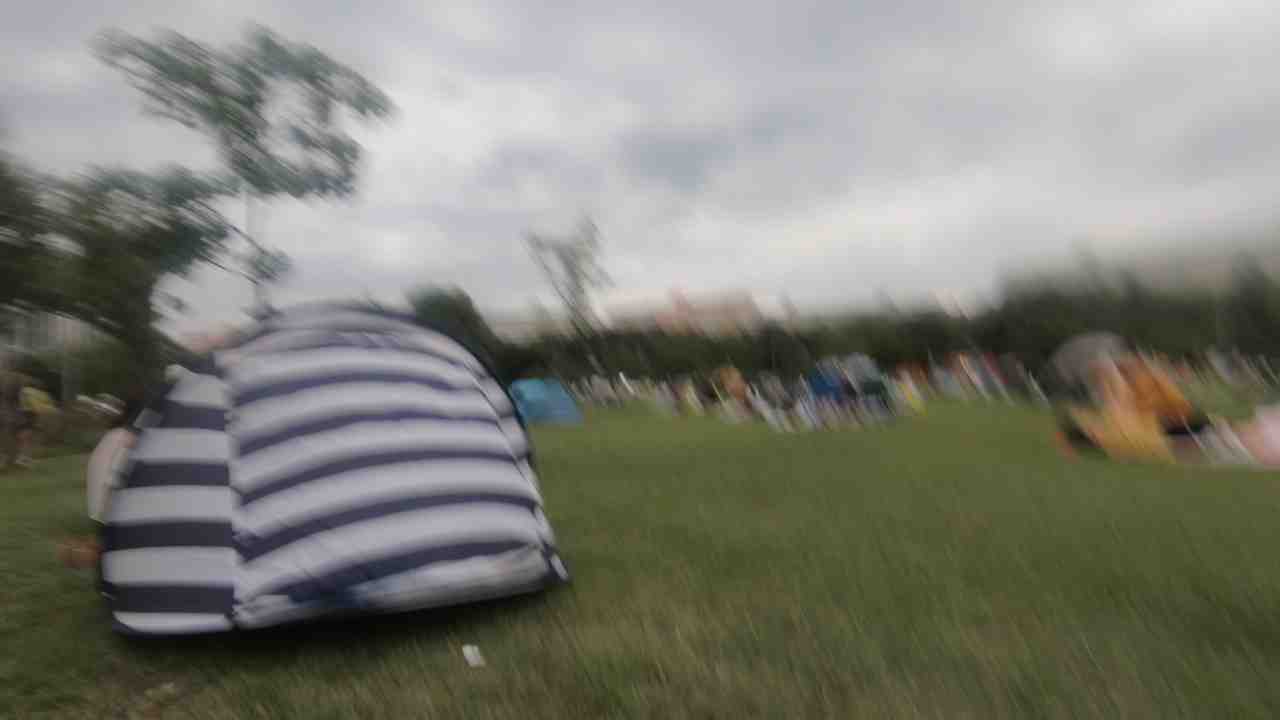}
    \includegraphics[width=0.19\textwidth]{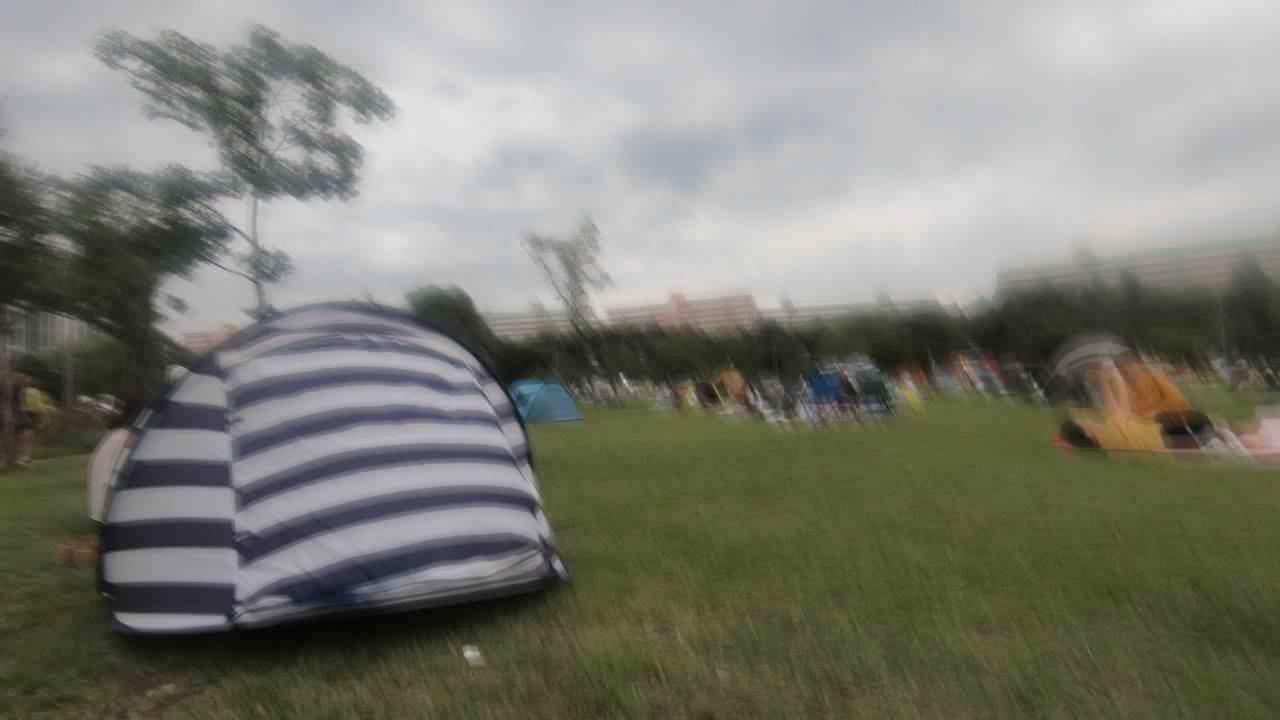}
    \includegraphics[width=0.19\textwidth]{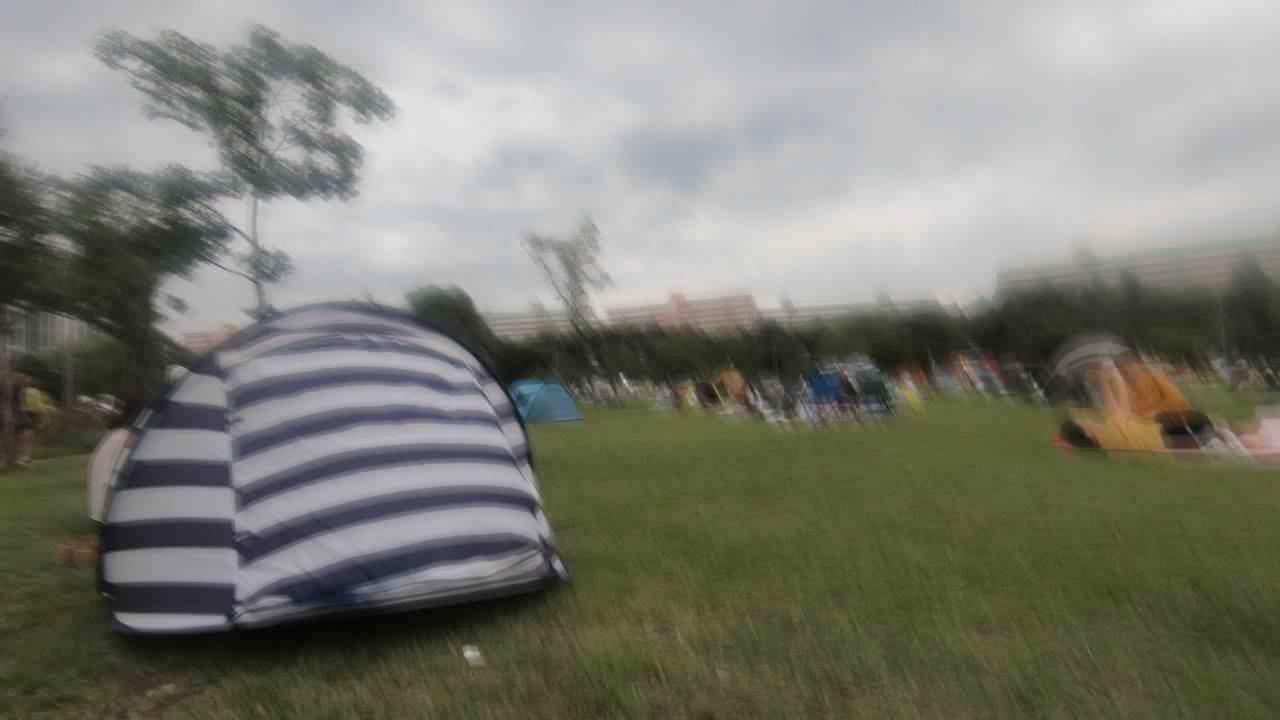}
    \includegraphics[width=0.19\textwidth]{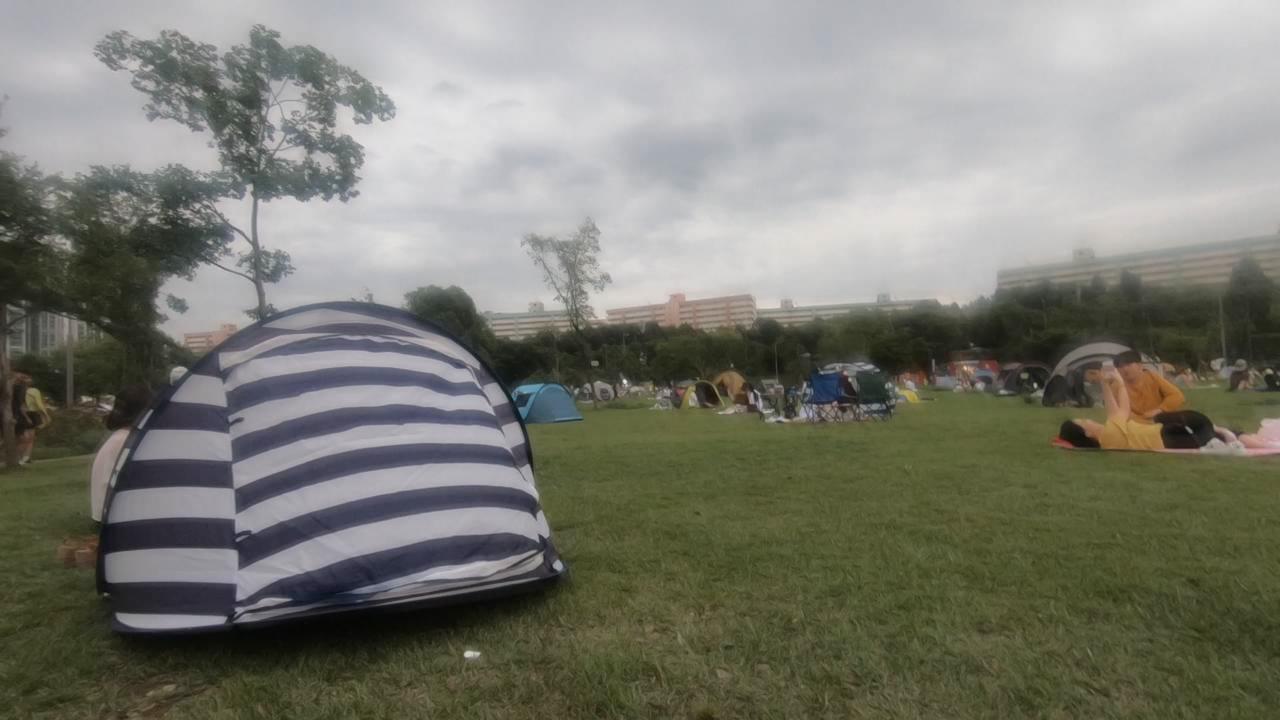}
    \includegraphics[width=0.19\textwidth]{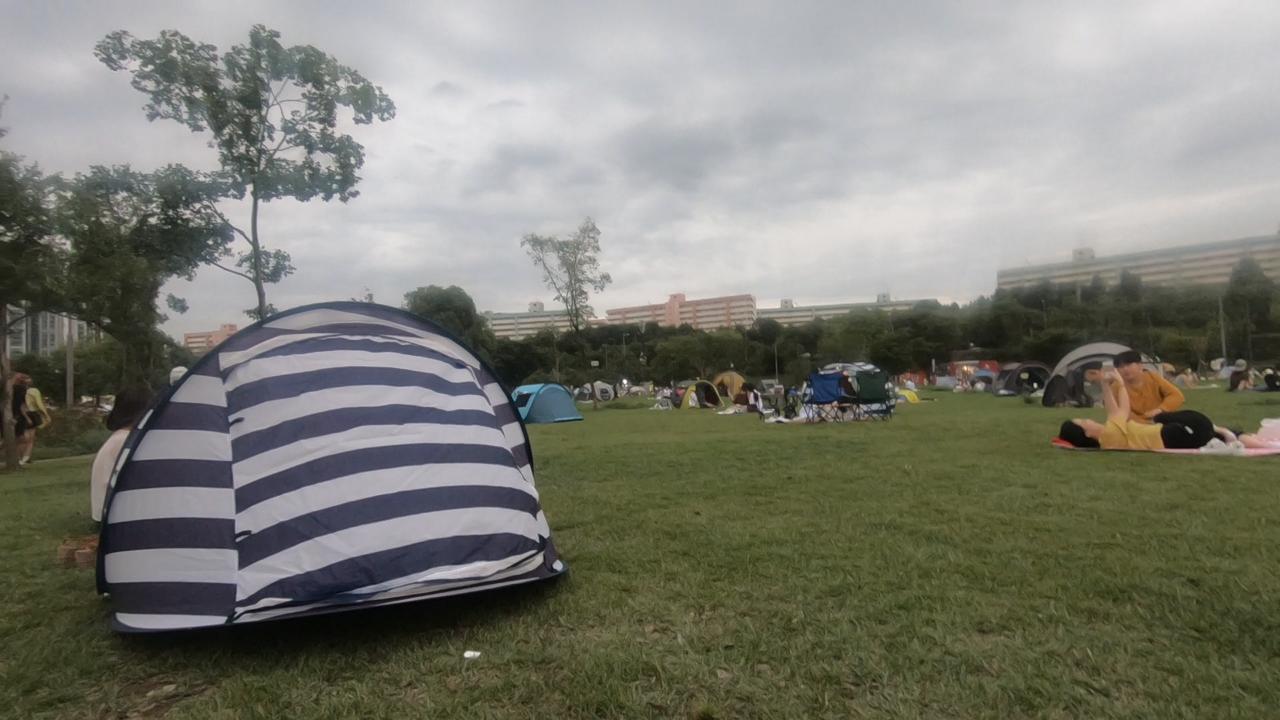}
    \\

    \vspace{2mm}
    
    \raisebox{0.01\height}{\makebox[0.02\textwidth]{\rotatebox{90}{\makecell{\small Haze\&Snow~\cite{chen2020srrs}}}}}
    \includegraphics[width=0.19\textwidth]{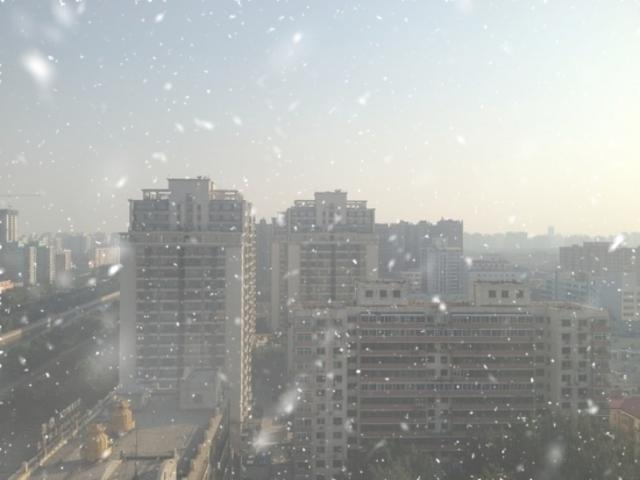}
    \includegraphics[width=0.19\textwidth]{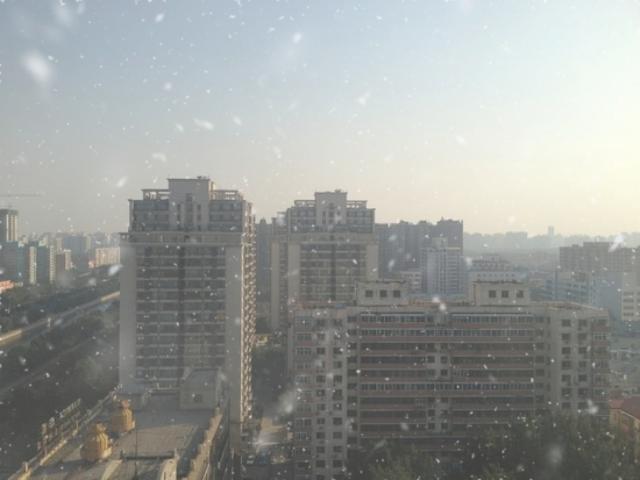}
    \includegraphics[width=0.19\textwidth]{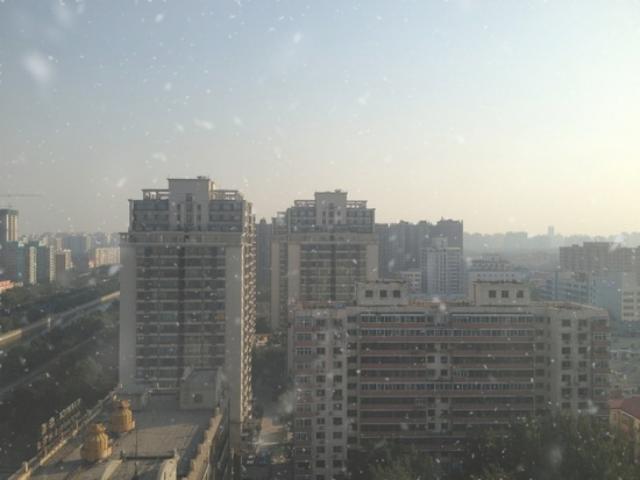}
    \includegraphics[width=0.19\textwidth]{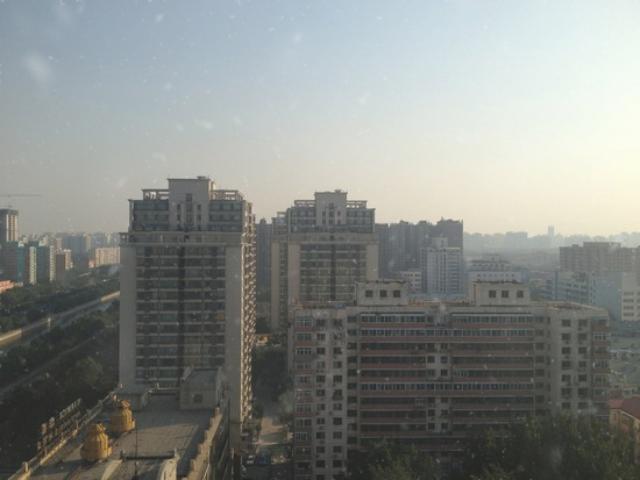}
    \includegraphics[width=0.19\textwidth]{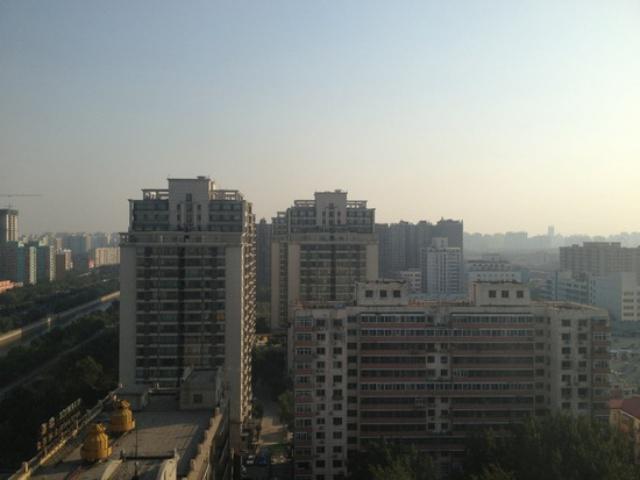}
    \\
    
    \makebox[0.02\textwidth]{}
    \makebox[0.19\textwidth]{\small Input}
    \makebox[0.19\textwidth]{\small Restormer~\cite{zamir2022restormer}}
    \makebox[0.19\textwidth]{\small PromptIR~\cite{potlapalli2024promptir}}
    \makebox[0.19\textwidth]{\small Ours-OH}
    \makebox[0.19\textwidth]{\small Ground truth}

    \caption{Qualitative results on mixed degradation scenarios. Specifically, blur and JPEG on the REDS~\cite{nah2019reds} dataset and haze and snow in the SRRS~\cite{chen2020srrs} dataset.} 
    \label{fig:qualitative_mixed}
\end{figure*}

\Cref{fig:qualitative_known} presents additional qualitative results for known degradations, including examples from both seen and unseen datasets. \Cref{fig:qualitative_unknown} highlights results for unknown degradations, with the corresponding mean absolute error (MAE) map displayed below each image to emphasize the differences. Finally, \Cref{fig:qualitative_mixed} shows qualitative results for mixed degradation scenarios.

{
    \small
    \bibliographystyle{ieeenat_fullname}
    \bibliography{main}
}

\end{document}